\documentclass[12pt,a4paper]{article}
\usepackage{amsmath,amssymb,amsthm,mathrsfs}
\usepackage{dsfont}
\usepackage[english]{babel}
\usepackage{graphicx,color}
\usepackage{subfigure,enumitem}
\usepackage[colorlinks]{hyperref}
\usepackage{mathtools}
\mathtoolsset{showonlyrefs}

\hypersetup{
linkcolor=blue,
citecolor=blue,
}
\usepackage{natbib}
\usepackage{siunitx} %

\graphicspath{{Figures/}}

\RequirePackage{algorithm}
\RequirePackage{algorithmic}
\renewcommand{\algorithmiccomment}[1]{\#\textit{#1}}

\makeatletter
\newenvironment{breakablealgorithm}
  {
   \begin{center}
     \refstepcounter{algorithm}
     \hrule height.8pt depth0pt \kern2pt
     \renewcommand{\caption}[2][\relax]{
       {\raggedright\textbf{\fname@algorithm~\thealgorithm} ##2\par}%
       \ifx\relax##1\relax 
         \addcontentsline{loa}{algorithm}{\protect\numberline{\thealgorithm}##2}%
       \else 
         \addcontentsline{loa}{algorithm}{\protect\numberline{\thealgorithm}##1}%
       \fi
       \kern2pt\hrule\kern2pt
     }
  }{
     \kern2pt\hrule\relax
   \end{center}
  }
\makeatother

\usepackage{microtype}
\usepackage{subfigure}
\usepackage{booktabs} 
\usepackage{hyperref}
\usepackage{afterpage}

\usepackage{tabularx}

\usepackage{amsmath,amsfonts}
\usepackage{scalerel,amssymb}
\usepackage{amsthm}

\usepackage{bm}
\newcommand{\R}{{\rm I\!R}}
\newcommand{\vect}{\mbox{vec}}
\newcommand{\svec}{\mbox{svec}}
\newcommand{\ivect}{\mbox{ivec}}
\newcommand{\diag}{\mbox{diag}}
\newcommand{\offd}{\mbox{offd}}
\newcommand{\mb}{\mathbf}
\newcommand{\A}{\mb{A}}

\newcommand{\G}{\mathcal{G}}
\newcommand{\N}{\mathrm{N}}
\newcommand{\KD}{K_{\mathrm{D}}}
\newcommand{\KF}{K_{\mathrm{F}}}

\newcommand{\KN}{K_{\mathrm{N}}}
\newcommand{\KG}{{\mathcal{K_G}}}

\newcommand{\projKG}{\mbox{Proj}_{{\mathcal{G}}}}
\newcommand{\projKN}{\mbox{Proj}_{{\mathcal{G}_{\N}}}}
\newcommand{\projKF}{\mbox{Proj}_{{\mathcal{G}_{\mathrm{F}}}}}
\newcommand{\projKD}{\mbox{Proj}_{{\mathrm{D}}}}

\newcommand{\AN}{A_{\N}}
\newcommand{\AF}{A_{\mathrm{F}}}
\newcommand{\spc}{\widehat{\mb{\Gamma}}_t(0)}
\newcommand{\spcone}{\widehat{\mb{\Gamma}}_t(1)}
\newcommand{\x}{\mb{x}}

\newcommand{\Ga}{\mathbf{\Gamma}_0}
\newcommand{\gaone}{\mathbf{\gamma}_1}
\newcommand{\as}{\mathbf{a}^s_1}
\newcommand{\wNone}{\mb{w}_{\N}^1}
\newcommand{\aslow}{\underline{\mathbf{a}}^s_1}
\newcommand{\bM}{\mb{b}_{\Bar{m}}}
\DeclareMathOperator*{\argmin}{arg\,min}

\newtheorem{theorem}{Theorem}[section]
\newtheorem{lemma}[theorem]{Lemma}
\newtheorem{assump}[theorem]{Assumption}
\newtheorem{corollary}{Corollary}[theorem]
\newtheorem{prop}[theorem]{Proposition}
\newtheorem{remark}{Remark}

\usepackage{mathabx}

\newcommand{\CB}[1]{{\color{blue} #1}}
\newcommand{\CR}[1]{{\color{black} #1}}
\usepackage{soul}

\numberwithin{equation}{section} 

\newcommand{\CM}[1]{\color{black} #1} 

\title{Online Graph Topology Learning from Matrix-valued Time Series}

\author{Yiye Jiang$^{1}$\footnote{Part of the work of this author was conducted while she was preparing her PhD at Université de Bordeaux.}, J\'{e}r\'{e}mie Bigot$^{2}$, Sofian Maabout$^{3}$\\
$^{1}$Universit\'e Grenoble Alpes, CNRS, Inria, Grenoble INP, LJK
\\  $^{2}$Institut de Math\'ematiques de Bordeaux, Universit\'e de Bordeaux\\  $^{3}$Laboratoire Bordelais de Recherche en Informatique, Universit\'e de Bordeaux
\vspace{0.1cm}   }

\begin{document}






\maketitle
\begin{abstract}
\CR{
The focus is on the statistical analysis of matrix-valued time series, where data is collected over a network of sensors, typically at spatial locations, over time. Each sensor records a vector of features at each time point, creating a vectorial time series for each sensor. The goal is to identify the dependency structure among these sensors and represent it with a graph. When only one feature per sensor is observed, vector auto-regressive (VAR) models are commonly used to infer Granger causality, resulting in a causal graph.  The first contribution extends VAR models to matrix-variate models for the purpose of graph learning. Additionally, two online procedures are proposed for both low and high dimensions, enabling rapid updates of coefficient estimates as new samples arrive. In the high-dimensional setting, a novel Lasso-type approach is introduced, and homotopy algorithms are developed for online learning. An adaptive tuning procedure for the regularization parameter is also provided. Given that the application of auto-regressive models to data typically requires detrending, which is not feasible in an online context, the proposed AR models are augmented by incorporating trend as an additional parameter, with a particular focus on periodic trends. The online algorithms are adapted to these augmented data models, allowing for simultaneous learning of the graph and trend from streaming samples. Numerical experiments using both synthetic and real data demonstrate the effectiveness of the proposed methods.}

\end{abstract}



\section{Introduction}
\label{sec: Intro}

The identification of graph topology responds to increasing needs for data representation and visualization in many disciplines, such as meteorology, finance, neuroscience, and social science. It is crucial to reveal the underlying relationships between data entries, even \textcolor{black}{in} settings where natural graphs are available. For example, \textcolor{black}{in} \citet{mei2016signal}, a temperature graph is inferred from the multivariate time series recording temperatures of cities around the continental United States over \textcolor{black}{a} one-year period. It differs from the distance graph, \textcolor{black}{yet it} exhibits better performance in predicting weather trends. Many methods have been proposed to infer graphs for various data processes and application settings \citep{sandryhaila2014big, dong2019learning}. The problem of graph learning is, given the observations of multiple features represented by random variables or processes, \textcolor{black}{to build or infer} the cross-feature relationship that takes the form of a graph, with the features termed as nodes. According to \textcolor{black}{the nature of} the data and the type of relationship, there are two main lines of work in the graph learning domain using statistical tools.

The first line considers the features represented by $N$ random variables $\mb{x} = (\bm{x}_1, \ldots, \bm{x}_N)^\top \in \R^N$ with i.i.d. observations. Moreover, it assumes that 
\begin{equation}\label{eq: ggm}
 \mb{x} \sim \mathcal{N}(0, P^{-1}).   
\end{equation}
The works are interested in inferring the conditional dependency structure among $\bm{x}_i, i = 1, \ldots, N$, which is encoded in the sparsity structure of precision matrix $P$. The resulting models are known as Gaussian graphical models \citep{meinshausen2006high, friedman2008sparse}, and the sparse estimators are called graphical lasso. There are also variants for stationary vector processes, see \citet{bach2004learning} and \citet{songsiri2010topology}, whereas the relationship considered is still the conditional dependence. 

The second line considers $N$ scalar processes, denoted by $\mb{x}_t \in \R^N$, and the inference of Granger causality \textcolor{black}{relationships} among them from the observed time series. \textcolor{black}{In contrast} to the Gaussian assumption, this line supposes that $\mb{x}_t$ is \textcolor{black}{a} vector auto-regressive (VAR) process
\begin{equation}\label{eq: varp}
    \mb{x}_t = \sum\limits_{l = 1}^p A^l\mb{x}_{t-l} + \mb{z}_t, \quad t \in \mathbb{Z},
\end{equation}
where $\mb{z}_t \sim \mathrm{WN}(0, \Sigma) $ is a white noise process with variance $\Sigma$, and $A^l \in \R^{N\times N}$ are the coefficients. For simplicity, we omit the intercept term here by assuming the process has zero mean. The process mean will be treated properly in Section \ref{sec: online learning aug}. Granger causality is defined pairwise: $\bm x_{it}$ is said to Granger cause $\bm x_{jt}, \, j \neq i$, if $\bm x_{jt}$ can be predicted more efficiently when the knowledge of $\bm x_{it}$ in the past and present is taken into account. \textcolor{black}{For a more technical definition, see} \citet[Section 2.3.1]{lutkepohl2005new}. The causal graph then refers to such a graph where each node represents a scalar process, and the edges represent Granger causality.

The advantage of \textcolor{black}{the} VAR assumption is that the topology of \textcolor{black}{the} causal graph is encoded in the sparsity structure of the coefficient matrices. More specifically, if the processes are generated by a stationary VAR($p$) model, then $\bm x_{it}$ does not cause $\bm x_{jt}$ if and only if all the $ji$-th entries of the true coefficient matrices $A^l_{ji} = 0, \, l = 1, \ldots, p,$ \citep[Corollary 2.2.1]{lutkepohl2005new}. Thus, the graph topology \textcolor{black}{can be retrieved} from the common sparsity structure in $A^l$. In \textcolor{black}{a} low-dimensional regime, this structure can be identified through \textcolor{black}{the} Wald test, which tests linear constraints for the coefficients.

The works in \textcolor{black}{the} literature therefore focus on the inference of causal graphs in \textcolor{black}{a} high-dimensional regime. The inference of the exact Granger causal graph is mainly considered in \citet{bolstad2011causal,zaman2020online}. \citet{bolstad2011causal} propose to use the group lasso penalty, $\lambda\sum_{i\neq j}\|(A^1_{ij}, \ldots, A^p_{ij})\|_{\ell_2}$, to the usual least squares problem of VAR($p$) models, in order to infer the common sparsity structure of coefficient matrices $A^l, l = 1, \ldots, p$. \citet{zaman2020online} \textcolor{black}{develop} the online procedure for this estimation problem. \citet{mei2016signal} define a variant of \textcolor{black}{the} VAR model, where the sparse structure of coefficients $A^l$ does not directly equal the graph topology, but the topology of $l$-hop neighborhoods. 
More specifically, they \textcolor{black}{assume} that $A^l = c_{l0}I + c_{l1}W + \ldots + c_{ll}W^l$, where $W$ is the adjacency matrix to infer, and $I$ is the identity matrix. Such models can thus capture the influence from more nodes. The estimation of the underlying adjacency matrix relies on the Lasso penalty to promote sparsity.

\begin{figure}[t]
  \begin{minipage}[b]{0.35\linewidth}
    \centering
    \includegraphics[width=\linewidth]{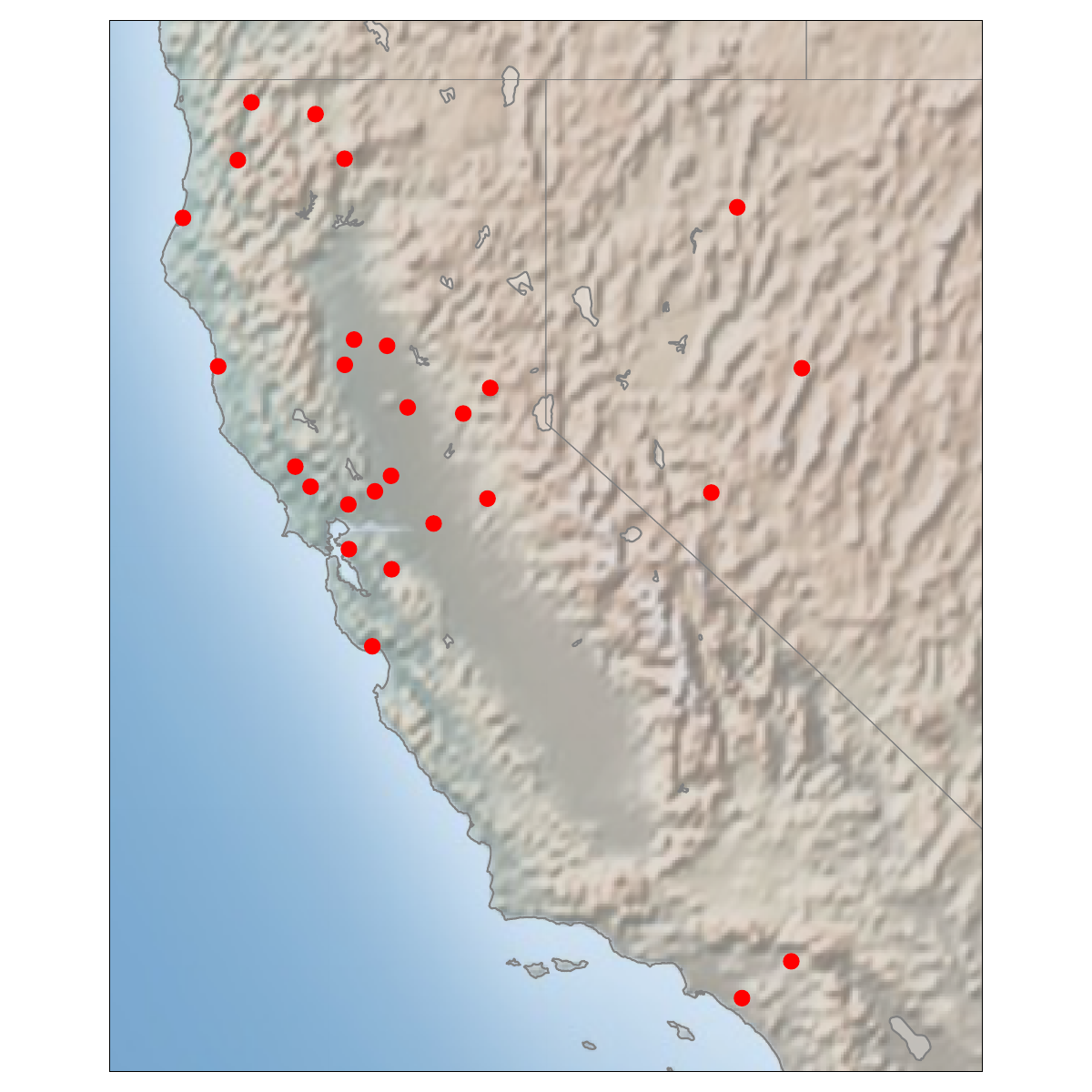}
  \end{minipage}%
  \begin{minipage}[b]{0.65\linewidth}
    \centering
    \includegraphics[width=\linewidth]{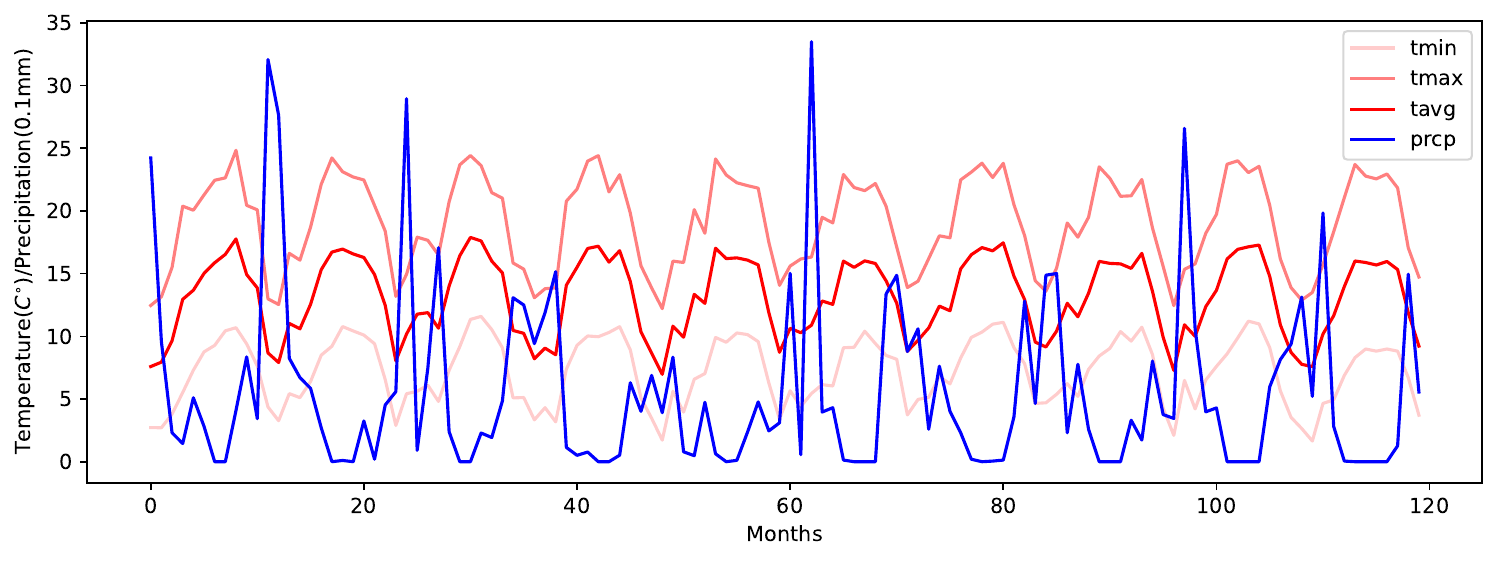} 
  \end{minipage}
  \caption{\textit{Monthly climatological records of weather stations in California.} On the left is the network of weather stations in California. On the right are demonstrated the \textit{vectorial} observations on a certain station $i$, where a vector $\x_{it} \in \R^4$ of min/max/avg temperature and precipitation is recorded per time $t$ at $i$, leading to $4$ scalar time series. The interest is in learning a graph of weather dependency for the network.}
  \label{fig: vector}
\end{figure}

We continue the second line of work by considering random processes in graph learning. However, instead of a scalar process for each feature, let us consider a vector process $\x_{it} \in \R^{F}$. The goal is to learn a graph of $N$ nodes that represents the causality structure among $\x_{it}, i = 1, \ldots, N$. An example data set of this setting is given in Figure \ref{fig: vector}. 
To this end, the classical VAR$(1)$ model is extended to a matrix-valued time series $\mb{X}_t = (\x_{1t}, \ldots, \x_{Nt}) \in \R^{N \times F}$. Second, estimation methods are developed, both in both low and high dimensions. In particular, the estimation is pursued in an \textit{online} fashion.

Graph learning from matrix-variate time series has not been considered previously in the literature, not even in the offline setting. In contrast, Gaussian graphical models and graphical Lassos (from the first line of work) have already been extended to the matrix-variate, and more generally to tensor-variate observations, $\mathcal{X}_t \in \R^{m_1 \times \ldots \times m_K}$, in several works \citep{zhou2014gemini, kalaitzis2013bigraphical, greeneWald2019tensor,wang2020sylvester}. 

In these extended works, the authors pointed out that one may straightforwardly apply the vector models to the vectorized data $\vect(\mathcal{X}_t)$. However, the resulting models will suffer from the quadratic growth of the number of edges with respect to the graph size $\prod_{k = 1}^K m_k$. The samples will \textcolor{black}{soon be} insufficient, and computational issues will appear. Additionally, different data dimensions will be lost after the vectorization. \textcolor{black}{To make models parsimonious yet flexible, the authors proposed applying} the vector models to the vectorized data \textcolor{black}{while imposing} certain structures on the parameter matrices. Such structures not only decrease the number of parameters but \textcolor{black}{are also} chosen to allow matrix representations and/or interpretations of the vector models. \textcolor{black}{This approach} brings twofold benefits. \textcolor{black}{First}, the data dimensions are \textcolor{black}{preserved} through the matrix representations, \textcolor{black}{enhancing} model interpretability. \textcolor{black}{Second}, the equivalent vector representations allow to apply the classical theoretical results and develop from that.

For example, \citet{zhou2014gemini} focused on the i.i.d. observations of a random matrix $\mathbf{X} = (\bm x_{ij})_{i,j} \in \R^{N\times F}$ and the inference of the conditional dependency structure among $\bm x_{ij}$. To this end, the author firstly applied the classical setting \eqref{eq: ggm} by assuming \begin{equation}\label{eq: mn vec}
    \vect(\mathbf{X})\sim \mathcal{N}(\mb{0}, P^{-1}). 
\end{equation}
Then a Kronecker product (KP) structure \citep{horn2012matrix} is imposed on the parameter matrix:
$$P = A \otimes B, A \in \R^{N\times N}, B \in \R^{F\times F}.$$
The KP structure reduces the number of parameters from $\mathcal{O}(N^2F^2)$ to  $\mathcal{O}(N^2 + F^2)$. It also allows a matrix representation of the model by 
\begin{equation}\label{eq: mn mat}
    \mathbf{X} = C^{-1}\mathbf{Z}D^{-1}, \; \mbox{ where } \vect(\mathbf{Z}) \sim \mathcal{N}(\mb{0}, I), A = D^\top D, B = C C^\top. 
\end{equation}
Note that, Equation \eqref{eq: mn vec} and  Equation \eqref{eq: mn mat} are equivalent definitions. 
Thanks to the vector representation, the conditional dependency structure among the $NF$ random variables $\bm x_{ij}$ can be easily retrieved from the sparsity structure of $P$. In this case, the KP structure also allows an interpretation of the proposed model. Since the KP structure in an adjacency matrix corresponds to a tensor product of sub-graphs, when taking $P$ as an adjacency matrix, the embedding of KP structure implies that the total conditional dependence structure is a product graph of the two sub-graphs defined respectively by $A$ and $B$. The two sub-precision matrices $A$ and $B$ are indeed separately associated with the row and column dimensions of $\bm X$, which can be equivalently shown by matrix representation \eqref{eq: mn mat}. To furthermore infer sparse graph structures, sparsity is supposed in $A$ and $B$.


In addition to the KP structure, another common structure to impose is the Kronecker sum (KS) \citep{horn2012matrix}. For example, \citet{kalaitzis2013bigraphical} performed matrix-variate graph learning by relying on: 
\begin{equation}\label{eq: ggm ks}
 \vect(\mb{X})\sim \mathcal{N}(\mu, P^{-1}), P = A \oplus B, A \in \R^{N\times N}, B \in \R^{F\times F}. 
\end{equation}
In this case, the \textcolor{black}{KS structure} does not allow an equivalent definition in matrix notation. However, it still \textcolor{black}{provides} an interpretation, similarly to the \textcolor{black}{KP structure}. By contrast to a tensor product, the KS structure in an adjacency matrix corresponds to a Cartesian product. Thus, the above model implies that the total conditional dependence structure is a Cartesian product graph of the two sub-graphs \textcolor{black}{corresponding to} the \textcolor{black}{row} and column dimensions.

Even though Models \eqref{eq: ggm ks} and \eqref{eq: mn mat} are  defined by the vectorized data, they take into account the row and column index of each entries $\bm x_{ij}$. Therefore, such models should be classified as matrix models instead of vector models.


Following the same principles to extend VAR$(1)$ models for matrix-variate time series $\bm X_t$, we first apply VAR$(1)$ models onto the vectorized data: 
\begin{equation}\label{eq: var mat}
    \vect{(\mb{X}_t)} = A\vect{(\mb{X}_{t-1})} + \mb{z}_t , \quad t \in \mathbb{Z}.
\end{equation}
Secondly, the KS structure is imposed on the coefficient matrix $A$:
$$A = A_F \oplus A_N, \, A_N \in \R^{N \times N}, A_F \in \R^{F \times F}.$$ Given the links between the KS and the matrix vectorization, the following matrix representation of Model \eqref{eq: var mat} holds:
\begin{equation}\label{eq: var mat mat}
    \mb{X}_t = A_N\mb{X}_{t-1} + \mb{X}_{t-1}A_F^\top + \mb{Z}_t, \quad t \in \mathbb{Z}. 
\end{equation}
Similarly to the cited works on matrix-variate Gaussian graphical models, the vector definition~\eqref{eq: var mat} indicates that the Granger causality structure among the $NF$ scalar processes $\bm x_{ijt}$ still corresponds to the sparsity structure of $A$. The KS structure furthermore implies this total causality structure is the Cartesian product of a row and a column sub-graphs, defined respectively by $A_N$ and $A_F$. More detailed explanation of how these $NF$ processes $\bm x_{ijt}$ interact through the KS product will be given in the next section. 

Given this extended auto-regressive model, this work mainly focuses on the estimation of $A_N$ and $A_F$. In particular, the estimation is pursued in an online fashion with sparsity supposed only in $A_N$, which raises technical challenges in both low and high dimensions. In this context, \textcolor{black}{the} primary contribution is \textcolor{black}{the extension of} the homotopy algorithms of classical Lasso to a new Lasso-type problem, which is motivated by the high dimensional inference. The derived algorithms are able to update the related Lasso estimator as either i) time evolves from $t$ to $t+1$; ii) the penalization parameter is updated; or iii) both i) and ii) occur simultaneously. Additionally, a data-adaptive tuning procedure is also derived, which can tune the penalization parameter value of the Lasso problem automatically when time evolves from $t$ to $t+1$. The contribution also relies on the fact that the derivations and the algorithms do not depend on the specific structure assumed in $A$, thus they can be applied to the online inference of many other parametric models under structure and sparsity assumptions. 

In addition to the new Lasso algorithms dedicated to high-dimensional inference, an online estimation procedure is also proposed \textcolor{black}{for} low dimension. Both online procedures take into account the presence of trends in time series; therefore, they can be applied directly \textcolor{black}{to} real time series. All the proposed methods constitute a realistic methodology to perform online graph learning of matrix-variate time series, which is \textcolor{black}{a novel data type} in the graph learning domain \textcolor{black}{to the best of our knowledge}. 

\textcolor{black}{Lastly, the literature on another domain, that of matrix auto-regressive models, should be mentioned.} Even though the extended VAR models are designed for graph learning, \textcolor{black}{they also enrich} the family of matrix auto-regressive (MAR) models. \textcolor{black}{The literature on matrix-valued time series models is thus briefly reviewed below.}

The MAR models are a very recent topic of interest to the statistics and econometrics communities. The first MAR model was introduced by \cite{chen2021autoregressive} in 2021, and can be written as:
\begin{equation}
    \mb{X}_t = A_r\mb{X}_{t-1}A_c^\top + \mb{Z}_t, \quad t \in \mathbb{Z}. 
\end{equation}
It can be defined equivalently in a vector form as:
\begin{equation}\label{eq: chen mar vat}
    \vect(\mb{X}_t) = A\vect(\mb{X}_{t-1}) + \mb{z}_t, \quad \mbox{where } A = A_c\otimes A_r, \quad t \in \mathbb{Z}. 
\end{equation}
Thus, this MAR model is also a special case of VAR models with a structure imposed on the parameter matrix. Indeed, the model followed the same principles of extension as we explained previously. The vector representation allowed them to derive the theoretical results, such as the stationarity conditions, by applying the general results of VAR models.  

Compared to our KS structure, their model corresponds to a KP structure. \textcolor{black}{The next section will also explain in detail how the $NF$ scalar processes $\bm x_{ijt}$ interact through the KP structure, providing a comparison to the chosen structure.} In addition to this difference in model formulation, there are distinctions in estimation. Firstly, our estimation method is online, \textcolor{black}{meaning that previous samples older than $t-1$ do not need to be stored} to calculate the estimates at time $t$. Additionally, our estimators for the row effect $\A_F$ are sparse, \textcolor{black}{facilitating the inference of} a sparse graph. The \textcolor{black}{corresponding} regularization also reduces the number of samples required in high-dimensional settings. \textcolor{black}{Numerical experiments will demonstrate that the proposed high-dimensional estimator requires fewer samples to be available and outperforms the estimators of \citet{chen2021autoregressive} with small sample sizes for both synthetic and real data.}

Given this first MAR models, several variants have been proposed. We classify them by their directions of extensions as follows. 
\begin{itemize}
    \item Tensor auto-regressive models: \cite{wang2024high} and \cite{li2021multi}.
    \item MAR Models with vector covariates: \cite{celani2023matrix} and \cite{sun2023matrix}. 
    \item Specialized MAR models for spatial data matrices:  \cite{sun2023matrix} and \cite{hsu2021matrix}. 
    \item MAR models with non-linear auto-regressive effects: mixture model \cite{wu2023mixture} which is defined by a probabilistic mixture of $K$ normal MAR models, and smooth transition model \cite{bucci2022smooth} which allows to switch smoothly between two MAR models. 
    \item MAR models with moving average terms: \cite{wu2023autoregressive}. 
\end{itemize}

As the initial MAR model of \cite{chen2021autoregressive}, these works in literature all rely on a KP structure, equivalently, a bilinear form to describe the row and column effects between $\mb{X}_t, \mb{X}_{t-1}$, and they also only consider the offline estimation. Thus our work has the corresponding contributions to the field of matrix auto-regressive models as well. It is of interest to compare the KS and the KP effects in the MAR model context. Since our model formulation and the data setting are closer to the MAR model in \cite{chen2021autoregressive} compared to other variants. We propose to compare our model with the one of \cite{chen2021autoregressive} with numerical experiments. \textcolor{black}{A comparison with the classical VAR(1) model is also included in the experiments to contrast the matrix models. The comparison results are provided in Section \ref{sec: experiments}.}

\paragraph*{Outlines}
\textcolor{black}{The proposed matrix-variate AR$(1)$ model is first recalled in Section \ref{sec: VAR model}, along with technical details. The different impacts of the KS and KP structures on the total causality relations are also explained. Next, in Section \ref{sec: online learning}, two online algorithms are developed for low and high dimensional settings, respectively. In Section \ref{sec: online learning aug}, the derived algorithms are augmented by additionally taking into account the trends of time series, primarily considering periodic trends, which consist of finite values. Lastly, the results from numerical experiments using both synthetic and real data are presented in Section \ref{sec: experiments}. All proofs and large algorithms are gathered in the technical appendices, and all notations are collected in Table \ref{tbl}.}

\begin{table}[ht]
    \centering
    \begin{tabularx}{\textwidth}{m{1.5cm} X}
     $\vect$ & Vectorized representation of a matrix. \\\vspace{0.1in}
     $\ivect$ & Inverse vectorized representation of a vector, such that $\ivect\circ\vect = id$.\\\vspace{0.1in}
     $[\cdot]_{\cdot}$ &  
     Extraction by index. The argument in $[]$ can be a vector or a matrix. For a vector, the index argument can be a scalar or an  \textit{ordered} list of integers. For example, $[\mbox{v}]_{k}$ extracts the $k$-th entry of v, while $[\mbox{v}]_{K} = ([\mbox{v}]_{k_i})_{i}$ extracts a sub-vector indexed by $K = (k_i)_i$ in order. 
     
     For a matrix, the index argument can be a pair of scalars or a pair of \textit{ordered} lists of integers. For example, $[M]_{k,k^\prime}$ extracts the $(k,k^\prime)$-th entry of $M$, while $[M]_{K,K^\prime} = ([M]_{k_i, k_j})_{i,j}$ extracts a sub-matrix indexed by $K = (k_i)_i$ in row order, and $K^\prime = (k^\prime_j)_j$ in column order. When $K = K^\prime$, we denote $[M]_{K,K^\prime}$ by $[M]_K$. \\\vspace{0.1in}
     $[M]_{:,i}$ & Extraction of the $i$-th column vector of matrix $M$. \\\vspace{0.1in}
     
     $[M]_{i,:}$ & Extraction of the $i$-th row vector of matrix $M$. \\\vspace{0.1in}
     
     $\svec(M)$ &  Vectorized representation of the upper diagonal part of matrix $M$, that is, $\left(\left(M\right]_{1,2}, \left[M\right]_{1,3}, \cdots, \left[M\right]_{2,3}, \cdots \right)^\top$. \\\vspace{0.1in}
     
     $\diag(M)$ & Diagonal vector of matrix $M$. \\\vspace{0.1in}
     
     $\offd(M)$ & $M$ with the diagonal elements replaced by zeros.  
\end{tabularx}
    \caption{\textit{Notations.}}
    \label{tbl}
\end{table}

\section{Causal Product Graphs and Matrix-variate AR(1) Models}\label{sec: VAR model}
In this section, \textcolor{black}{the complementary construction and assumptions of the proposed MAR model are presented}, which are needed for the theoretical properties. Then, \textcolor{black}{the interaction of processes described by the KS structure is explained in detail}, in contrast with the KP structure.

\textcolor{black}{The proposed KS-based MAR(1) model for a matrix-variate time series $\mb{X}_t \in \R^{N \times F}$ is first recalled as follows:}
\begin{equation}
    \mb{X}_t = A_N\mb{X}_{t-1} + \mb{X}_{t-1}A_F^\top + \mb{Z}_t, \quad t \in \mathbb{Z}.
\end{equation}
It can be defined equivalently as a special VAR(1) model as
\begin{equation}
    \vect{(\mb{X}_t)} = A\vect{(\mb{X}_{t-1})} + \mb{z}_t, \quad \mbox{where } A = \AF \oplus \AN, \quad t \in \mathbb{Z}. 
\end{equation}
This model is not completely identifiable, because $\AF \oplus \AN = (\AF + cI_F) \oplus (\AN-cI_N)$ holds for any scalar $c$. Thus to address this technical issue, we fix the diagonals of $\AN, \AF$ by $\diag(M_{\mathrm{F}}) = 0, \; \diag(M_{\N}) = 0$. In return, we estimate the diagonal of $A$ independently of $\AN, \AF$. In terms of formula, this means the model becomes 
\begin{equation}
\begin{aligned}
\vect{(\mb{X}_t)} = A\vect{(\mb{X}_{t-1})} + \mb{z}_t, \quad \mbox{where } &\offd(A) = \AF \oplus \AN, \\
&\diag(\AN) = 0, \mbox{and } \diag(\AF) = 0, \quad t \in \mathbb{Z}.    
\end{aligned}
\end{equation}
This actually gives more freedom to the model. \textcolor{black}{In addition}, we require that the component graphs, and hence the product graph, be symmetric. Namely, $\AN = \AN^\top$ and $\AF = \AF^\top$. This is because \textcolor{black}{it is observed that existing causal graphs are usually directed}, which disables their further use in methods that require undirected graphs as prior knowledge, such as kernel methods and graph Fourier transform-related methods. Therefore, the focus is on learning undirected graphs. Nevertheless, it is emphasized that the derived approaches do not depend on the specific structure of the coefficients, \textcolor{black}{and thus can be adapted to, for example, a relaxed constraint set without the symmetry assumption}.

The model definition so far is concluded as follows. The matrix-variate stochastic process $\mb{X}_t \in \R^{N \times F}$ is said matrix-variate AR($1$) process if the multivariate process $\mb{x}_t := \vect \left(\mb{X}_t \right)$ is a VAR($1$) process 
\begin{equation}\label{eq: var1_vect}
    \mb{x}_t =A\mb{x}_{t-1} + \mb{z}_t, \quad t \in \mathbb{Z}, 
\end{equation}
with $A$ having the particular KS structure $\KG$
\begin{eqnarray}\label{eq: KG def}
\KG  &= &\big\{M \in \R^{NF \times NF}: \exists \, M_{\mathrm{F}} \in \R^{F \times F}, M_{\N} \in \R^{N \times N}, \; \mbox{such that} , \\
& &  \hspace{0.1in} \mbox{offd}(M) = M_{\mathrm{F}} \oplus M_{\N}, \, \mbox{with,} \; \diag(M_{\mathrm{F}}) = 0, \; \diag(M_{\N}) = 0,\\\vspace{0.1in}
& & \hspace{0.1in} M_{\mathrm{F}} = M_{\mathrm{F}}^\top, \; M_{\mathrm{N}} = M_{\mathrm{N}}^\top\big\}.
\end{eqnarray}
Accordingly, the matrix representation changes to 
\begin{equation}\label{eq: var1_matrix}
    \mb{X}_t = D\circ\mb{X}_{t-1} +  \AN\mb{X}_{t-1} + \mb{X}_{t-1}\AF^\top + \mb{Z}_t.  
\end{equation}
where $\circ$ is Hadamard product, $D \in \R^{N\times F} = \ivect(\diag(A))$, $\AN$ and $\AF$ are the adjacency matrices such that $\mbox{offd}(A) = \AF \oplus \AN$, and $\mb{Z}_t = \ivect(\mb{z}_t)$. 

Lastly, to make the model above stationary, corresponding model conditions are needed. Note that the vector representation is a special case of VAR(1) models, as we mentioned for general VAR models in Equation \eqref{eq: varp}, the sparsity structure of $A$ defines the Granger causality dependency structure of the $NF$ component processes $x_{ijt}$ only if the model is stationary. Thus, we apply the classical stationarity results on Model \eqref{eq: var1_vect}, which leads to the model condition 
\begin{assump}\label{assup: stationarity}
    $\|A\|_2 < 1$.
\end{assump}
Note that we assume here the process mean is zero and derive the main frameworks in Section \ref{sec: online learning}. In Section \ref{sec: online learning aug}, we will study the model with non-zero but time-variant process mean, namely, process trend, and we will adapt the derived frameworks to the augmented model. 

\begin{figure}[ht]
    \centering
    \includegraphics[width=0.7\textwidth]{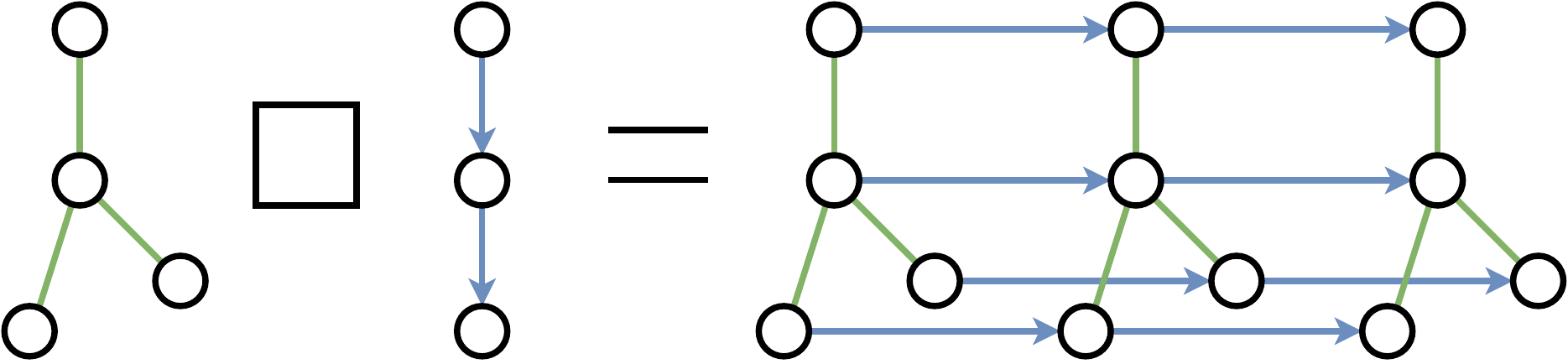}\vspace{0.2in}
    \includegraphics[width=0.7\textwidth]{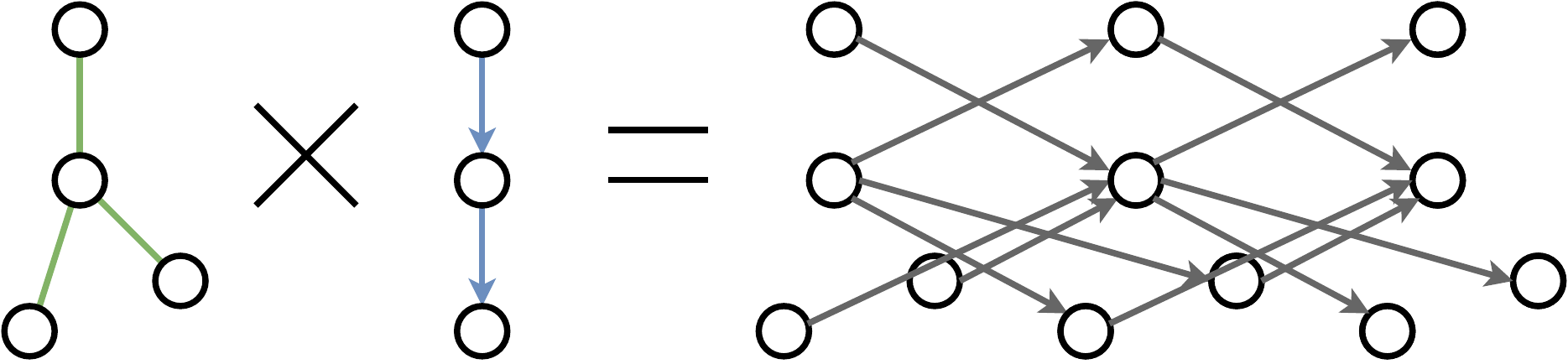}
    \caption{\textit{Comparison of the Cartesian and the tensor products of graphs.} The node set of both product graphs is the Cartesian product of the components' node sets, yet follows the different adjacencies. The example is based on \citet[Figure 2]{sandryhaila2014big}.}
    \label{fig-intro: product graphs}
\end{figure}

\textcolor{black}{The KS (our model) and KP (model of \citet{chen2021autoregressive}) structures are now compared when imposed on the coefficient $A$ of VAR(1) models.} As mentioned before, when KP and KS structures are present in adjacency matrices, it implies in both cases that, the corresponding graphs can factorize into smaller graphs. To see the difference of factorization, let $\AF, \AN$ be the adjacency matrices of two graphs $\G_{\mathrm{F}}, \G_{\mathrm{N}}$, then the KP $\AF\otimes \AN$ and the KS $\AF\oplus \AN$ are respectively the adjacency matrices of their tensor product graph $\G_\mathrm{N}\times\G_\mathrm{F}$ and Cartesian product graph  $\G_\mathrm{N}\Box\G_\mathrm{F}$ \citep{sandryhaila2014big}. We illustrate these two product graphs in Figure \ref{fig-intro: product graphs}. For the formal definitions of Cartesian and tensor products of graphs, we refer to \citet{hammack2011handbook,chen2015weak,imrich2018cartesian}. 

Figure \ref{fig-intro: product graphs} shows that, the two product graphs differ greatly. 
For example, the lattice-like structure of the Cartesian product preserves the sub-graphs in all sections of both dimensions. By contrast, the tensor product focuses on the cross-dimensional connection, yet abandoning the intra-dimensional dependency. This later property actually refers to, in the Gaussian process literature, the cancellation of inter-task transfer, see for example, \citet[Section 2.3]{bonilla2007multi}. Therefore when the nodes represent $\bm x_{it}^f, i = 1,\ldots,N, f = 1, \ldots, F$, imposing KP structure \citep{chen2021autoregressive} implies assuming no causality dependencies among $\bm x_{it}^f, i = 1,\ldots,N$ for each $f$ fixed, which represent the observations of the feature $f$ at different nodes across the network. By contrast, the coefficient matrix $A$ of KS structure is able to take such dependencies into account during inference, which are in effect present in many applications. This justifies our choice. 

\section{Online Graph Learning}\label{sec: online learning}
In this section, \textcolor{black}{two learning frameworks are developed} to estimate $A$ recursively. The first method is valid in low dimensional regime, where the number of samples along time is assumed to be sufficiently large with respect to the number of parameters. By contrast, the second method based on a Lasso-type problem requires fewer samples, and it is thus adapted to high-dimensional regime. \textcolor{black}{A general learning framework is especially considered,} where the partial sparsity is pursued in the estimation of only $\AN$. This is motivated by the fact that, merely a very small number of features $F$ are usually present in applications. Thus, the feature graph can be reasonably assumed fully-connected. On the other hand, since the partial sparsity constraint is also a technically more complicated case for the proposed high dimensional learning method, given its corresponding resolution, the adaption to the case of fully sparsity does not require novel techniques. In the following section, we firstly introduce the tools on constraint set $\KG$, which are crucial to derive the proposed frameworks.

\subsection{Orthonormal Basis and Projection Operator of $\KG$}\label{sec: basis}
\begin{figure}[ht]
\begin{center}
\includegraphics[width=0.35\textwidth]{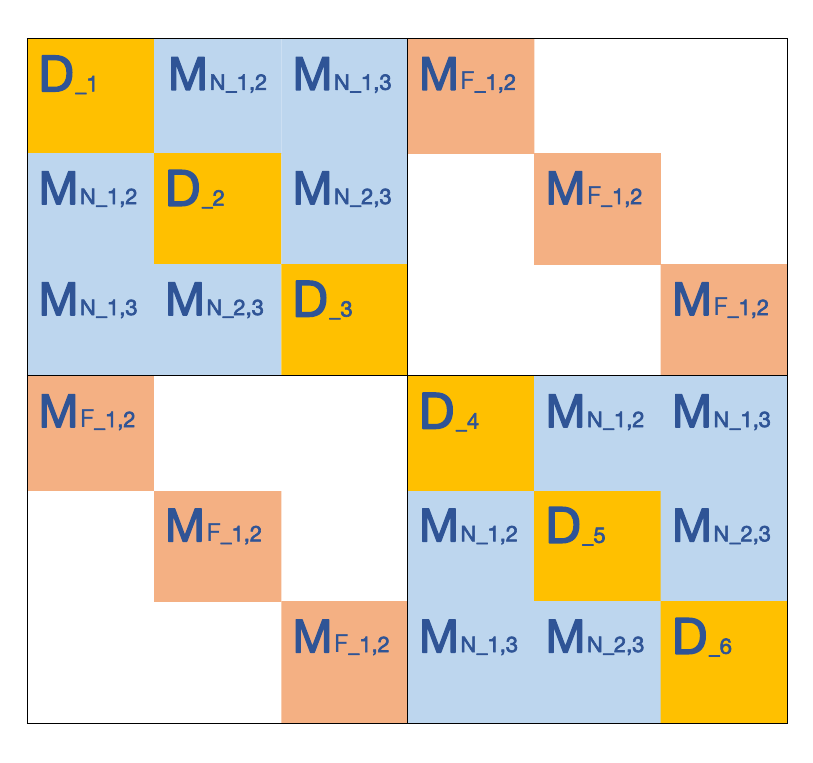}
\includegraphics[width=0.6\textwidth]{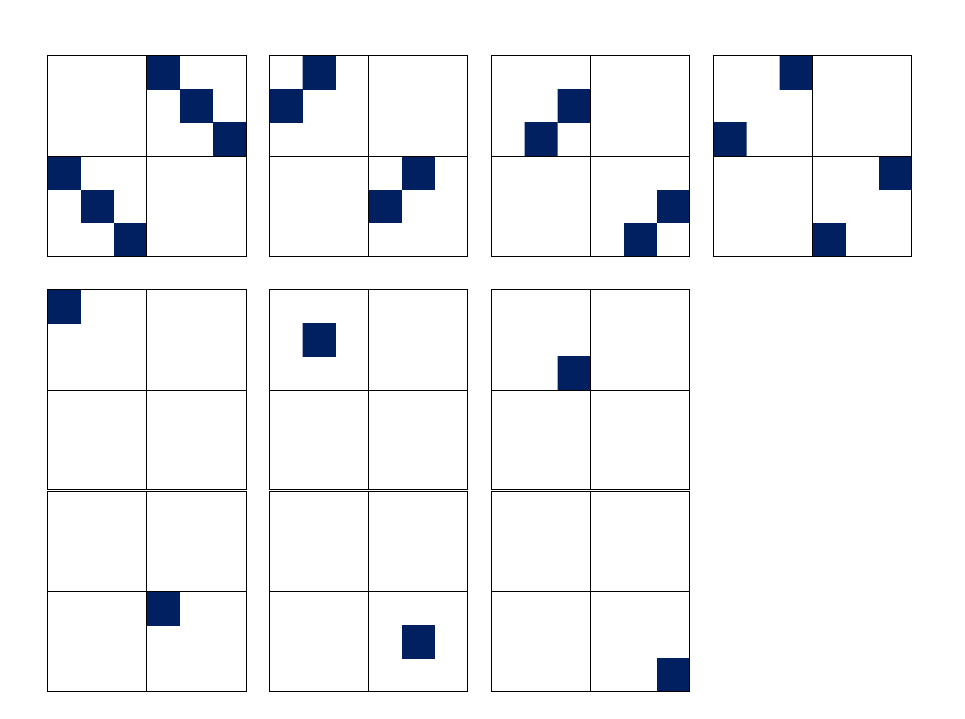}
\caption{\small \textit{Matrices $(U_k)_k$ as entry locators, which characterise the structure of $\KG$.}}
\label{fig-online: ep orthogonal basis}
\end{center}
\end{figure}

$\KG$ defined as Equation \eqref{eq: KG def} is a linear space of dimension $NF + \frac{1}{2}F(F-1) + \frac{1}{2}N(N-1)$. We now endow $\KG$ with the Frobenius inner product of matrix, that is $\langle B,C\rangle_{\mathbf{F}} = tr(B^\top C)$. The orthogonal basis of $\KG$ is then given in the following Lemma.
\begin{lemma}\label{lem: basis_KG}
The set of matrices $U_k, \; k \in K := \{1, \ldots, NF + \frac{1}{2}F(F-1) + \frac{1}{2}N(N-1)\}$, 
defined below form an orthogonal basis of $\KG$
\begin{equation}\label{eq: ortho basis}
U_{k} = 
\begin{cases}
    &E_{k}, \quad k \in \KD := \{ 1, ..., NF\}, \\
    &\frac{1}{2N} E_k \otimes I_N, \quad k \in \KF := NF + \{ 1, \ldots, \frac{1}{2} F(F-1)\},\\
    &\frac{1}{2F} I_F \otimes E_k , \quad k \in \KN := NF + \frac{1}{2} F(F-1) + \{1, \ldots, \frac{1}{2}N(N-1) \},
\end{cases}
\end{equation}
where when $k \in \KD$, $E_k \in \R^{NF\times NF}, \mbox{ with } [E_k]_{i,j} = 1, \mbox{ if } i=j=k, \mbox{ otherwise } 0$, when $k \in \KF$, $E_k \in \R^{F\times F}$ is almost a zero matrix except
\begin{equation}
\begin{cases}
    &[E_k]_{1,2} = [E_k]_{2,1} = 1, \mbox{if } k = {NF+1}, \\
    &[E_k]_{1,3} = [E_k]_{3,1} = 1, \mbox{if } k = {NF+2}, \\
    &[E_k]_{2,3} = [E_k]_{3,2} = 1, \mbox{if } k = {NF+F}, \\ &\vdots\\
    &[E_k]_{F-1,F} = [E_k]_{F,F-1} = 1, \mbox{if } k = {NF+ \frac{1}{2} F(F-1)}, 
\end{cases}    
\end{equation}
when $k \in \KN$, $E_k \in \R^{N\times N}$ is almost a zero matrix except
\begin{equation}
\begin{cases}
    &[E_k]_{1,2} = [E_k]_{2,1} = 1, \mbox{if } k = {NF+\frac{1}{2} F(F-1)+1}, \\
    &[E_k]_{1,3} = [E_k]_{3,1} = 1, \mbox{if } k = {NF+\frac{1}{2} F(F-1)+2}, \\
    &[E_k]_{N-1,N} = [E_k]_{N,N-1} = 1, \mbox{if } k = {NF+ \frac{1}{2} F(F-1)+\frac{1}{2} N(N-1)}. 
\end{cases}    
\end{equation}
\end{lemma}
In Figure \ref{fig-online: ep orthogonal basis}, an example of this orthogonal basis of $\KG$ for $N=3, F=2$, \textcolor{black}{is given}, where $U_k$ are visualized with respect to their non-zero entries.
\textcolor{black}{It can be found that} each $U_k$ relates to one variable of $\diag(M), M_\mathrm{F}$ and $M_\N$, and characterises how it contributes to the structure of $M$ by repeating at multiple entries. Thus, taking the inner product with $U_k$ actually calculates the average value of an arbitrary matrix over these entries. This is important to understand how to project an arbitrary matrix onto $\KG$.

It is easy to verify that $\langle U_k,U_{k^\prime}\rangle_{\mathbf{F}} = 0$ for any $k \neq k^\prime$ in $K$, and $(U_k)_k$ spans $\KG$. 
Thus the normalized matrices $U_k/\|U_k\|_\mathbf{F}, k \in K$ form an orthonormal basis of $\KG$. We introduce the orthogonal projection onto $\KG$ and provide an explicit formula to calculate it using $(U_k/\|U_k\|_\mathbf{F})_k$ in Proposition \ref{prop: projection_KG}. 
\begin{prop}\label{prop: projection_KG}
For a matrix $A  \in \R^{NF\times NF}$, its orthogonal projection onto $\KG$ is defined by 
\begin{equation}
    \projKG(B) = \argmin_{M \in \KG} \left\|B - M\right\|_\mb{F}^2.
\end{equation}
Then given the orthonormal basis $U_k/\|U_k\|_\mathbf{F}, k \in K$, the projections can be calculated explicitly as  
\begin{equation}
  \projKG(B) = \sum_{k\in K} \left\langle U_k,B \right\rangle \, \frac{1}{\|U_k\|_\mathbf{F}^2} U_k. 
\end{equation}
\end{prop}
The projection is straightforward to understand. To obtain a variable in $\diag(M), M_F$, and $M_N$ related to $U_k$, \textcolor{black}{the average value of $B$ is calculated using $\langle U_k, B\rangle$ as explained previously. This average value is then placed at} the corresponding entries to construct the structure by multiplying the locator $U_k/\|U_k\|_\mathbf{F}^2$.

Furthermore the orthogonality of the basis implies the direct sum
\begin{equation}\label{eq: direct sum}
    \KG = \mathcal{K}_{\mathrm{D}} \oplus \mathcal{K}_{\mathrm{F}} \oplus \mathcal{K}_{\N},
\end{equation}
where $\mathcal{K}_{\mathrm{D}}$, $\mathcal{K}_{\mathrm{F}}$, and  $\mathcal{K}_{\N}$ are respectively spanned by $(U_k)_{k \in \KD}$, $(U_k)_{k \in \KF}$, and $(U_k)_{k \in \KN}$. Given the construction of $(U_k)_k$, Equation \eqref{eq: direct sum} actually reveals the product graph decomposition, note that equally we have
$$\mathcal{K}_{\mathrm{D}} =
\{M \in \R^{NF \times NF}: \mbox{offd}(M) = 0\},$$
\begin{eqnarray*}
\mathcal{K}_{\mathrm{F}} &=&
\{M \in \R^{NF \times NF}: \exists M_{\mathrm{F}} \in \R^{F \times F}, \; \mbox{such that} , \\
& &  \hspace{0.1in} M = M_{\mathrm{F}} \otimes I_N, \, \mbox{with,} \; \diag(M_{\mathrm{F}}) = 0, M_{\mathrm{F}} = M_{\mathrm{F}}^\top\}, 
\end{eqnarray*}
\begin{eqnarray*}
\mathcal{K}_{\mathrm{N}} &=&
\{M \in \R^{NF \times NF}: \exists M_{\N} \in \R^{N \times N}, \; \mbox{such that} , \\
& &  \hspace{0.1in} M = I_F \otimes M_{\N}, \, \mbox{with,} \; \diag(M_{\N}) = 0, M_{\mathrm{N}} = M_{\mathrm{N}}^\top\}. 
\end{eqnarray*}
The projection onto these subspaces can also be computed analogously:
\begin{equation}
    \projKD(B) = \sum_{k\in \KD} \langle U_k,B\rangle \, \frac{1}{\|U_k\|_\mathbf{F}^2} U_k, \mbox{ that is the diagonal part of } B,
\end{equation}
\begin{equation}
    \mbox{Proj}_{\mathrm{F}}(B) = \sum_{k\in \KF} \langle U_k,B\rangle \, \frac{1}{\|U_k\|_\mathbf{F}^2} U_k = \left[\sum_{k\in \KF} \langle U_k,B\rangle E_k\right] \otimes I_N,
\end{equation}
\begin{equation}
    \mbox{Proj}_{\mathrm{N}}(B) = \sum_{k\in \KN} \langle U_k,B\rangle \, \frac{1}{\|U_k\|_\mathbf{F}^2} U_k =  I_F \otimes \left[\sum_{k\in \KF}\langle U_k,B\rangle E_k\right].
\end{equation}
$\projKF(B)$ and $\projKN(B)$ \textcolor{black}{are used} to denote the small matrices $\sum_{k\in \KF} \langle U_k,B\rangle E_k$ and $\sum_{k\in \KF}\langle U_k,B\rangle E_k$, with an extra subscript $\mathcal{G}$, with which we will represent the proposed estimators of $\AF, \AN$ in the following sections. Finally, \textcolor{black}{it holds:}
\begin{equation}\label{eq: projection_KG}
  \projKG(B) = \projKD(B) + \projKF(B) \oplus \projKN(B).
\end{equation}

\subsection{Approach 1: Projected OLS Estimators and Wald Test} \label{sec:app1}
In low dimensional regime, Model \eqref{eq: var1_vect} can be fitted by the ordinary least squares method as a VAR model. Assume that we start receiving samples $\x_0, \x_1, ..., \x_t$ from time $\tau = 1$, the OLS estimator for an intercept-free VAR($1$) model is given by
\begin{equation}
    \widecheck{\A}_t = \spcone \left[\spc\right]^{-1},  
\end{equation}
where 
$$\spc = \frac{1}{t} \sum_{\tau = 1}^t \x_{\tau - 1}\x_{\tau - 1}^\top,$$
and 
$$\spcone = \frac{1}{t} \sum_{\tau = 1}^t \x_{\tau}\x_{\tau - 1}^\top,$$
respectively estimate the auto-covariance matrices $\Gamma(0)$ and $\Gamma(1)$, with $\Gamma(h) = \mathbb{E}\left(\mb{x}_t\mb{x}_{t-h}^{\top}\right), \, h \geq 0$. 

The classical results \citep[Section 3.2.2]{lutkepohl2005new} show that the OLS estimator of a stationary VAR model is consistent, moreover, permits a central limit theorem (CLT), if the following assumptions are satisfied. 
\begin{assump}\label{assump: ols}
\begin{enumerate}
    \item The white noise is independent, namely, $\mb{z}_t \perp\!\!\!\!\perp \mb{z}_{t^\prime}, \; t \neq t^\prime$.
    \item The white noise has singular covariance, and bounded fourth moments. 
    \item The initial sample is already drawn from a stationary model, in other words, it permits the causal representation $\mb{x}_0 = \sum_{j=0}^\infty A^j\mb{z}_{t-j}$.
\end{enumerate}
\end{assump}
Given Assumption \ref{assump: ols} and the stationary condition \ref{assup: stationarity}, the classical results show that 
\begin{enumerate}
\item $\widecheck{\A}_t \xrightarrow{p} A$,
\item $\sqrt{t} \, \vect(\widecheck{\A}_t - A) \xrightarrow{d} \mathcal{N}(0, \Sigma_{ols})$,
where $\Sigma_{ols} = \left[\Gamma(0)\right]^{-1} \otimes \Sigma$,
\end{enumerate}
where $A$ is the true coefficient which has the structure of $\KG$. 

However, in practice, due to model misspecification and limited samples, $\widecheck{\A}_t$ will not have the same structure as $A \in \KG$. Because the structure is crucial to retrieve the inferred graphs. We propose to perform the projection of $\widecheck{\A}_t$ onto $\KG$, which leads to the projected OLS estimator: $$\widehat{\A}_t := \projKG(\widecheck{\A}_t).$$
Given the representation \eqref{eq: projection_KG}, it is natural to define the estimators of $\diag(A)$, $\AF$, and $\AN$ by $\projKD(\widecheck{\A}_t)$, $\projKF(\widecheck{\A}_t)$, and $\projKN(\widecheck{\A}_t)$, respectively, denoted by $\widehat{\A_{\mathrm{D}}}_{,t}$, $\widehat{\bm \AF}_{,t}$, and $\widehat{\bm \AN}_{,t}$. 
We now establish the Wald test with $\widehat{\bm \AN}_{,t}$ to identify the sparsity structure of the true $\AN$. To this end, we provide the CLT in Theorem \ref{thm: asym_AN}.  

\begin{theorem}\label{thm: asym_AN}
Assume samples $\x_0, \x_1, ..., \x_t$ follow Model \eqref{eq: var1_vect}, and Assumptions \ref{assump: ols} and \ref{assup: stationarity} is satisfied, then the following CLT holds for $\widehat{\bm \AN}_{,t}$, as $t \to + \infty$,
\begin{equation}
\sqrt{t} \, \svec(\widehat{\bm \AN}_{,t} - \AN) \xrightarrow{d} \mathcal{N}(0, \Sigma_{\N}),
\end{equation}
where 
$$
\Sigma_{\N} = \hspace{-0.2cm} \sum_{k, k' \in \KN} \hspace{-0.2cm} \vect(U_k)^\top \Sigma_{ols} \vect(U_{k'}) \, \left(\svec(E_k)\svec(E_{k'})^\top\right).
$$
\end{theorem}

The proof is done through applying Cram\'er-Wold theorem on $\sqrt{t} \, \svec(\widehat{\bm \AN}_{,t} - \AN)$, given the linearity of $\projKN(\widecheck{\A}_t)$ and the CLT of classical OLS estimator $\widecheck{\A}_t$. For details, see  \ref{sec: proof CLT AN_hat}, where we also derive a CLT for $\widehat{\A}_t$. 

It is straightforward to understand the asymptotic distribution of $\widehat{\bm \AN}_{,t}$. The asymptotic covariance between its two entries is assigned the mean of covariance values $\vect(U_k)^\top \Sigma_{ols} \vect(U_{k'})$, following the construction of the corresponding  estimators $\langle U_k, \widecheck{\A}_t\rangle$ and $\langle U_{k'}, \widecheck{\A}_t\rangle$ as averages as well. 

Based on this large sample result, to retrieve the sparsity structure of $\AN$, we now test the nullity of $P$ given variables $\left[\AN\right]_{i_k, j_k}, \; k = 1, ..., P$, with $i_k < j_k$ as
$$
H_0: \alpha = 0 \;  \mbox{versus} \; H_1: \alpha \neq 0,
$$
where $\alpha \in \R^{P} := \left(\cdots ,\left[\AN\right]_{i_k, j_k}, \cdots\right)^\top$.
The test statistic is given by
\begin{equation}\label{eq: Wald statistic}
\lambda_{W, t} = t \, \hat{\bm \alpha}_t^\top \left[\widehat{\mb{\Sigma}}_{W,t}\right]^{-1} \hat{\bm \alpha}_t,     
\end{equation}
where $\hat{\bm \alpha}_t \in \R^{P} := \left(\cdots ,\left[\widehat{\bm \AN}_{,t}\right]_{i_k, j_k}, \cdots\right)^\top$,  
and $\widehat{\mb{\Sigma}}_{W,t} \in \R^{P \times P}$ is defined as
$$
\left[\widehat{\mb{\Sigma}}_{W,t}\right]_{k, k'} \hspace{-0.2in} = \vect(U_{h_k})^\top\widehat{\mb{\Sigma}}_{ols,t}\vect(U_{h_{k'}}),
$$
such that $U_{h_k}$ is the matrix corresponding to variable $\left[\AN\right]_{i_k, j_k}$, 
$$\widehat{\mb{\Sigma}}_{ols,t} = \left[\spc\right]^{-1} \otimes \widehat{\mb{\Sigma}}_t, \mbox{ and } \widehat{\mb{\Sigma}}_t = \spc - \spcone \left[\spc\right]^{-1}\spcone^\top,$$
are the consistent estimators. CLT \eqref{thm: asym_AN} implies the following result. 
\begin{corollary}\label{coro: Wald test} The asymptotic distribution of $\lambda_{W,t}$ as $t \to + \infty$ is given by
$$\lambda_{W,t} \xrightarrow{d} \mathcal{\chi}^2(P), \quad \mbox{Under } \; H_0.$$
\end{corollary}
\begin{remark}
We can also consider the test statistic $\lambda_{F,t} := \lambda_{W,t}/P$ as suggested in \citet[Section 3.6]{lutkepohl2005new} in conjunction with the critical values from $F(P, t - NF - 1)$.
\end{remark}

\begin{remark}
In practice, the significance level of $\chi^2$ test is a hyperparameter which can control sparsity of the estimator.
\end{remark}

The Wald test above theoretically completes the approach. In practice, we propose to test the $p$ entries of the  smallest estimate magnitudes, jointly each time, as $p$ grows from $1$ to possibly largest value $|\KN|$. Specifically, for a given estimation $\widehat{\bm \AN}_{,t}$, we first sort its entries such that 
$$|[\widehat{\bm \AN}_{,t}]_{i_1, j_1}| \leq |[\widehat{\bm \AN}_{,t}]_{i_2, j_2}| \leq \ldots \leq |[\widehat{\bm \AN}_{,t}]_{i_{|\KN|}, j_{|\KN|}}|.$$
Then, we set up the sequence of joint tests
$$H_0(1), \, H_0(2), \, ..., \, H_0(|\KN|), \mbox{ where } H_0(p): \left([{\AN}_{,t}]_{i_1, j_1}, \cdots, [{\AN}_{,t}]_{i_p, j_p}\right)^\top = 0, 
$$
\textcolor{black}{These tests are performed} sequentially until $H(p_0 + 1)$ is rejected for some $p_0$. Lastly, the entries $[\widehat{\bm \AN}_{,t}]_{i_1, j_1}, ..., [\widehat{\bm \AN}_{,t}]_{i_{p_0}, j_{p_0}}$ are replaced with $0$ in $\widehat{\bm \AN}_{,t}$ as the final estimate of $\AN$. Note that searching for $p_0$ resembles root-finding, since the output from each point $p$ is binary. Thus, the search can be accelerated by using the bisection method, with the maximal number of steps being approximately $\log_2(|\KN|)$. 

The previous procedure is performed at the $t$-th iteration, given the OLS estimator $\widecheck{\A}_t$ and the consistent estimator $\widehat{\mb{\Sigma}}_{ols,t}$. 
When new sample $\x_{t+1}$ comes, $\widecheck{\A}_{t+1}$ and $\widehat{\mb{\Sigma}}_{ols,t+1}$ can be calculated efficiently by applying \textit{Sherman Morrison formula} on $[\spc]^{-1}$. The pseudo code is given in Algorithm \ref{alg: proj_Wald}.

\subsection{Approach 2: Structured Matrix-variate Lasso and Homotopy Algorithms}  \label{sec:app2}
As discussed at the introduction, a common practice in the literature to identify the sparsity structure of VAR coefficients in high dimensional regime is to adopt Lasso estimators. The one used in \citet{bolstad2011causal,zaman2020online} is defined as the minimizer of Lasso problem \eqref{eq-intro: vecLasso} in the VAR($1$) case. 
\begin{equation}\label{eq-intro: vecLasso}
    \min_{A} \frac{1}{2t} \sum\limits_{\tau = 1}^t \left\|\x_{\tau} - A\x_{\tau - 1}\right\|_{\ell_2}^2 + \lambda_t \left\|A\right\|_{\ell_1}, 
\end{equation}
where $\x_{\tau}$ is a vector of sample, which can be taken as $\vect(\bm X_{\tau})$ for example. Lasso \eqref{eq-intro: vecLasso} is the most standard Lasso in literature \citep[Section 3.4.2]{hastie2009elements}. A wide variety of frameworks from convex analysis and optimization have been adapted to compute its solutions for different scenarios, for example, coordinate descent \citep{friedman2010regularization}, proximal gradient methods \citep{beck2009fast}, and a more Lasso-specific technique least angle regression \citep{efron2004least}. However, Lasso \eqref{eq-intro: vecLasso} is not able to estimate the structured $A$ with the sparse component $\AN$. Therefore motivated by the estimation, we propose the novel Lasso type problem \eqref{eq: matLasso} 
\begin{equation}\label{eq: matLasso}
    \A(t, \lambda_t) = \argmin_{A \in \KG} \frac{1}{2t} \sum\limits_{\tau = 1}^t \left\|\x_{\tau} -  A\x_{\tau - 1}\right\|_{\ell_2}^2 + \lambda_t F \left\|\AN\right\|_{\ell_1}.
\end{equation}
\textcolor{black}{It can be observed that} the proposed Lasso problem differs from the classical Lasso \eqref{eq-intro: vecLasso} by the structure constraint $\A \in \KG$ and the partial sparsity regularization on $\AN$ instead of on all parameters $A$. The ordinary resolution of Lasso \eqref{eq: matLasso} can be done by applying, for example, the proximal gradient descent \citep{parikh2014proximal}. \textcolor{black}{Within the algorithm framework,} the structure constraint and the partial sparsity do not pose additional difficulties, since only the gradient with respect to $\R^{NF \times NF}$ is calculated in the forward step. \textcolor{black}{These details are provided in} \ref{app: pgd}.

\textcolor{black}{At this point, the focus is on providing} algorithms to quickly update the previous solutions for the change in the hyperparameter value or in the data term. This different goal requires to consider specific methods. For classical Lasso, the framework of homotopy continuation methods \citep{osborne2000new} has been explored \citep{malioutov2005homotopy,garrigues2008homotopy} to calculate the fast updating. Since the homotopy algorithm is derived from the optimality condition, which is with respect to the matrices in $\KG$ for Lasso \eqref{eq: matLasso}, requiring to consider the gradient with the structure, thus the existing homotopy algorithms for classical Lasso are not applicable. Therefore in the following, we first calculate the optimality condition of Lasso \eqref{eq: matLasso} in Section \ref{sec: op cond}, based on the expression of projection onto $\KG$. Then we derive the two homotopy algorithms in Sections \ref{sec: homo1} and \ref{sec: data path}, respectively for the updating paths $\A(t, \lambda_1) \rightarrow \A(t, \lambda_2)$ and $\A(t, \lambda_2) \rightarrow \A(t+1, \lambda_2)$, together with an adaptive tuning procedure for the regularization hyperparameter. 

Therefore, the online algorithm consists in performing the three steps in the order:
\begin{equation}\label{eq-intro: online homo pipeline}
\begin{aligned}
&\mbox{Step 1 : }\ \lambda_t \rightarrow \lambda_{t+1}, \quad \mbox{Step 2 : } \A(t, \lambda_t) \rightarrow \A(t, \lambda_{t+1}), \\
&\mbox{Step 3 : } \A(t, \lambda_{t+1}) \rightarrow \A(t+1, \lambda_{t+1}). 
\end{aligned}
\end{equation}

\subsubsection{Optimality Conditions}\label{sec: op cond}
The key point in deriving the optimality conditions arising from the variational problem (\ref{eq: matLasso}) is to transfer the structure of $A$ onto the data vector $\x_{\tau - 1}$, using an orthonormal basis of $\KG$.
We introduce the auxiliary variable $A^0$, such that $A = \projKG(A^0)$, and rewrite Problem (\ref{eq: matLasso}) with respect to $A^0$
\begin{equation}\label{eq: prob_A0}
\begin{aligned}
    \min_{A^0 \in \R^{NF\times NF}} \;  &\frac{1}{2t} \sum\limits_{\tau = 1}^t \left\|\x_{\tau} -\sum_{k \in K} \langle U_{k},A^0\rangle  \, \frac{1}{\|U_k\|_\mathbf{F}^2}U_{k} \x_{\tau - 1}\right\|_{\ell_2}^2 \\
    &\hspace{0.5in}+ \lambda \left\|\sum_{k \in \KN} \langle U_{k},A^0\rangle  \, \frac{1}{\|U_k\|_\mathbf{F}^2}U_{k}\right\|_{\ell_1}.  
\end{aligned}
\end{equation}
Problem \eqref{eq: prob_A0} is weakly convex, since a minimizer of (\ref{eq: matLasso}) can be projected from infinitely many minimizers of (\ref{eq: prob_A0}).
We still use $L_{\lambda,t}$ to denote the objective function above. A minimizer $\A^0$ of (\ref{eq: prob_A0}) satisfies the optimality conditions
\begin{equation}\label{eq: opcon_A0}
\begin{aligned}
0 \in \; \frac{\partial L_{\lambda,t}}{\partial A^0} = & \sum_{k, k' \in K}  \, \langle U_{k},U_{k'}\spc\rangle \, \langle \frac{1}{\|U_{k'}\|_\mathbf{F}^2} U_{k'}, \A^0 \, \rangle \frac{1}{\|U_k\|_\mathbf{F}^2} U_{k} \\
& - \sum_{k \in K} \langle U_{k},\spcone\rangle  \, \frac{1}{\|U_k\|_\mathbf{F}^2}U_{k} + \lambda \sum_{k \in \KN} \partial \left |\langle U_{k},\A^0\rangle \right| \, \frac{1}{\|U_k\|_\mathbf{F}^2}U_{k}.
\end{aligned}    
\end{equation}
Assume $\A^0$ is a matrix which satisfies Equation \eqref{eq: opcon_A0}, hence a minimizer of Problem \eqref{eq: prob_A0}. Then $\A = \projKG(\A^0)$ is a minimizer of Lasso \eqref{eq: matLasso}. We denote its active set $\{k \in \KN: \langle U_k, \A^0 \rangle \neq 0\}$ by $\KN^1$, that is all the non-zero variables of $\bm \AN$, and its non-active set by $\KN^0$, that is $\KN \setminus \KN^1$. Since $(U_{k})_{k\in K}$ is an orthogonal family, Equation \eqref{eq: opcon_A0} is equivalent to 
\begin{equation}\label{eq: op cond DF}
0 = \sum_{k\in \KD \bigcup \KF} \left[\sum_{k' \in K}  \, \langle U_{k},U_{k'}\spc\rangle \, \langle \frac{1}{\|U_{k'}\|_\mathbf{F}^2}U_{k'}, \A^0 \, \rangle  -  \langle U_{k},\spcone\rangle \right] \frac{1}{\|U_k\|_\mathbf{F}^2} U_{k},
\end{equation}
\begin{equation}\label{eq: op cond KN1}
\begin{aligned}
0 = & \sum_{k\in \KN^1} \left[\sum_{k' \in K}  \, \langle U_{k},\spc\rangle \, \langle \frac{1}{\|U_{k'}\|_\mathbf{F}^2} U_{k'}, \A^0 \, \rangle  -  \langle U_{k},\spcone\rangle \right] \frac{1}{\|U_k\|_\mathbf{F}^2} U_{k} \\
& \hspace{1in} + \lambda \sum_{k \in \KN^1} \mbox{sign} \langle U_{k},\A^0\rangle \, \frac{1}{\|U_k\|_\mathbf{F}^2}U_{k},
\end{aligned}    
\end{equation}
\begin{equation}\label{eq: op cond KN0}
\begin{aligned}
0 = & \sum_{k\in \KN^0} \left[\sum_{k' \in K}  \, \langle U_{k},U_{k'}\spc\rangle \, \langle \frac{1}{\|U_{k'}\|_\mathbf{F}^2} U_{k'}, \A^0 \, \rangle  -  \langle U_{k},\spcone\rangle \right] \frac{1}{\|U_k\|_\mathbf{F}^2} U_{k} \\
& \hspace{1in} + \lambda \sum_{k \in \KN^0} \partial \left |\langle U_{k},\A^0\rangle \right| \, \frac{1}{\|U_k\|_\mathbf{F}^2}U_{k}, \mbox{ where } \partial \left |\langle U_{k},\A^0\rangle \right| \in [-1,1].
\end{aligned}    
\end{equation}

The optimality conditions above are an extension of those for classical Lasso. Compared to the conditions of classical Lasso which are given by two linear systems, ours are furthermore refined to three corresponding to the addition situation where a subset of variables in the problem are not penalized. 

\textcolor{black}{Matrix representations for} Equations \eqref{eq: op cond DF}, \eqref{eq: op cond KN1}, and \eqref{eq: op cond KN0} \textcolor{black}{are also derived}, which facilitate the interpretation of the condition. \textcolor{black}{For details, see} \ref{app: op mat}.

\subsubsection{Homotopy from $\A(t, \lambda_1)$ to $\A(t, \lambda_{2})$}\label{sec: homo1}

\textcolor{black}{To develop the homotopy algorithm for changes in the $\lambda$ value, it is necessary to obtain} the formulas of the active variables indexed by $\KN^1$ in terms of $\lambda$. To this end, \textcolor{black}{all the model variables are first reorganized into a vector:}
$$\mathbf{a}^s := \left(\langle \frac{1}{\|U_k\|_\mathbf{F}^2} U_k, \A^0 \rangle\right)_{k \in K} = \left(\langle \frac{1}{\|U_k\|_\mathbf{F}^2} U_k, \A \rangle\right)_{k \in K}.$$
Note that $\mathbf{a}^s$ is in fact the scaled Lasso solution by the time the variable repeats. Then optimality conditions \eqref{eq: op cond DF}, \eqref{eq: op cond KN1}, and \eqref{eq: op cond KN0} are essentially a system of linear equations of unknown $\mathbf{a}^s$, with $\lambda$ in the coefficients. Thus we aim to firstly represent this linear system in vector form, in order to solve the unknowns. We shall introduce the following notations. 

\paragraph*{Notations of Proposition \ref{prop: opcon}.} $\Ga \in \R^{|K| \times |K|}$ is a large matrix defined as 
$$\left[\Ga\right]_{k,k'} = \langle U_{k}, U_{k'}\spc\rangle.$$
$\gaone \in \R^{|K|}$ is a long vector defined as $$\left[\gaone\right]_{k} = \langle U_{k},\spcone\rangle. $$
$\mb{w} \in \R^{|K^1|}$ is a long vector where $[\mb{w}]_k$ is defined as 
\begin{equation}
\begin{cases}
= 0, & k \in \KD \bigcup \KF, \\
= \mbox{sign} [\mb{a}^s]_k, & k \in \KN^1, \\
\in [-1,1], & k \in \KN^0. 
\end{cases}   
\end{equation}
We define $K^1 := \KD \bigcup \KF \bigcup \KN^1 $, that are all the \textit{non-zero} variables. Note that except the computational coincidence, the variables in $\KD \bigcup \KF$ are usually non-zero. Then we denote the extractions 
\begin{equation}\label{eq: extraction}
\begin{aligned}
&\Ga^1 = \left[\Ga\right]_{K^1}, 
\Ga^0 = \left[\Ga\right]_{\KN^0, K^1}, \gaone^1 = \left[\gaone\right]_{K^1}, 
\gaone^0 = \left[\gaone\right]_{\KN^0}, \\
&\as = [\mb{a}^s]_{K^1}, \mb{w}_1 = [\mb{w}]_{K^1}, \mb{w}_0 = [\mb{w}]_{\KN^0}.
\end{aligned}
\end{equation}

\textcolor{black}{Any order can be assigned to the elements in} $K^1, \KN^0$ to extract the rows/columns/entries above, as long as the order is applied consistently to all extractions. \textcolor{black}{With these notations, a system of linear equations can be retrieved from} Equations \eqref{eq: op cond DF}, \eqref{eq: op cond KN0}, \eqref{eq: op cond KN1} \textcolor{black}{for the unknowns} $\as$. Each equation is obtained by equating the entries of one $U_{k}$. The resulting system is given in Proposition \ref{prop: opcon}.

\begin{prop}\label{prop: opcon} 
A minimizer of Lasso problem (\ref{eq: matLasso}) satisfies the linear system 
\begin{equation}\label{eq: opcon}
\Bigg\{
\begin{aligned}
& \Ga^1 \as - \gaone^1 + \lambda \mb{w}_1 = 0, \\
& \Ga^0 \as - \gaone^0 + \lambda \mb{w}_0 = 0.
\end{aligned}   
\end{equation}
\end{prop}
The representation of the optimality conditions in Equation (\ref{eq: opcon}) are similar to those of classical Lasso \citep{garrigues2008homotopy, malioutov2005homotopy}, where $\Ga$, $\gaone$ with the embedded structures correspond to $\mb{X}^\top\mb{X}$,  $\mb{X}^\top\mb{y}$ in the optimality conditions of classical Lasso. However in our case, the non-zero and sign pattern are only with respect to the entries of $\AN$, thus $\mb{w}_1$, which is the equivalent of sign vector, has $|\KD| + |\KF|$ zeros. 

Suppose that $\A(t, \lambda)$ is the unique solution for a fixed $\lambda$ of the optimization problem \eqref{eq: matLasso}, then we invert $\Ga^1$ in Proposition \ref{prop: opcon} and get the formulas of $\as$
\begin{equation}\label{eq: closed form}
\Bigg\{
\begin{aligned}
& \as  = \left[\Ga^1\right]^{-1} (\gaone^1 - \lambda \mb{w}_1), \\
& \lambda \mb{w}_0 = \gaone^0 - \Ga^0 \as.
\end{aligned}   
\end{equation}
Formula \eqref{eq: closed form} is determined by the active set and the sign pattern of the optimal solution at $\lambda$. It shows that $\as$ is a piecewise linear function of $\lambda$, while $\mb{w}_0$ is also a piecewise smooth function. 

Therefore, with the assumptions that $\left[\mb{a}^s\right]_{\KN^1} \neq 0$ (element-wise), and $|\left[\mb{w}\right]_{\KN^0}| < 1$ (element-wise), due to continuity properties, there exists a range $(\lambda_l, \lambda_r)$ containing $\lambda$, such that for any $\lambda' \in (\lambda_l, \lambda_r)$, element-wise,  $\left[\mb{a}^s\right]_{\KN^1}$ remains nonzero with the signs unchanged, and  $\left[\mb{w}\right]_{\KN^0}$ remains in $(-1, 1)$. Hence, Formula \eqref{eq: closed form} is the closed form of all the optimal solutions $\A(t, \lambda')$, for $\lambda' \in (\lambda_l, \lambda_r)$. $\lambda_l, \lambda_r$ are taken as the closest critical points to $\lambda$. Each critical point is a $\lambda$ value which makes either an $\left[\mb{a}^s\right]_{k}, \, k \in \KN^1$ become zero, or a $\left[\mb{w}\right]_{k}, \, k \in \KN^0$ reach $1$ or $-1$. By letting $\left[\mb{a}^s\right]_{k} = 0, \, k \in \KN^1$ and $\left[\mb{w}\right]_k = \pm 1, \, k \in \KN^0$ in Formula \eqref{eq: closed form}, we can compute all critical values. We now use $k_i$ to denote the orders of $K^1, \KN^0$ that we used in the extraction \eqref{eq: extraction}. The critical values are then given by
\begin{equation}\label{eq: crit_point}
\begin{aligned}
    &\lambda^0_{k_i} = \left[ \left[\Ga^1\right]^{-1} \gaone^1\right]_{i} / \left[ \left[\Ga^1\right]^{-1} \mb{w}_1\right]_{i}, \quad k_i \in K^1 \mbox{ such that } k_i \in \KN^1, \\
    &\lambda^+_{k_i} = \frac{\left[\gaone^0 -  \Ga^0\left[\Ga^1\right]^{-1} \gaone^1\right]_{i}}{\left[1 - \Ga^0\left[\Ga^1\right]^{-1} \mb{w}_1\right]_{i}}, \quad k_i \in \KN^0, \\  
    &\lambda^-_{k_i} = \frac{\left[\gaone^0 -  \Ga^0\left[\Ga^1\right]^{-1} \gaone^1\right]_{i}}{\left[-1 - \Ga^0\left[\Ga^1\right]^{-1} \mb{w}_1\right]_{i}}, \quad k_i \in \KN^0. 
\end{aligned}
\end{equation}
Thus, the closet critical points from both sides are
\begin{equation}\label{eq: lambda_range}
\begin{aligned}
&\lambda_l := \max\big\{\max\{\lambda^0_k, k \in \KN^1: \lambda^0_k < \lambda\}, \\
& \hspace{1in}\max\{\lambda^+_k, k \in \KN^0: \lambda^+_k < \lambda\}, \max\{\lambda^-_k, k \in \KN^0: \lambda^-_k < \lambda\}\big\}, \\
&\lambda_r := \min\big\{\min\{\lambda^0_k, k \in \KN^1: \lambda^0_k > \lambda\}, \\
& \hspace{1in}\min\{\lambda^+_k, k \in \KN^0: \lambda^+_k > \lambda\}, \min\{\lambda^-_k, k \in \KN^0: \lambda^-_k > \lambda\}\big\}.
\end{aligned}
\end{equation}
If $\lambda_l = \emptyset$ then $\lambda_l := 0$, while if $\lambda_r = \emptyset$ then $\lambda_r := +\infty$.
After $\lambda'$ leaves the region by adding or deleting one variable to or from the active set, we update in order the corresponding entry in $\mb{w}$, $K^1, \KN^0$, and the solution formula \eqref{eq: closed form} (Sherman Morrison formula for one rank update of $[\Ga^1]^{-1}$) to calculate the boundary of the new region as before. We proceed in this way until we reach the region covering the $\lambda$ value at which we would like to calculate the Lasso solution, and use Formula (\ref{eq: closed form}) in this final region to compute the $\as$ with the desired $\lambda$ value. Lastly, we retrieve the matrix-form optimal solution based on $\as$ and the latest $K^1$.
This completes the first homotopy algorithm. The detailed algorithm see  \ref{app: homo1}. 

\subsubsection{Homotopy from $\A(t,\lambda)$ to $\A(t+1,\lambda)$}\label{sec: data path}

We recall again the classical Lasso in Equation \eqref{eq: vecLasso}. We formulate it with vectorial parameter here. 
\begin{equation}\label{eq: vecLasso}
\bm\theta(t, \lambda) =  \argmin_{\theta \in \R^{d}} \frac{1}{2} \|\mb{y} - \mb{X}\theta\|_{\ell_2}^2 + t\lambda\|\theta\|_{\ell_1},
\end{equation}
where $\mb{y} = (\bm y_{1}, \ldots, \bm y_{t})^\top$, $\mb{X} = (\x_{1}, \ldots, \x_{t})^\top$, and $\bm y_{\tau} \in \R, \x_{\tau} \in \R^{d}$ are the samples at time $\tau$. \citet{garrigues2008homotopy} propose to introduce a continuous variable $\mu$ in Lasso \eqref{eq: vecLasso}, leading to the optimization Problem \eqref{eq: vecLassohomo} 
\begin{equation}\label{eq: vecLassohomo}
\min_{\theta \in \R^{d}} \frac{1}{2} \|\mb{y} - \mb{X}\theta\|_{\ell_2}^2 + \frac{1}{2}(\mu \bm y_{t+1} - \mu\x_{t+1}^\top\theta)^2 +  t\lambda\|\theta\|_{\ell_1},
\end{equation}
in order to let the problem of learning from $t$ samples evolve to that of learning from $t+1$ samples, as $\mu$ goes from $0$ to $1$. Therefore, representing the Lasso solution as a continuous function of $\mu$ permits the development of homotopy algorithm, which computes the path $\bm\theta(t, \lambda)$ to $\bm\theta(t+1, \frac{t}{t+1}\lambda)$.

This homotopy algorithm is derived based on the fact that, the term of new sample will only result in a rank-1 update in the covariance matrix as $\mb{X}^\top\mb{X} + \mu^2\x_{t+1}\x_{t+1}^\top$, because only $1$ response variable is present. 
Thus, the corresponding matrix inverse in the closed form of optimal solution can be still expressed as an explicit function of $\mu$ using the Sherman Morrison formula, which furthermore allows the calculation of critical points of $\mu$.
However, for the matrix-variate Lasso (\ref{eq: matLasso}), a new sample will cause a rank-$NF$ update in $\Ga$, 
that is the number of response variables in the Lasso problem.
To formally understand this change, we rewrite $\Ga$ as the sum of $t$ reorganized samples analogous to usual $\spc$:
$$
\Ga = \frac{1}{t}\sum_{\tau=1}^t\widetilde{\mb{X}}_{\tau-1}\widetilde{\mb{X}}_{\tau-1}^\top, \mbox{ where } \widetilde{\mb{X}}_{\tau-1} \in \R^{|K|\times NF} \mbox{ with } [\widetilde{\mb{X}}_{\tau-1}]_{k,i} = [U_k]_{i,:}\x_{\tau-1}.
$$
Note that a new $\x_{t+1}$ corresponds to the change $\widetilde{\mb{X}}_{t}\widetilde{\mb{X}}_{t}^\top$ in $\Ga$, which is a rank $NF$ matrix. 
Thus it is impossible to express $\left[\Ga^1\right]^{-1}$ as an explicit and simple function of one single $\mu$. However, note that each column (rank) $[\widetilde{\mb{X}}_t]_{:,i}$ corresponds to introducing new sample of one response variable ${\bm x}_{t+1,i}:= [\x_{t+1}]_i$ at node $i$ in $\G$, by rewriting the incremental term of Lasso \eqref{eq: matLasso}:
\begin{equation}
\begin{aligned}
 \|\x_{t+1} - A\x_{t}\|_{\ell_2}^2 &= \left\|\x_{t+1} - \sum_{k \in K} \langle U_{k},A^0\rangle \frac{1}{\|U_k\|_\mathbf{F}^2}U_{k} \x_{t}\right\|_{\ell_2}^2 \\
  &= \sum\limits_{i = 1}^{NF} \left({\bm x}_{t+1,i} - \sum_{k \in K} \langle U_{k},A^0\rangle \frac{1}{\|U_k\|_\mathbf{F}^2}[U_k]_{i,:} \x_{t}\right)^2. 
\end{aligned}
\end{equation}
Therefore, we propose to introduce $NF$ continuous variables $\mu_1, ..., \mu_{NF}$ in Lasso \eqref{eq: matLasso}, and to consider the following problem
$$\A_{\lambda,t}(\mu_1, ..., \mu_{NF}) = \argmin_{A \in \KG} L_{\lambda,t}(\mu_1, ..., \mu_{NF}), $$
\begin{equation}\label{eq: matLassohomo}
\begin{aligned}
\mbox{ where } \; L_{\lambda,t}(\mu_1, ..., \mu_{NF}) & = \frac{1}{2(t+1)} \sum\limits_{\tau = 1}^t \|\x_{\tau} - A\x_{\tau - 1}\|_{\ell_2}^2 + \lambda F \|\AN\|_{\ell_1} \\
& + \frac{1}{2(t+1)} \sum\limits_{i = 1}^{NF} \mu_i \left({\bm x}_{t+1,i} - \sum_{k \in K} \langle U_{k},A^0\rangle \frac{1}{\|U_k\|_\mathbf{F}^2}[U_k]_{i,:} \x_{t}\right)^2.
\end{aligned}
\end{equation}
Given solution $\A(t, \lambda)$, we first apply the homotopy Algorithm of Section \ref{sec: homo1} on it with $\lambda_1 = \lambda$ and $\lambda_2 = \frac{t+1}{t}\lambda$ to change the constant before the old data term from $\frac{1}{t}$ to $\frac{1}{t+1}$.
Then, we have $\A(t, \frac{t+1}{t}\lambda) = \A_{\lambda,t}(0, ..., 0)$ and $\A(t+1, \lambda) = \A_{\lambda,t}(1, ..., 1)$. We let evolve the optimization problem (\ref{eq: matLasso}) from time $t$ to $t+1$ by sequentially varying all $\mu_i$ from $0$ to $1$, along the paths 
$$L_{\lambda,t}(0, 0, ..., 0) \rightarrow L_{\lambda,t}(1, 0, ..., 0) \rightarrow L_{\lambda,t}(1, 1, ..., 1) = L_{\lambda, t+1}.$$ 
\begin{prop}\label{prop: opconhomo} A minimizer $\A_{\lambda,t}(..., 1, \mu_i, 0, ...)$ of $\min_{A \in \KG}L_{\lambda,t}(..., 1, \mu_i, 0, ...)$ 
satisfies the linear system
\begin{equation}\label{eq: opconhomo}
\Bigg\{
\begin{aligned}
& \Ga^1(\mu_i) \as - \gaone^1(\mu_i) + (1 + \frac{1}{t})\lambda \mb{w}_1 = 0, \\
& \Ga^0(\mu_i) \as - \gaone^0(\mu_i) + (1 + \frac{1}{t})\lambda \mb{w}_0 = 0,
\end{aligned}   
\end{equation}
where $\mb{a}^s, \KN^0, \KN^1, K^1, \mb{w}$ are with respect to $\A = \A_{\lambda,t}(..., 1, \mu_i, 0, ...)$, defining furthermore the extractions through \eqref{eq: extraction}, 
\begin{equation}
\Ga(\mu_i) = \Ga + \frac{1}{t}\sum_{n = 1}^{i-1}[\widetilde{\mb{X}}_t]_{:,n} [\widetilde{\mb{X}}_t]_{:,n}^\top +  \frac{\mu_i}{t}[\widetilde{\mb{X}}_t]_{:,i}[\widetilde{\mb{X}}_t]_{:,i}^\top \, ,
\end{equation}
and
\begin{equation}
\gaone(\mu_i) = \gaone + \frac{1}{t}\sum_{n = 1}^{i-1}{\bm x}_{t+1,n}[\widetilde{\mb{X}}_t]_{:,n} +  \frac{\mu_i}{t}{\bm x}_{t+1,i}[\widetilde{\mb{X}}_t]_{:,i} \, ,
\end{equation}
with $\Ga, \gaone$ are the same ones as in Proposition \ref{prop: opcon}.
\end{prop}

The optimal conditions given in Proposition \ref{prop: opconhomo} show that, each path only relates to the one rank change: $\frac{\mu_i}{t} [\widetilde{\mb{X}}_t]_{:,i}[\widetilde{\mb{X}}_t]_{:,i}^\top$, for the latest updated $\Ga$. Thus the Sherman Morrison formula on $\left[\Ga^1(\mu_i)\right]^{-1}$ \textcolor{black}{can be applied} to retrieve the smooth function of $\mu_i$, and express $\as$ and $\mb{w}_0$ as smooth functions of $\mu_i$, which furthermore makes the calculation of the critical points of $\mu_i$ explicit.  
To leverage these continuity properties, \textcolor{black}{it should be still assumed that} $\left[\mb{a}^s\right]_{\KN^1} \neq 0$ (element-wise), and $|\left[\mb{w}\right]_{\KN^0}| < 1$ (element-wise). For the algorithm of path $\A_{\lambda,t}(0, ..., 0)$ to $\A_{\lambda,t}(1, ..., 1)$, it is sufficient to impose such assumption only on $\A_{\lambda,t}(0, ..., 0)$. By arguing as in Section \ref{sec: homo1}, we can derive the homotopy algorithm for the whole data path. For details, see Algorithm \ref{alg: HomoAlgo2} in the appendices. 

\subsubsection{Update from $\lambda_t$ to $\lambda_{t+1}$}\label{sec: lambda selection}
Given the previous solution $\A(t, \lambda_t)$, one way to select the hyperparameter value $\lambda$ 
is to introduce the empirical objective function \citep{monti2018adaptive, garrigues2008homotopy}, which takes the form 
\begin{equation}\label{eq: pred_err}
    f_{t+1}(\lambda) = \frac{1}{2}\|\x_{t+1} - \A(t, \lambda)\x_{t}\|_{\ell_2}^2, 
\end{equation}
and to employ the updating rule
\begin{equation}\label{eq: lambda_update}
    \lambda_{t+1} = \lambda_t - \eta \frac{\mbox{d}f_{t+1}(\lambda)}{\mbox{d}\lambda}\big|_{\lambda = \lambda_t},
\end{equation}
where $\eta$ is the step size. 

For convenience, \textcolor{black}{the expression} $\frac{\mbox{d}f_{t+1}(\lambda)}{\mbox{d}\lambda}\big|_{\lambda = \lambda_t}$ \textcolor{black}{is written as $\frac{\mbox{d}f_{t+1}(\lambda_t)}{\mbox{d}\lambda}$. Analogously, the notation $\frac{\mbox{d}\A(t, \lambda_t)}{\mbox{d}\lambda}$ is adopted} to denote the derivative with respect to $\lambda$, taken at the value $\lambda = \lambda_t$. The objective function
$f_{t+1}$ can be interpreted as an one step prediction error on unseen data. 
Since the Lasso solution is piece-wise linear with respect to $\lambda$, it follows that when $\lambda$ is not a critical point, the derivative can be calculated as
\begin{equation}
\begin{aligned}
 \frac{\mbox{d}f_{t+1}(\lambda_t)}{\mbox{d}\lambda} & = \left\langle \mb{G}_t, \frac{\mbox{d}\A(t, \lambda_t)}{\mbox{d}\lambda}\right\rangle  \\
 & = \left\langle \projKG(\mb{G}_t), \frac{\mbox{d}\A(t, \lambda_t)}{\mbox{d}\lambda}\right\rangle = - \left[\mb{a}^{\mb{G}_t}_1\right]^\top \left[\Ga^1\right]^{-1}\mb{w}_1,
\end{aligned}
\end{equation}
where $\mb{a}^{\mb{G}_t}_1 \in \R^{|K^1|}$ is defined as  
$\left(\mb{a}^{\mb{G}_t}_1\right)_{i} = \langle U_{k}, \mb{G}_t\rangle, \, k_i \in K^1$, with $K^1$, $\mb{w}_1$, $\left[\Ga^1\right]^{-1}$ associated with $\A(t, \lambda_t)$, and 
$$\mb{G}_t = \left(\A(t, \lambda_t)\x_t - \x_{t+1}\right)\x_t^\top.$$
The derivatives of the entries of $\A(t, \lambda)$ indexed by $K^1$ at $\lambda_t$ can be calculated through the formula \eqref{eq: closed form} of $\as$. By contrast, the derivatives of the entries of $\A(t, \lambda)$ indexed by $\KN^0$ all equal zero. To obtain the non-negative parameter value, we project $\lambda_{t+1}$ onto interval $[0, +\infty)$ by taking $\max\{\lambda_{t+1}, 0\}$, whenever the result from Equation (\ref{eq: lambda_update}) is negative. 

Note that $\lambda_{t+1}$ defined in Equation (\ref{eq: lambda_update}) can be interpreted as the online solution from the projected stochastic gradient descent derived for the batch problem
\begin{equation}\label{eq: batch problem lambda}
 \lambda^*_n = \argmin\limits_{\lambda \geq 0} \frac{1}{2n} \sum\limits_{t = 1}^n \|\x_{t+1} - \A(t, \lambda)\x_{t}\|_{\ell_2}^2.   
\end{equation}
Therefore, the sublinear regret property of projected stochastic gradient descent implies that, when $\eta$ is given as $\mathcal{O}(\frac{1}{\sqrt{n}})$, we have
\begin{equation}\label{eq: sublinear regret}
   \frac{1}{2n} \sum\limits_{t = 1}^n \|\x_{t+1} - \A(t, \lambda_t)\x_{t}\|_{\ell_2}^2 - \frac{1}{2n} \sum\limits_{t = 1}^n \|\x_{t+1} - \A(t, \lambda^*_n)\x_{t}\|_{\ell_2}^2 = \mathcal{O}(\frac{1}{\sqrt{n}}). 
\end{equation}
Equation (\ref{eq: sublinear regret}) implies that in the sense of average one step prediction error defined as Equation (\ref{eq: batch problem lambda}), the adaptive hyperparameter sequence $\{\lambda_t\}_t$ will perform almost as well as the best parameter $\lambda^*_n$, for a large number of online updates, with sufficiently small step size $\eta$. 
This completes the online procedure in the high dimensional domain, which we conclude in Algorithm \ref{alg: online matrix Lasso}.

\begin{algorithm}[t]
   \caption{Online Structured matrix-variate  Lasso}
   \label{alg: online matrix Lasso}
\begin{algorithmic}
   \STATE {\bfseries Input:} $\A(t, \lambda_t)$, $\Ga$, $\gaone$, $\KN^1$(ordered list), $\wNone$, $\lambda_t$, $\left[\Ga^1\right]^{-1}$, $\x_{t+1}$, $\widetilde{\mb{X}}_{t}$, $ t$, where $\KN^1$, $\wNone$, $\left[\Ga^1\right]^{-1}$ are associated with $\A(t, \lambda_t)$, and $\wNone = [\mb{w}]_{\KN^1}$.
   \STATE Select $\lambda_{t+1}$ according to Section \ref{sec: lambda selection}.
   \STATE Update $\A(t, \lambda_t) \rightarrow \A(t, \frac{t+1}{t}\lambda_{t+1})$ using Algorithm \ref{alg: HomoAlgo1}.
   \STATE Update $\A(t, \frac{t+1}{t}\lambda_{t+1}) \rightarrow \A(t+1, \lambda_{t+1})$ using Algorithm \ref{alg: HomoAlgo2}.
   \STATE {\bfseries Output:} $\A(t+1, \lambda_{t+1})$, $\Ga$, $\gaone$, $\KN^1$, $\wNone$, $\lambda_{t+1}$, $\left[\Ga^1\right]^{-1}$.
\end{algorithmic}
\end{algorithm}
\begin{figure}[t]
\begin{center}
\centerline{\includegraphics[width=0.7\textwidth]{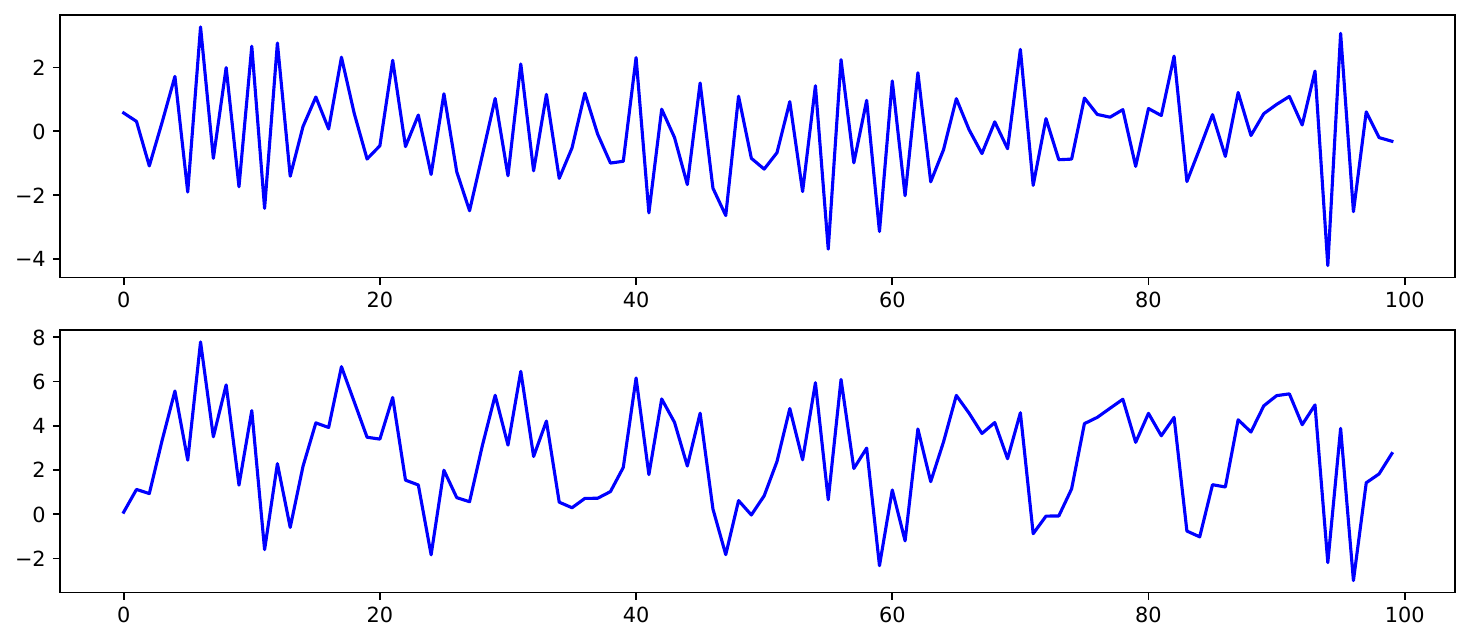}}
\caption{Top is the stationary time series from Model \eqref{eq: var1_matrix} at one component, bottom is the time series from the augmented Model \eqref{eq: var1_vect_aug} in the same realisation.}
\label{fig: added_trend}
\end{center}
\vskip -0.2in
\end{figure}
\section{Augmented Model for Periodic Trends}\label{sec: online learning aug}
The online methods derived previously are based on the data process \eqref{eq: var1_vect}, which assumes the samples $(\x_{\tau})_{{\tau}\in\mathbb{N}}$ have a time-invariant mean of zero. In this section, \textcolor{black}{a more realistic data model is proposed, which considers trends, and the online methods for stationary data are adapted to this augmented model}.

In the literature of time series analysis, stationarity is very often assumed due to its analytic advantage. However, raw data is usually not stationary, as shown in Figure \ref{fig: vector}. \textcolor{black}{Furthermore, Figure \ref{fig: added_trend} provides a comparison between stationary and non-stationary time series.} 

In offline learning, to fit the models to data, a \textit{detrending} step is typically required. This step involves approximating the trend function using the entire data set and then removing it from the raw data. However, since online learning does not require the presence of all data, such pre-processing steps are not feasible. Therefore, it is necessary to consider the trend as explicit parameters in addition to the graph parameters $\AN, \AF$ in the online model. \textcolor{black}{The following augmented model is proposed:}
\begin{equation}\label{eq: var1_vect_aug}
\begin{cases}
    &\mb{x}_t = \mbox{b}_t^0 + \mb{x}_t^{\prime}, \\
    &\mb{x}^{\prime}_t = A \mb{x}^{\prime}_{t-1} + \mb{z}_t, \; \mbox{with} \; A \in \KG, \; \|A\|_2 < 1, \quad t \in \mathbb{N}^+,
\end{cases}
\end{equation}
where $\mb{x}_t = \vect \left(\mb{X}_t \right)$, $\mbox{b}_t^0 \in \R^{NF}$, $\mb{z}_t \in \R^{NF} \sim \mathrm{IID}\left(0, \Sigma\right)$ with non-singular covariance matrix $\Sigma$
and bounded fourth moments, and $\mb{x}^{\prime}_0 = \sum_{j=0}^\infty A^j\mb{z}_{t-j}$. The observations of Model \eqref{eq: var1_vect_aug} is $\mb{x}_t$, while $\mb{x}_t^{\prime}$ has the similar role as the unobserved state in the state space models however the observation equation here is much simplified. Therefore the estimators are built on the series $\mb{x}_t$. Note that Model \eqref{eq: var1_vect_aug} admits another reparameterization with intercept
\begin{equation}\label{eq: var1_vect_aug_bm}
    \mb{x}_t =  \mbox{b}_t +A\mb{x}_{t-1} + \mb{z}_t, \quad  \mbox{b}_t = \mbox{b}_t^0 -A\mbox{b}_{t-1}^0.
\end{equation}
The augmented model assumes that the non-stationarity of the observations $\mb{x}_t$ is caused by the trend $\mbox{b}_t^0$. We consider in particular in this work, the periodic trend 
\begin{equation}\label{eq: periodic trend}
\mbox{b}_t^0 = \mbox{b}_m^0, \; m = 0, ..., M-1, \; m = t \; \mbox{mod}
\; M,
\end{equation}
where $M$ is the length of period and it is a hyperparameter to be preassigned. This type of trend is frequently encountered in practice. For example, an annual recurrence ($M = 12$) can be found in many monthly data sets recorded over years, such as the weather data in Figure \ref{fig: vector}. In the following sections, we will adapt the two learning frameworks presented in Section \ref{sec: online learning} to the augmented model for the periodic trends, in order to infer the trends and graphs simultaneously from non-stationary time series $\mb{x}_t$, in an online fashion.

\subsection{New OLS Estimators and Asymptotic Distributions}
For the augmented model (\ref{eq: var1_vect_aug}), \textcolor{black}{a new OLS estimator of $A$ is proposed}, based on the new sample auto-covariances, together with the OLS estimator of $\mbox{b}_m^0$. Because two crucial properties for deriving the Wald tests in Section \ref{sec:app1} are the consistency of sample auto-covariances $\spc, \, \spcone$, and the CLT of the OLS estimator $\widecheck{\A}_t$, the corresponding asymptotic results for the new estimators are derived, showing that these asymptotics are exactly the same as in the stationary case. Therefore, all the results and procedures presented in Section \ref{sec:app1} can be applied directly to the new estimators. \textcolor{black}{The estimator of $A$, still denoted as $\widecheck{\A}_t$, is first defined using the general least squares (GLS) method:}
\begin{equation}\label{eq: op new GLS}
    \widecheck{\A}_t, \widehat{\mb{b}}_{m,t} = \argmin_{A, \mbox{b}_{m}} \sum_{m = 0}^{M-1} \bm S_m(A, \mbox{b}_m),
\end{equation}
where 
$$\bm S_m = \sum_{\tau \in I_{m,t}}
\Tilde{\mb{z}}_{\tau}^\top\Sigma^{-1}\Tilde{\mb{z}}_{\tau}, \quad \Tilde{\mb{z}}_{\tau} = \x_{\tau} - \mbox{b}_m -A\x_{\tau - 1},$$
with $I_{m,t} = \{\tau = 1,...,t : \tau \; \mbox{mod} \; M = m\}$, and $\Sigma^{-1}$ the true white noise covariance given in Model (\ref{eq: var1_vect}). Note that $\Tilde{\mb{z}}_{\tau}$ represents the residual of the prediction of sample $\x_{\tau}$. The explicit forms of $\widecheck{\A}_t, \widehat{\mb{b}}_{m,t}$ can be found through straightforward calculation, which yields new sample auto-covariances, denoted still as $\spc, \, \spcone$, and the estimator of trend $\widehat{\mb{b}}_{m,t}^0$. Specifically, we have 
\begin{equation}\label{eq: new OLS}
\Bigg\{
\begin{aligned}
&\widecheck{\A}_t = \spcone \left[\spc\right]^{-1}, \\
& \widehat{\mb{b}}_{m,t} = \Bar{\x}_{m,t} - \widecheck{\A}_t \underline{\x}_{m-1,t} \; \Rightarrow \; \widehat{\mb{b}}_{m,t}^0 = \underline{\x}_{m,t} (\mbox{or}\; \Bar{\x}_{m,t}), 
\end{aligned}
\end{equation}
with
$$
\begin{aligned}
& \spc = \sum_{m=0}^{M-1} \frac{p_{m,t}}{t}\left(\frac{\sum\limits_{\tau \in I_{m,t}}\x_{\tau-1}\x_{\tau-1}^\top}{p_{m,t}} - \underline{\x}_{m-1,t}\underline{\x}_{m-1,t}^\top\right), \\
& \spcone = \sum_{m=0}^{M-1} \frac{p_{m,t}}{t}\left(\frac{\sum\limits_{\tau \in I_{m,t}}\x_{\tau}\x_{\tau-1}^\top}{p_{m,t}} - \Bar{\x}_{m,t}\underline{\x}_{m-1,t}^\top\right), \\
& p_{m,t} = |I_{m,t}|, \; \Bar{\x}_{m,t} = \sum_{\tau \in I_{m,t}} \frac{\x_{\tau}}{p_{m,t}}, \; m = 0, ..., M-1, \\
&\underline{\x}_{m-1,t} = \sum_{\tau \in I_{m,t}} \frac{\x_{\tau-1}}{p_{m,t}}, \; m = 0, ..., M-1.
\end{aligned}
$$
Note that $\underline{\x}_{-1,t}$ denotes $\underline{\x}_{M-1,t}$. It is also straightforward to understand the new auto-covariance estimators. Each $\bm S_m(A, \mbox{b}_m)$ leads to an OLS problem of regression equation \eqref{eq: var1_vect_aug_bm}. Its minimization introduces two sample covariance matrices. The weighted average of all such sample auto-covariance matrices for $m = 0, \ldots, M-1$ is the new sample auto-covariance for Model \eqref{eq: var1_vect_aug}. $p_{m,t}$ denotes the number of times that the samples from the $m$-th state point in the period have been predicted in the sense of Equation (\ref{eq: op new GLS}). As $t$ grows, $\underline{\x}_{m,t}$ becomes $\Bar{\x}_{m,t}$ quickly, and $p_{m,t}$ becomes $\frac{t}{M}$. For the augmented model, GLS and OLS estimators are still identical, with the latter defined as
$$\argmin_{A, \mbox{b}_{m}} \sum_{m = 0}^{M-1} \sum_{\tau \in I_{m,t}}
\Tilde{\mb{z}}_{\tau}^\top\Tilde{\mb{z}}_{\tau}.$$
The estimators given by Formula \eqref{eq: op new GLS} enjoy the asymptotic properties in Proposition \ref{prop: asymptotics new OLS}.

\begin{prop}\label{prop: asymptotics new OLS} The following asymptotic properties hold for the estimators $\spc, \, \spcone, \, \widecheck{\A}_t,$ $\, \widehat{\mb{b}}_{m,t}^0$, as $t \to + \infty$,

\begin{enumerate}
\item $\spc \xrightarrow{p} \Gamma(0)$, $\spcone \xrightarrow{p} \Gamma(1)$,
\item $\widehat{\mb{b}}_{m,t}^0 \xrightarrow{p} \mbox{\normalfont b}_{m}^0$, $\widecheck{\A}_t \xrightarrow{p} A$, 
\item $\sqrt{t} \, \vect(\widecheck{\A}_t - A) \xrightarrow{d} \mathcal{N}(0, \left[\Gamma(0)\right]^{-1} \otimes \Sigma)$,
\end{enumerate}
\textit{where} $\Gamma(h) = \mathbb{E}\left(\mb{x}_t^{\prime}[\mb{x}_{t-h}^{\prime}]^{\top}\right), \, h \geq 0$, $\Sigma = \mathbb{E}\left(\mb{z}_t\mb{z}_{t}^{\top}\right)$.
\end{prop}
The proofs of the above results are given in  \ref{sec: proof aug asymp}. 
Thus, Theorem \ref{thm: asym_AN} and the bisection Wald test procedure are still valid using $\projKN(\widecheck{\A}_t)$ and $\spc, \, \spcone$ defined in this section.
On the other hand, $\spc$ and $\spcone$ satisfy the one rank update formulas
\begin{equation}\label{eq: update autocov}
\begin{aligned}
   &\widehat{\mb{\Gamma}}_{t+1}(0) =  \frac{t}{t+1}\widehat{\mb{\Gamma}}_{t}(0) + \frac{1}{t+1} \left[\frac{p_{\Bar{m},t}}{p_{\Bar{m},t} + 1}(\x_t - \underline{\x}_{\Bar{m}-1,t})(\x_t - \underline{\x}_{\Bar{m}-1,t})^\top\right], \\
   &\widehat{\mb{\Gamma}}_{t+1}(1) =  \frac{t}{t+1}\widehat{\mb{\Gamma}}_{t}(1) + \frac{1}{t+1} \left[\frac{p_{\Bar{m},t}}{p_{\Bar{m},t} + 1}(\x_{t+1} - \underline{\x}_{\Bar{m},t})(\x_t - \underline{\x}_{\Bar{m}-1,t})^\top\right],
\end{aligned}
\end{equation}
where $\Bar{m} = (t+1) \; \mbox{mod} \; M$.
Thus, when new sample comes, $[\widehat{\mb{\Gamma}}_{t+1}(0)]^{-1}$  can still be calculated efficiently given the matrix inverse at the previous time. The details of the extended low dimensional learning procedure see Algorithm \ref{alg: proj_Wald aug} in the appendices.

\subsection{Augmented Structured Matrix-variate Lasso and the Optimality Conditions}\label{sec:app2 aug}
To adapt the Lasso-based approach to Model \eqref{eq: var1_vect_aug_bm}, the corresponding trend and graph estimators can be obtained by minimizing the augmented Matrix-variate Lasso problem 
\begin{equation}\label{eq: matLasso aug}
    \A(t, \lambda), \mb{b}_m(t, \lambda) = \argmin_{A \in \KG, \mbox{b}_m} \frac{1}{2t} \sum\limits_{m = 0}^{M-1}\sum\limits_{\tau \in I_{m,t}} \|\x_{\tau} - \mbox{b}_m -A\x_{\tau - 1}\|_{\ell_2}^2 + \lambda F \|\AN\|_{\ell_1}.
\end{equation}
As in the extension of our first approach, the extra bias terms $\mbox{b}_m, m = 0, ..., M-1,$ do not affect the core techniques, rather they force the methods to consider the $M$ means in the sample autocovariances.
Since $\mbox{b}_m$ only appear in the squares term, the minimizers $\mb{b}_m(t, \lambda)$ have the same dependency with $\A(t, \lambda)$ as in Equation \eqref{eq: new OLS}. Thus the trend $\mbox{b}_m^0$ can still be estimated by $\underline{\x}_{m,t}$, and we extend the algorithms in Section \ref{sec:app2} to update the batch solution of augmented Lasso \eqref{eq: matLasso aug} from $\A(t, \lambda_t)$ to $\A(t+1, \lambda_{t+1})$, given new sample $\mb{x}_{t+1}$. 
To compute the regularization path $\A(t, \lambda_t) \rightarrow \A(t, (1+\frac{1}{t})\lambda_{t+1})$, Proposition \ref{prop: opcon aug} implies that Algorithm \ref{alg: HomoAlgo1} can still be used, with the adjusted definitions of $\Ga$ and $\gaone$.
\begin{prop}\label{prop: opcon aug} A minimizer $\A(t, \lambda)$ of Lasso problem \eqref{eq: matLasso aug} satisfies the linear system
\begin{equation}\label{eq: opcon aug}
\Bigg\{
\begin{aligned}
& \Ga^1 \as - \gaone^1 + \lambda \mb{w}_1 = 0, \\
& \Ga^0 \as - \gaone^0 + \lambda \mb{w}_0 = 0,
\end{aligned}   
\end{equation}
where $\mb{a}^s$ is the vectorized scaled Lasso solution $\A(t, \lambda)$, $\mb{w}, K^1, \KN^0$ are also defined analogously from $\A(t, \lambda)$, while $\spc$ and $\spcone$ used in the definitions of $\Ga$ and $\gaone$ are the new sample auto-covariance matrices in Equation \eqref{eq: new OLS}. 
\end{prop} 

For the data path $\A(t, (1+\frac{1}{t})\lambda_{t+1}) \rightarrow \A(t+1, \lambda_{t+1})$, in the same spirit of Problem (\ref{eq: matLassohomo}), variables $\mu_1, ..., \mu_{NF}$ are introduced to let evolve Lasso problem \eqref{eq: matLasso aug} from time $t$ to $t+1$ through the variational problem \eqref{eq: matLassohomo aug}.

We recall that \textcolor{black}{the solution is updated along the path:} 
$$L_{\lambda_{t+1},t}(0, 0, ..., 0) \rightarrow L_{\lambda_{t+1},t}(1, 0, ..., 0) \rightarrow L_{\lambda_{t+1},t}(1, 1, ..., 1) = L_{\lambda_{t+1}, t+1}.$$ 
At each step $L_{\lambda_{t+1},t}(..., 1, \mu_i, 0, ...), \mu_i \in [0,1]$, the optimal solution $\A_{\lambda_{t+1},t}(..., 1, \mu_i, 0, ...)$ is piece-wise smooth with respect to $\mu_i$, element-wise. The linear system of $\as$ in terms of $\mu_i$ is retrieved in Proposition \ref{prop: opconhomo aug} for the optimality condition of each $\A_{\lambda_{t+1},t}(..., 1, \mu_i, 0, ...)$.

\begin{equation}\label{eq: matLassohomo aug}
\begin{aligned}
   &\A_{\lambda_{t+1},t}(\mu_1, ..., \mu_{NF}), \; \mb{b}_{m,\lambda_{t+1},t}(\mu_1, ..., \mu_{NF}) = \argmin_{A \in \KG, \mbox{b}_{m}} \; L_{\lambda_{t+1},t}(\mu_1, ..., \mu_{NF}), \\
   &\mbox{ where } L_{\lambda_{t+1},t}(\mu_1, ..., \mu_{NF}) = 
   \frac{1}{2(t+1)} \sum\limits_{m = 0}^{M-1}\sum\limits_{\tau \in I_{m,t}} \|\x_{\tau} - \mbox{b}_m -A\x_{\tau - 1}\|_{\ell_2}^2 \\
   & + \lambda_{t+1} F \|\AN\|_{\ell_1} + \frac{1}{2(t+1)} \sum\limits_{i = 1}^{NF} \mu_i ({\bm x}_{t+1,i} - b_{\Bar{m},i} -  \sum_{k \in K} \langle U_{k},A^0\rangle \frac{1}{\|U_k\|_\mathbf{F}^2}[U_k]_{i,:} \x_{t})^2,
\end{aligned}
\end{equation}
where $b_{\Bar{m},i} = [\bM]_i$, $\Bar{m} = (t+1) \; \mbox{mod} \; M$. 

\begin{prop}\label{prop: opconhomo aug}
A minimizer $\A_{\lambda_{t+1},t}(..., 1, \mu_i, 0, ...)$ of Lasso $L_{\lambda_{t+1},t}(..., 1, \mu_i, 0, ...)$ satisfies the linear system
\begin{equation}\label{eq: opconhomo aug}
\Bigg\{
\begin{aligned}
& \Ga^1(\mu_i) \as - \gaone^1(\mu_i) + (1 + \frac{1}{t})\lambda_{t+1} \mb{w}_1 = 0, \\
& \Ga^0(\mu_i) \as - \gaone^0(\mu_i) + (1 + \frac{1}{t})\lambda_{t+1} \mb{w}_0 = 0, 
\end{aligned}  
\end{equation}
where $\mb{a}^s, \mb{w}, K^1, \KN^0$ are with respect to $\A_{\lambda_{t+1},t}(..., 1, \mu_i, 0, ...)$, defining the extractions through \eqref{eq: extraction},
$$
\begin{aligned}
\Ga(\mu_i) = \Ga &+ \frac{1}{t}\sum_{n = 1}^{i-1}\frac{p_{\Bar{m},t}}{p_{\Bar{m},t} + 1}[\widetilde{\mb{X}}_t - \underline{\widetilde{\mb{X}}}_{\Bar{m}-1,t}]_{:,n} [\widetilde{\mb{X}}_t - \underline{\widetilde{\mb{X}}}_{\Bar{m}-1,t}]_{:,n}^\top \\
&+ \frac{\mu_i}{t}\frac{p_{\Bar{m},t}}{p_{\Bar{m},t} + \mu_i}[\widetilde{\mb{X}}_t - \underline{\widetilde{\mb{X}}}_{\Bar{m}-1,t}]_{:,i}[\widetilde{\mb{X}}_t - \underline{\widetilde{\mb{X}}}_{\Bar{m}-1,t}]_{:,i}^\top,
\end{aligned}
$$ 
$$
\begin{aligned}
\gaone(\mu_i) = \gaone &+ \frac{1}{t}\sum_{n = 1}^{i-1}\frac{p_{\Bar{m},t}}{p_{\Bar{m},t} + 1}(\x_{t+1,n}-(\underline{\x}_{\Bar{m},t})_n)[\widetilde{\mb{X}}_t - \underline{\widetilde{\mb{X}}}_{\Bar{m}-1,t}]_{:,n} \\
&+ \frac{\mu_i}{t}\frac{p_{\Bar{m},t}}{p_{\Bar{m},t} + \mu_i}(\x_{t+1,i}-(\underline{\x}_{\Bar{m},t})_i)[\widetilde{\mb{X}}_t - \underline{\widetilde{\mb{X}}}_{\Bar{m}-1,t}]_{:,i},
\end{aligned}
$$
with $[\underline{\widetilde{\mb{X}}}_{\Bar{m}-1,t}]_{k,i} = [U_k]_{i,:}\underline{\x}_{\Bar{m}-1,t}$, $p_{\Bar{m},t} := t \; \mbox{mod} \; M$, $\Ga, \gaone$ the same as Proposition \ref{prop: opcon aug}.
\end{prop}

Therefore, the derived homotopy algorithm is essentially the previous homotopy Algorithm \ref{alg: HomoAlgo2} with minor changes. For details, see Algorithm \ref{alg: online matrix Lasso aug} in the appendices.

Lastly, \textcolor{black}{the updating rule for the regularization parameter is derived. The one-step prediction error is still considered, which is expressed as} the following objective function in the case of Model \eqref{eq: var1_vect_aug_bm}:
\begin{equation}
f_t(\lambda) = \frac{1}{2}\|\x_{t+1} - \mb{b}_{\Bar{m}}(t,\lambda) - \A(t, \lambda)\x_{t}\|_{\ell_2}^2.
\end{equation}
Given the previous solution $\A(t, \lambda_t)$ and $\mb{b}_{\Bar{m}}(t,\lambda_t)$, we assume that $\lambda_t$ is not a critical point. Then the derivative of $f_t$ with respect to $\lambda$ is calculated as
\begin{equation}
\begin{aligned}
 \frac{\mbox{d}f_t(\lambda_t)}{\mbox{d}\lambda} & = \left\langle \frac{\mbox{d}f_t(\lambda)}{\mbox{d}\mb{b}_{\Bar{m}}(t,\lambda)}\Big|_{\lambda = \lambda_t}, \frac{\mbox{d}\mb{b}_{\Bar{m}}(t,\lambda_t)}{\mbox{d}\lambda}\right\rangle + \left\langle \frac{\mbox{d}f_t(\lambda)}{\mbox{d}\A(t,\lambda)}\Big|_{\lambda = \lambda_t}, \frac{\mbox{d}\A(t, \lambda_t)}{\mbox{d}\lambda}\right\rangle \\
 &= \left\langle \mb{G}_t^{\mb{b}}, -\frac{\mbox{d}\A(t, \lambda_t)}{\mbox{d}\lambda}\underline{\x}_{\Bar{m}-1,t}\right\rangle + \left\langle \mb{G}_t, \frac{\mbox{d}\A(t, \lambda_t)}{\mbox{d}\lambda}\right\rangle \\
 &= \left\langle [\A(t, \lambda_t)\x_t - \x_{t+1} + \mb{b}_{\Bar{m}}(t,\lambda_t)][\x_t - \underline{\x}_{\Bar{m}-1,t}]^\top, \frac{\mbox{d}\A(t, \lambda_t)}{\mbox{d}\lambda}\right\rangle,
\end{aligned}
\end{equation}
where $\mb{G}_t^{\mb{b}} = \mb{b}_{\Bar{m}}(t,\lambda_t) - \x_{t+1} + \A(t,\lambda_t)\x_t$, $\mb{G}_t = [\A(t, \lambda_t)\x_t - \x_{t+1} + \mb{b}_{\Bar{m}}(t,\lambda_t)]\x_t^\top$. Analogous to Section \ref{sec: lambda selection}, we have $\langle \mb{G}_t, \frac{\mbox{d}\A(t, \lambda_t)}{\mbox{d}\lambda}\rangle = -\left[\mb{a}^{\mb{G}_t}_1\right]^\top \left[\Ga^1\right]^{-1}\mb{w}_1$. Using the same updating rules of the projected stochastic gradient descent presented in Section \ref{sec: lambda selection}, we can compute the online solution $\lambda_{t+1}$. 
We can see that, the introduction of bias terms $\mbox{b}_m$ into the original model makes them center the raw data automatically during the model fitting. This enables the direct learning over raw time series, while maintaining the performance of methods  comparable to the stationarity-based ones.

We summarize the complete learning procedure of this subsection in Algorithm \ref{alg: online matrix Lasso aug} in the appendices.

\section{Experiments}\label{sec: experiments}
We test the two proposed approaches for the online graph and trend learning on both synthetic and real data sets. 

\subsection{Synthetic Data}
\subsubsection{Evaluation Procedures}\label{eq: testing procedure}
\textcolor{black}{The evaluation procedure for the augmented model approaches is now presented.} 
In each simulation, a true graph $A$ is generated with the structure indicated by $\KG$. In particular, sparsity is imposed on its spatial graph $\AN$ by randomly linking a subset of node pairs. The values of the non-zero entries in $\AN$, as well as the entries in $\AF$ and $\mbox{diag}(A)$, are generated randomly.
Additionally,  {\CM a trend over a period of $M$ time points is generated} for each node and each feature. Therefore, the true $\mbox{b}^0_m, \; m = 0, ..., M - 1$, consists in these $NF$ trend vectors, each containing $M$ elements. Then, observations $\x_t$ are generated from Model \eqref{eq: var1_vect_aug}, where the noise follows a centered multivariate normal distribution with a randomly generated covariance. Figure \ref{fig: added_trend} shows an example of the synthetic time series $\x_t$, compared with its stationary source $\x^{\prime}_t$ before adding the periodic trend. The graph and trend estimators proposed in Section \ref{sec: online learning aug} only use $\x_t$. 

To start the two online procedures, \textcolor{black}{a very small number of samples is first synthesized} until time $t_0$. The batch Lasso problem \eqref{eq: matLasso aug} is set up with the generated samples, and its solution $\A(t_0, \lambda_0)$ is used to initialize the high-dimensional online procedure starting with the next synthetic sample. The batch problem is solved via the accelerated proximal gradient descent with backtracking line search \citep[Section 3.2.2]{parikh2014proximal}. \textcolor{black}{In particular, $\lambda_0$ is set as a large number to obtain an over-sparse initial solution. Therefore, a decreasing $\lambda_t$ is expected, along with a more accurate estimate $\mathbf{A}(t, \lambda_t)$ as $t$ increases.} For the low-dimensional procedure, since it requires $[\hat{\mathbf{\Gamma}}_{t}(0)]^{-1}$, the start is delayed until there are enough samples, which is simply set as $t = t_0 + \text{dim}(\KG)$.

To analyse the performance of the proposed online estimators, the following error metrics are defined. The average one step prediction error metric is given as
\begin{equation}\label{eq: pred_err_aug}
\sum\limits_{\tau=1}^t\frac{\| (\mbox{b}_{m(\tau+1)} + A\x_{\tau}) - (\widehat{\mathbf{b}}_{m(\tau+1), \tau} + \A^e_{\tau}\x_{\tau})\|_2}{t\|\mbox{b}_{m(\tau+1)} + A\x_{\tau}\|_2}, \quad m(\tau+1) = (\tau + 1) \; \mbox{mod} \; M, 
\end{equation}
and root mean square deviation (RMSD) as
\begin{equation}\label{eq: rmsd}
\frac{\|\A^e_t - A\|_{\mb{F}}}{\|A\|_{\mb{F}}},  
\end{equation}
where $\A^e_{\tau}$ denotes the estimates from either approach at time $\tau$, and 
\begin{equation}
\widehat{\mb{b}}_{m(\tau+1),\tau} =
\begin{cases}
\underline{\x}_{m(\tau+1),\tau} - 
\widehat{\A}_{\tau} 
\underline{\x}_{m(\tau+1)-1,\tau}, \quad & \mbox{in low-dimensional,} \\
\underline{\x}_{m(\tau+1),\tau} - \A(\tau, \lambda_{\tau}) \underline{\x}_{m(\tau+1)-1,\tau}, \quad & \mbox{in high-dimensional}.
\end{cases}
\end{equation}
\begin{figure}[ht]
\begin{center}
\centerline{\includegraphics[width=1.0\textwidth]{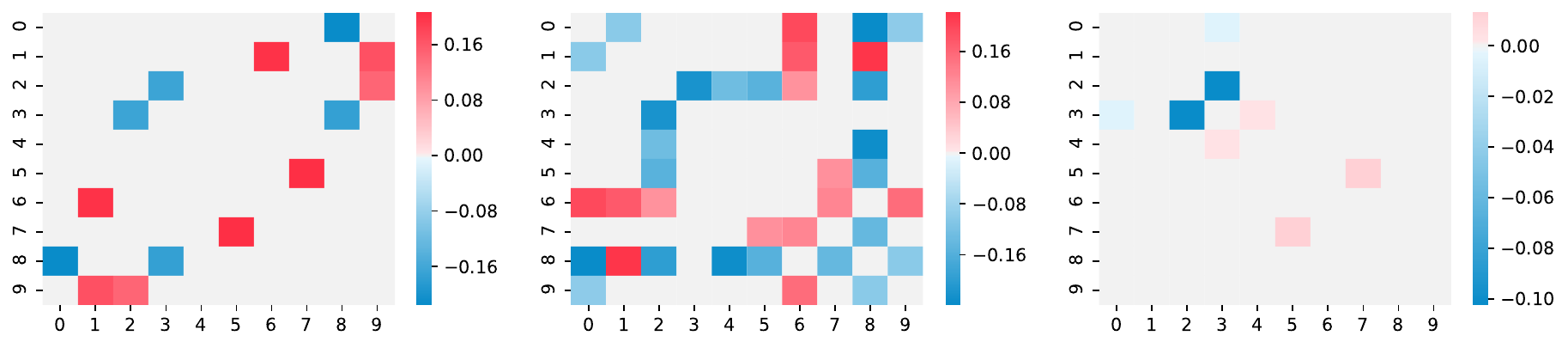}}
\caption{\textit{Initial spatial graph estimates which start the online procedures.} True $\AN$ (left), $\widehat{\bm \AN}_{,91}$ of the low-dimensional  procedure (middle), and $\A_{\N}(20, 0.05)$ of the high-dimensional procedure (right) are represented by heatmaps. Simulation settings: $N = 10$, $F = 4$, number of model parameters = $571$, significance level of $\chi^2$ test in Corollary \ref{coro: Wald test} $= 0.1$. }
\label{fig: Est_init_aug}
\end{center}
\vskip -0.2in
\end{figure}
\begin{figure}[ht]
\begin{center}
\centerline{\includegraphics[width=1.0\textwidth]{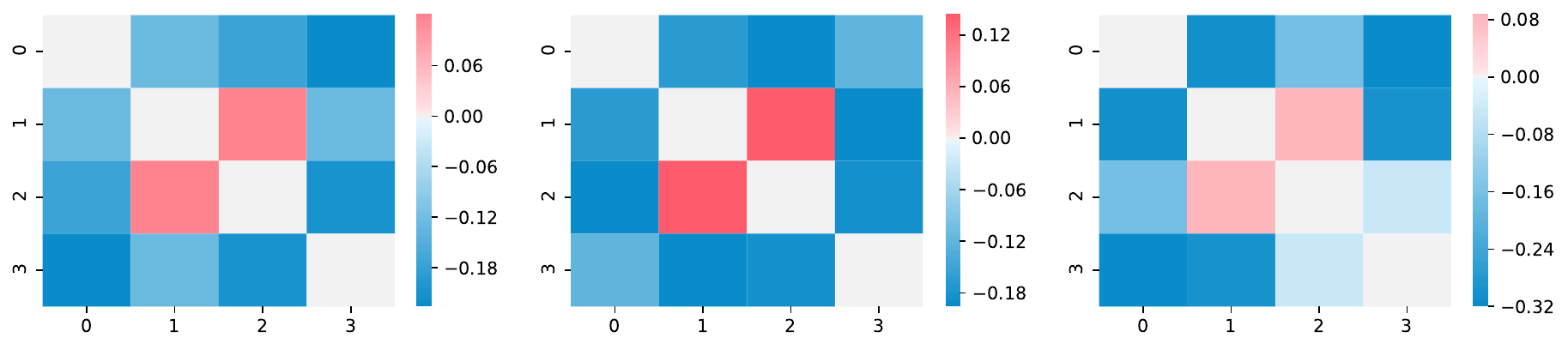}}
\caption{\textit{Initial feature graph estimates which start the online procedures.} True $\AF$ (left), $\widehat{\bm \AF}_{,91}$ of the low-dimensional  procedure (middle), and $\A_{\mathrm{F}}(20, 0.05)$ of the high-dimensional procedure (right). Simulation settings: $N = 10$, $F = 4$, number of model parameters = $571$.}
\label{fig: Est_init_AF_aug}
\end{center}
\vskip -0.2in
\end{figure}

Note that, the prediction error is defined with the conditional expectation (i.e. the oracle best prediction) instead of the new observation $\x_{\tau+1}$. {\CM Indeed, it is} observed in experiments that due to the presence of the random noise, the oracle prediction can be less optimal than other empirical predictions. To avoid this misleading by the noise, the proposed predictions {\CM are compared} with the oracle best prediction to evaluate their performance. On the top of it,  the direct prediction error at each step {\CM is not used} in the metric, because the corresponding metric curves are shown very noisy in practice. To facilitate the graphical presentation, we take additionally a moving average, as shown in Equation \eqref{eq: pred_err_aug}.

Given the error metrics, their values are collected over time. \textcolor{black}{Such simulation is performed multiple times} to obtain the means and standard deviations of the error metrics at each iteration (when the estimators are available) to better demonstrate performance. The true graph $A$ and trends $\mbox{b}_m^0$ are generated independently for each simulation.

To furthermore justify the proposed models, the MAR models of \cite{chen2021autoregressive} and the classical VAR(1) models {\CM are additionally implemented  } in the simulations above. Especially, \cite{chen2021autoregressive} proposed $3$ estimators for their KP-based MAR models, which we will all test. For the VAR(1) models, we use the OLS estimation. To make the comparison fair, another group of simulations {\CM is performed} where samples are generated from the MAR models of \cite{chen2021autoregressive}. To be more specific, in each simulation of the second group,  the KP component $A_c \in \R^{N \times N}$ and $A_r \in \R^{F \times F}$ {\CM are first generated} randomly. $A_c$ and $A_r$ are imposed no sparsity nor symmetricity. On the other hand, we randomly generate $NF$ trend vectors, all over a period of $M$ time points. Then, latent stationary samples {\CM are synthesized } according to Model \eqref{eq: chen mar vat}, where the noise follows a centered multivariate normal distribution with a randomly generated covariance. Afterwards,the periodic trends {\CM  are added } to the stationary samples to generate the synthetic observations. Therefore for this group of simulations, the coefficient matrix $A$ in the prediction error metric \eqref{eq: pred_err_aug} should be $A_c \otimes A_r$, and the intercepts b$_{m(\tau +1)}$ should be modified accordingly. The true model parameters are generated independently across the simulations. 

For the second group, our online procedures {\CM are started } the same way and at the same time as before. The procedures are applied directly on the observed time series, even though they are generated from the model of \cite{chen2021autoregressive}. We present how to start the competitor estimators in both groups. On one hand, because they are designed for stationary time series and are all offline methods, {\CM it is needed} to calculate them anew at each time step and with a pre-processing detrending step. The detrending step {\CM that is adopted} is the same one used in the numerical experiments of \cite[Section 5.2]{chen2021autoregressive}. The estimation of (periodic) trends in the detrending step is used therefore to calculate the prediction error metric for the competitor methods by replacing $\underline{\x}_{m(\tau+1),\tau}$ and $\underline{\x}_{m(\tau+1) - 1,\tau}$. On the other hand, the competitor estimators do not have extra regularization as the Lasso one in our case, so they are not as immediate as our high-dimensional estimator, which means that they are not calculable or the corresponding algorithms can not converge. Thus for the OLS estimator of VAR(1) model and the projection (Proj) estimator of \cite{chen2021autoregressive}, we start calculating them from the same time as our low-dimensional estimator. For the iterated least squares (MLE) and the structured maximum likelihood (MLEs) estimators of \cite{chen2021autoregressive}, we check at each step and record them whenever they are available.

\subsubsection{Simulation Results}\label{sec: simu results}
To facilitate the remarks, this section {\CM is organized } in two parts. In the first part,  the performance of our methods {\CM are reported} without comparing to the competitors and focus on its interpretation. In the second part, our methods {\CM  are examined} in comparison with other methods. 

In the following,  the representative estimates {\CM are visualized in} heatmap with $N = 10$, $F = 4$, and $M = 12$ for illustration purpose. Then, the evolution of error metrics and regularization parameter of $30$ simulations  {\CM are plotted}  for $N = 20$, $F = 5$, and $M = 12$. Lastly, the running time {\CM is reported}.
The hyperparameter settings are given in the captions of figures of corresponding results. 

\begin{figure}[!ht]
\begin{center}
\centerline{\includegraphics[width=1.0\textwidth]{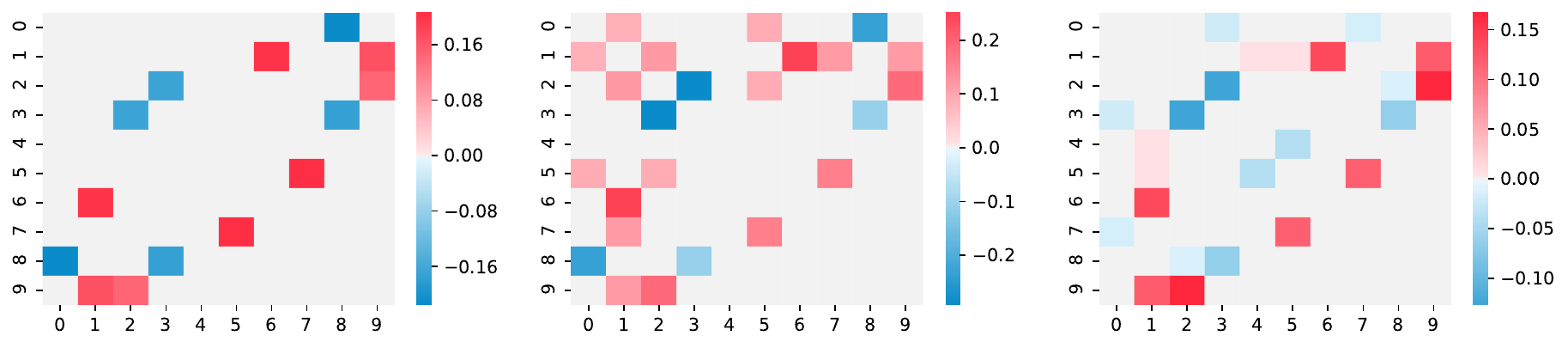}}
\caption{\textit{Spatial graph estimated at the arrival of the $182$-th sample.} True $\AN$ (left), $\widehat{\bm \AN}_{,182}$ of the low-dimensional  procedure (middle), and $\A_{\N}(182, 0.0286)$ of the high-dimensional procedure (right) are represented by heatmaps. Simulation settings: $N = 10$, $F = 4$, number of model parameters = $571$, $\mbox{significance level of } \chi^2 \mbox{ test} = 0.1$, $\eta = 5\times10^{-6}$. }
\label{fig: Est_2d_aug}
\end{center}
\vskip -0.2in
\end{figure}

\begin{figure}[!ht]
\begin{center}
\centerline{\includegraphics[width=1.0\textwidth]{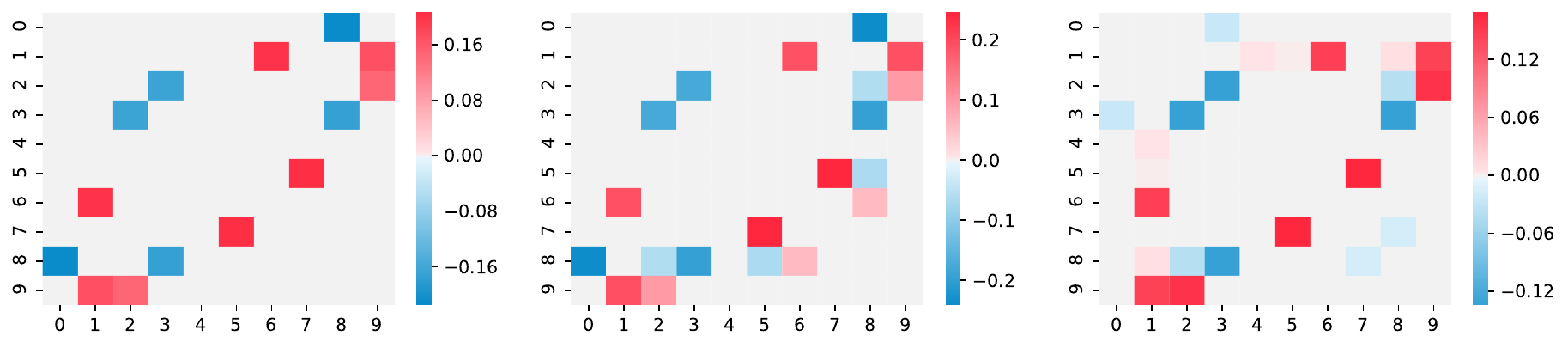}}
\caption{\textit{Spatial graph estimated at the arrival of the $591$-th sample.} True $\AN$ (left), $\widehat{\bm \AN}_{,591}$ of the low-dimensional  procedure (middle), and $\A_{\N}(591, 0.0130)$ of the high-dimensional procedure (right) are represented by heatmaps. Simulation settings: $N = 10$, $F = 4$, number of model parameters = $571$, $\mbox{significance level of } \chi^2 \mbox{ test} = 0.1$, $\eta = 5\times10^{-6}$. }
\label{fig: Est_last_aug}
\end{center}
\vskip -0.2in
\end{figure}

\begin{figure}[ht]
\begin{center}
\centerline{\includegraphics[width=1.0\textwidth]{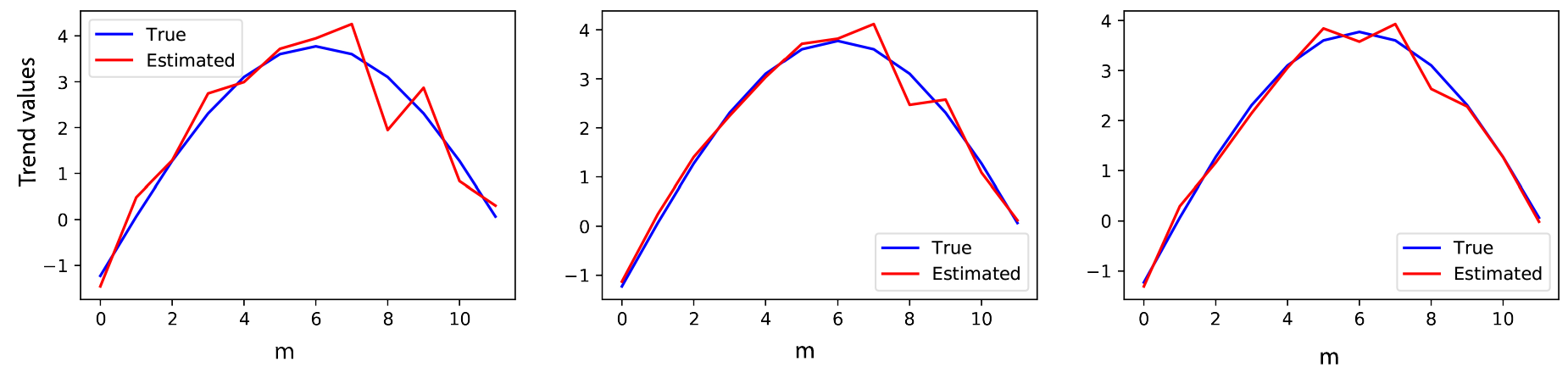}}
\caption{\textit{Trend of the first node, first feature, estimated at different times.} Estimation at $t = 182$ (left), $t = 273$ (middle), $t = 591$ (right). Simulation settings: $N = 10$, $F = 4$, $M = 12$, number of model parameters = $571$. }
\label{fig: Est_trend}
\end{center}
\vskip -0.2in
\end{figure}

Figures \ref{fig: Est_init_aug} and \ref{fig: Est_init_AF_aug} show the estimated graphs of two approaches when their corresponding online procedures start. In Figure \ref{fig: Est_init_aug}, {\CM it can be seen} that the batch solution which starts the high-dimensional procedure is over sparse due to the large $\lambda_0$. We can notice from Figure \ref{fig: Est_init_AF_aug} that the two initial estimations of $\AF$ are already satisfactory, especially the Lasso solution which uses only $20$ samples. Actually, estimations of $\AF$ and $\mbox{diag}(A)$ converge to the truth very quickly in both cases when $N$ is significantly larger than $F$. Figures \ref{fig: Est_2d_aug} and \ref{fig: Est_last_aug} show that the estimations of $\AN$ of both approaches tend to the true values as more samples are received. Meanwhile, Figure \ref{fig: Est_trend} shows the effectiveness of trend estimator $\underline{\x}_{m,t}$ defined in Equation \eqref{eq: new OLS}.

\begin{figure}[!ht]
\begin{center}
\centerline{\includegraphics[width=0.65\textwidth]{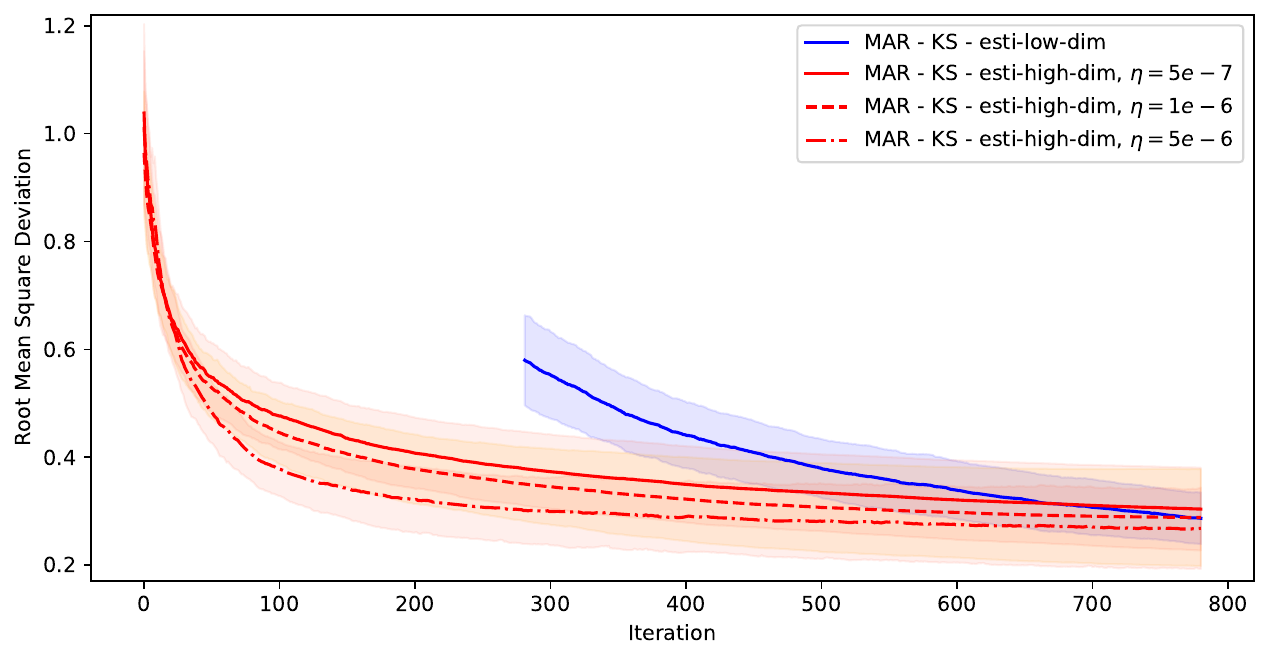}}
\caption{\textit{Root mean square deviation.} The red curves are the mean RMSDs of the high-dimensional procedure, taken over $30$ simulations each. The blue curve is the mean RMSD of the low-dimensional  procedure, taken over the same $90$ simulations. The shaded areas represent the corresponding one standard deviations. Other simulation settings: $N = 20$, $F = 5$, $M = 12$, number of model parameters = $1500$, $\mbox{significance level of } \chi^2 \mbox{ test} = 0.1$, $t_{0} = 20$, $\lambda_0 = 0.03$. In the first high dimensional phase, the accurate estimator of the low-dimensional  procedure is not available.}
\label{fig: rmsd_aug}
\end{center}
\vskip -0.2in
\end{figure}

\begin{figure}[!ht]
\begin{center}
\centerline{\includegraphics[width=0.65\textwidth]{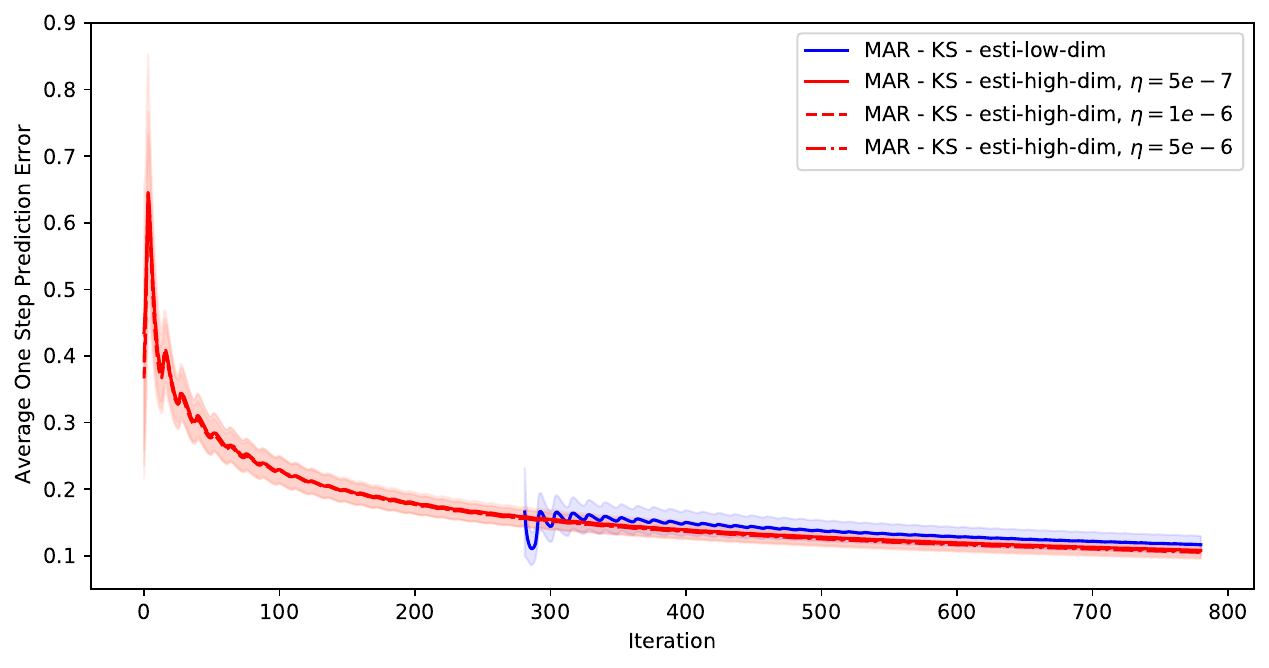}}
\caption{\textit{Average one step prediction error.} The red curves are the mean prediction error of the high-dimensional procedure, taken over $30$ simulations each. The blue curve is the mean prediction error of the low-dimensional  procedure, taken over the same $90$ simulations. The shaded areas represent the corresponding one standard deviations. Other simulation settings: $N = 20$, $F = 5$, $M = 12$, number of model parameters = $1500$, $\mbox{significance level of } \chi^2 \mbox{ test} = 0.1$, $t_{0} = 20$, $\lambda_0 = 0.03$.}
\label{fig: Average_one_step_prediction_error_aug}
\end{center}
\vskip -0.2in
\end{figure}

 The numeric results of \CB{$30$} simulations, with $N = 20$, $F = 5$, and $M = 12$ {\CM are now discussed}. We test three different step sizes $\eta$, $5\times10^{-7}$, $1\times10^{-6}$, and $5\times10^{-6}$. With each value \CB{$30$} independent simulations {\CM are performed}. Figure \ref{fig: rmsd_aug} and \ref{fig: Average_one_step_prediction_error_aug} plot the evolution of error metrics \eqref{eq: pred_err_aug} and \eqref{eq: rmsd}, respectively. For better visualization effect, since the performance of the low-dimensional  procedure does not depend on $\eta$, we only show one mean metric curve instead of $3$, in the two figures, which is calculated from the results of these \CB{$90$} simulations. 

Figure \ref{fig: rmsd_aug} shows the evolutions of estimator errors of the true coefficient $A$. Overall, {\CM it can be seen} that the step size $\eta$ influences the convergence speed of the high-dimensional estimators. Specifically, the RMSD of the high-dimensional estimator with $\eta = 5\times10^{-6}$ converges the most quickly for the first $200$ samples, after which it starts to slow down and decrease more slowly than the RMSDs of the other two step sizes. For the low-dimensional estimator, when it is available, the RMSD decreases very fast and faster in higher dimension than those of high-dimensional estimators. This was expected, because the regularization in the high-dimensional estimator will cause bias, while the low-dimensional estimator is consistent as {\CM shown} in Theorem \ref{prop: asymptotics new OLS} and the comments below. 

The impact of the step size is not significant in terms of the prediction error as shown in Figure \ref{fig: Average_one_step_prediction_error_aug}., possibly because the latter aggregates the estimator error of $A$ with other quantities such as the samples and the estimator error of trend. Besides, it shows that the prediction error of the low-dimensional estimator is worse than that of the high-dimensional one. This follows the fact that the estimator error of the former is worse almost all the time during the testing period, even though it starts to descend faster at the end.

For synthetic data, it is not surprising that the RMSD from the low-dimensional  procedure will tend toward zero, because these data are precisely sampled from the model used in the method derivation. On the other hand, at each online iteration, the OLS estimation is calculated accurately. In contrast, for the homotopy algorithms, they still introduce small errors, possibly due to the following assumptions used in the derivation of the method: 1. the active elements of $\KN^1$ of the algorithm inputs are not zero, which means that, some zero $\left(\as\right)_{i(k)}, k \in \KN^1$ should not satisfy the first equations of the optimality condition (\ref{eq: opcon}) and (\ref{eq: opconhomo}), due to the computation coincidence; 2. the sub-derivatives of those zero elements are strictly within $(-1, 1)$; 3. every $\lambda_t$ at which we calculate the derivative as in Section \ref{sec: lambda selection} is not a critical point. Thus, for example, small non-zero entry values in the inputs may cause the numerical errors.  
However, in real applications, the only available metric which allows the performance comparison is the prediction error \eqref{eq: pred_err_aug}. 

\begin{figure}[!ht]
\begin{center}
\centerline{\includegraphics[width=0.65\textwidth]{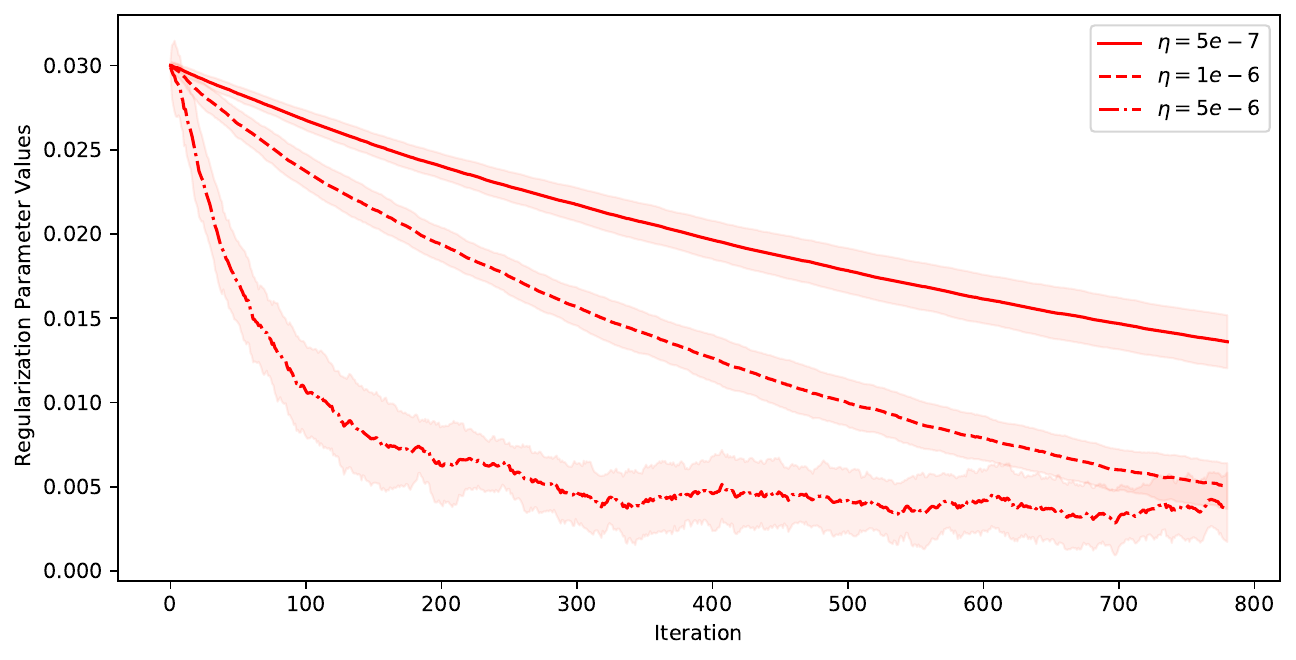}}
\caption{\textit{Regularization parameter evolution.} The red curves are the mean regularization parameter values, taken over $30$ simulations each. The shaded areas represent the corresponding one standard deviations. Other simulation settings: $N = 20$, $F = 5$, $M = 12$, number of model parameters = $1500$, $t_{0} = 20$, $\lambda_0 = 0.03$.}
\label{fig: lambda_aug}
\end{center}
\vskip -0.2in
\end{figure}

Figure \ref{fig: lambda_aug} demonstrates the performance of the updating method of the regularizing parameter $\lambda$, and the impact from different step size values $\eta$. The curves emphasize the convergence of the estimation updated by the high-dimensional procedure. Moreover, {\CM it can be observed} that all three $\lambda_t$ are decreasing, which was expected from the experiment design. On the other hand, the results show again that a larger step size will make the convergence faster. In this figure, {\CM it can be observed} additionally $\lambda_t$ curves of larger steps are noisier. 

\begin{figure}[!ht]
\begin{center}
\centerline{\includegraphics[width=0.65\textwidth]{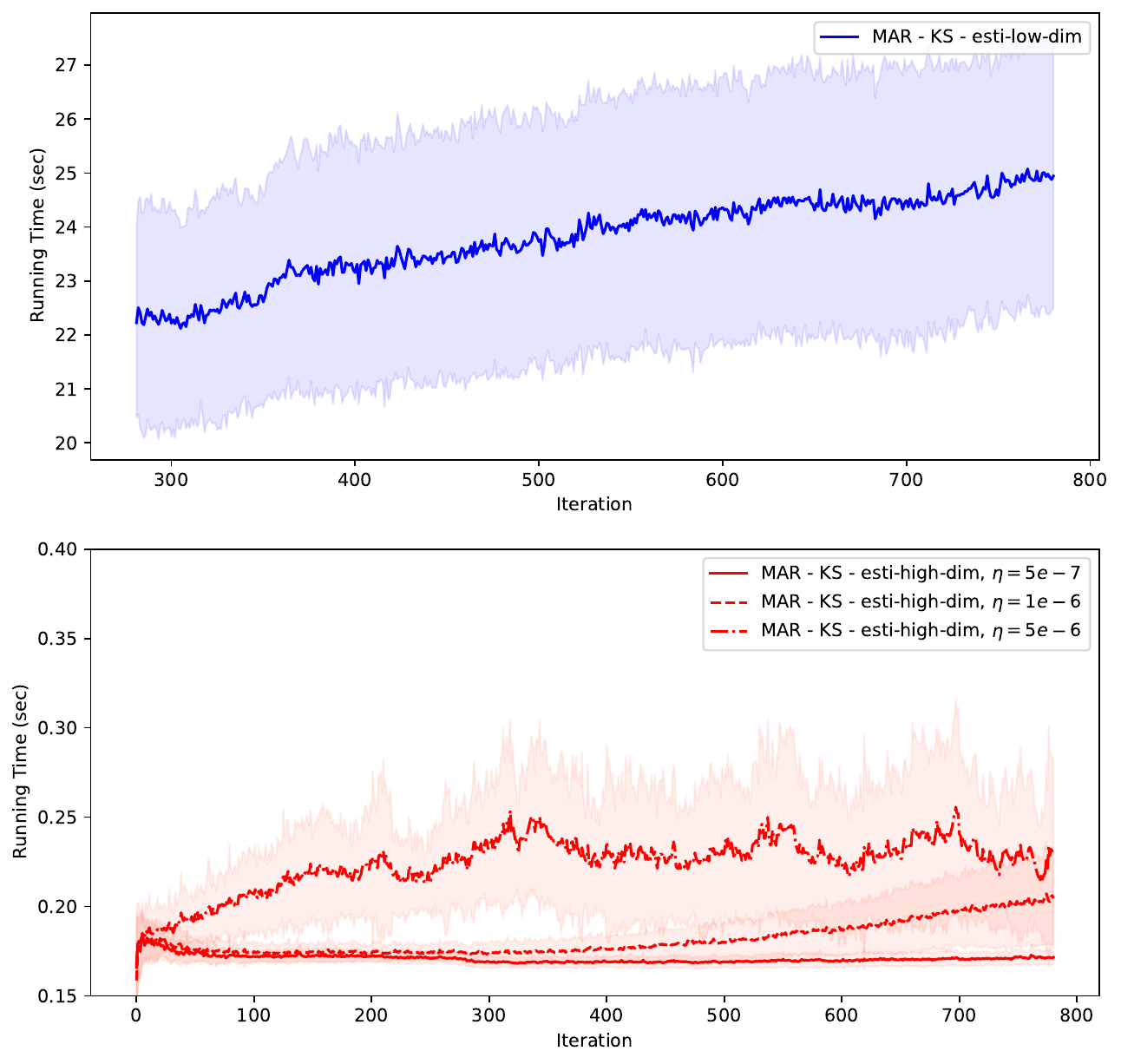}}
\caption{\textit{Running time of each online update.} The red curves are the mean running time of the high-dimensional procedure, taken over $30$ simulations each. The blue curve is the mean running time of the low-dimensional  procedure, taken over the same $90$ simulations. The shaded areas represent the corresponding one standard deviations. Other simulation settings: $N = 20$, $F = 5$, $M = 12$, number of model parameters = $1500$, $\mbox{significance level of } \chi^2 \mbox{ test} = 0.1$, $t_{0} = 20$, $\lambda_0 = 0.03$.}
\label{fig: run_time_aug}
\end{center}
\vskip -0.2in
\end{figure}

For our approaches, we also record the running time of their online updates in Figure \ref{fig: run_time_aug}. Firstly, it is clear that updating the Lasso solutions by the homotopy algorithms saves considerable time, which is on average $0.20$ seconds for the graph size $N = 20$, $F = 5$. The running time of the accelerated proximal gradient descent performed in the beginning of these simulations costs more than $3$ seconds. By contrast, an update using the low-dimensional  procedure takes $25$ seconds on average.  {\CM It can also be noticed} that the high-dimensional  procedure with larger step size runs slower, because the updated regularization parameter is quite different from the preceding one, as evidenced by the results in Figure \ref{fig: lambda_aug}. This aspect of evaluation is not used for the competitor methods. The reason is three-fold. Firstly, the computation time for the methods of \cite{chen2021autoregressive} depends on the stopping criteria of the corresponding algorithms, which simultaneously impact the performance of methods. Thus it is difficult to set ideal stopping criteria which allow us to compare both performance and running time. Secondly, they do not consider sparsity in estimation, thus the calculation of their estimators consists in nature fewer steps. Lastly, our way to code their algorithms also impacts their running time. Thus, we focus on the comparison with the performance by setting the stopping criteria sufficiently strict.

\begin{figure}[!ht]
\begin{center}
\centerline{\includegraphics[width=0.65\textwidth]{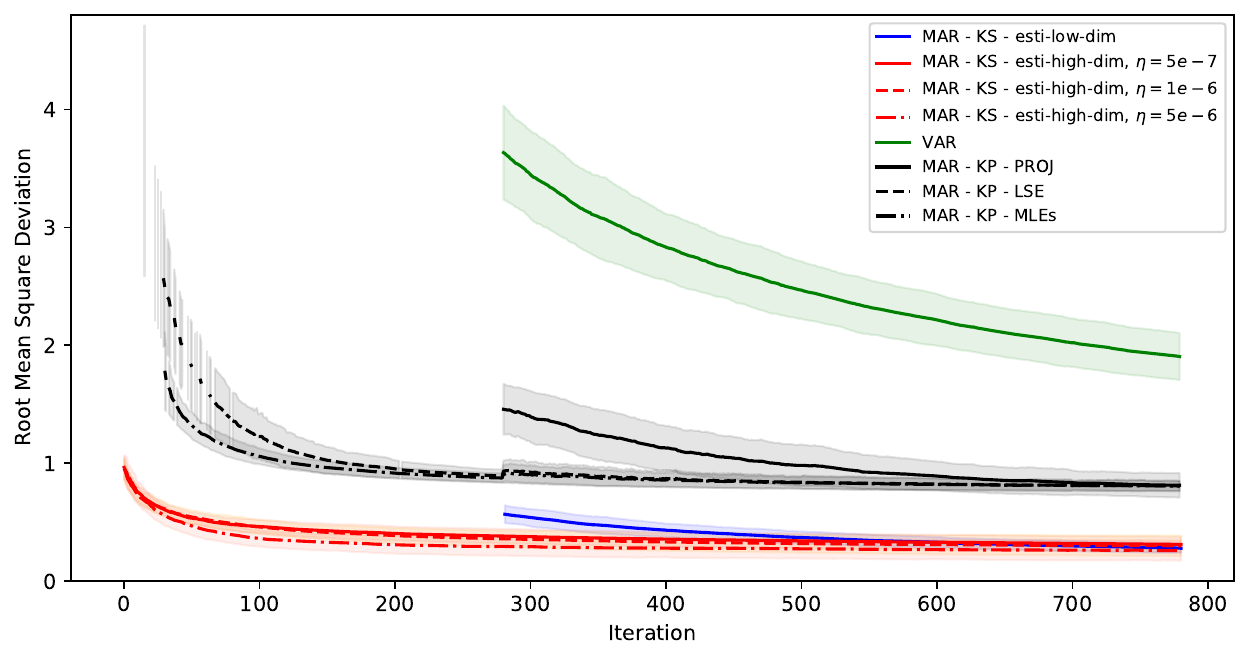}}
\caption{\textit{Root mean square deviation (true sample model is ours).} The red curves are the mean RMSDs of the high-dimensional procedure, each taken over $30$ simulations each. The blue curve is the mean RMSD of the low-dimensional procedure, taken over the same $90$ simulations. The black curves are the mean RMSDs of the model proposed by \cite{chen2021autoregressive}, each corresponding to an estimator and taken over the same $90$ simulations. The green curve is the mean RMSD of the VAR(1) using the OLS estimation, taken over the same $90$ simulations as well. The shaded areas represent the corresponding one standard deviations. Other simulation settings: $N = 20$, $F = 5$, $M = 12$. For our approaches: $\mbox{significance level of } \chi^2 \mbox{ test} = 0.1$, $t_{0} = 20$, $\lambda_0 = 0.03$. For the LSE approach of \cite{chen2021autoregressive}, the corresponding algorithm is initialized by $A = I_N$ and $B = I_F$ if the Proj estimator $B_{Proj} \otimes A_{Proj}$ is not yet available, by $A = A_{Proj}$ and $B = B_{Proj}$ otherwise. It is stopped by $mean(\|A-A_{old}\|_{\mathbf{F}}, \|B-B_{old}\|_{\mathbf{F}}) < 1e-5$. For the MLEs approach of \cite{chen2021autoregressive}, the corresponding algorithm is initialized by $A = I_N, B = I_F, \Sigma_r = I_N, \Sigma_N = I_F$ if the Proj estimator $B_{Proj} \otimes A_{Proj}$ is not yet available, by $A = A_{Proj}, B = B_{Proj}, \Sigma_r = I_N, \Sigma_N = I_F$ otherwise. It is stopped by $mean(\|A-A_{old}\|_{\mathbf{F}}, \|B-B_{old}\|_{\mathbf{F}}, \|\Sigma_r-\Sigma_r^{old}\|_{\mathbf{F}}, \|\Sigma_c-\Sigma_c^{old}\|_{\mathbf{F}}) < 1e-5.$ $I_N$ and $I_F$ denote respectively the $N \times N$ and $F \times F$ identity matrices. The index ``old'' denotes the iterate from the previous step. }
\label{fig: rmsd_aug_comp_trueisours}
\end{center}
\vskip -0.2in
\end{figure}
\begin{figure}[!ht]
\begin{center}
\centerline{\includegraphics[width=0.65\textwidth]{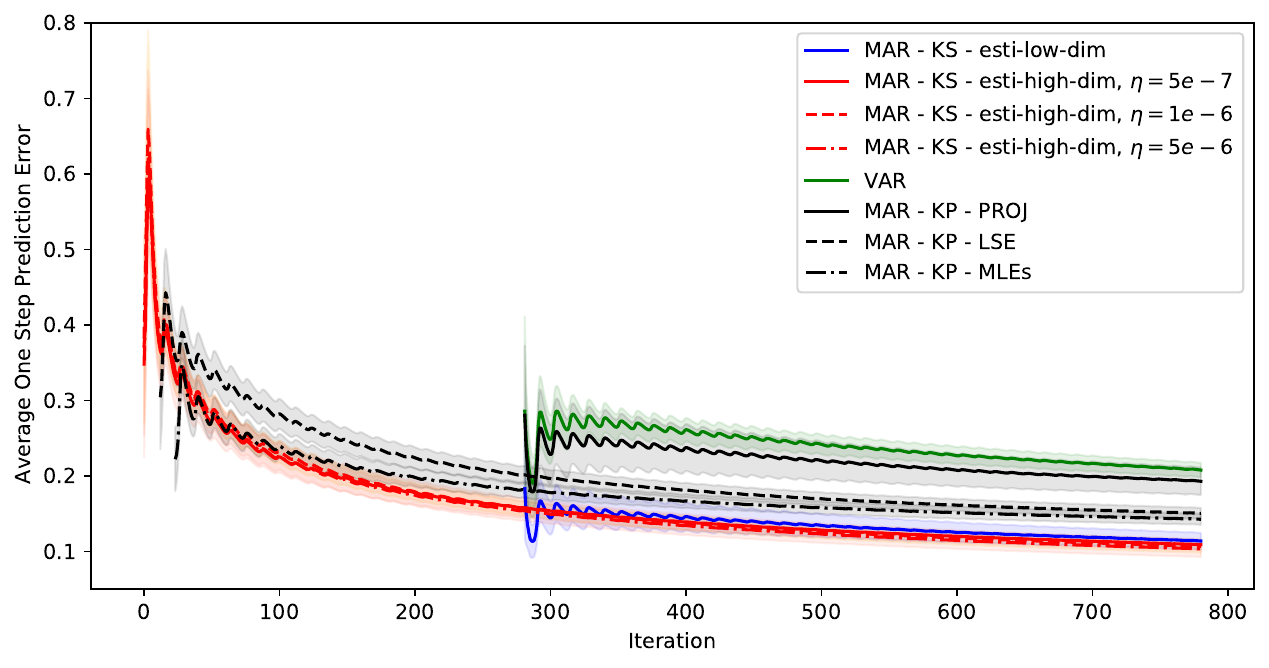}}
\caption{\textit{Average one step prediction error (true sample model is ours).} The red curves are the mean errors of the high-dimensional procedure, each taken over $30$ simulations each. The blue curve is the mean error of the low-dimensional procedure, taken over the same $90$ simulations. The black curves are the mean errors of the model proposed by \cite{chen2021autoregressive}, each corresponding to an estimator and taken over the same $90$ simulations. The green curve is the mean error of the VAR(1) using the OLS estimation, taken over the same $90$ simulations as well. The shaded areas represent the corresponding one standard deviations. Other settings see the caption of Figure \ref{fig: rmsd_aug_comp_trueisours}.}
\label{fig: Average_one_step_prediction_error_aug_comp_trueisours}
\end{center}
\vskip -0.2in
\end{figure}
\begin{figure}[!ht]
\begin{center}
\centerline{\includegraphics[width=0.65\textwidth]{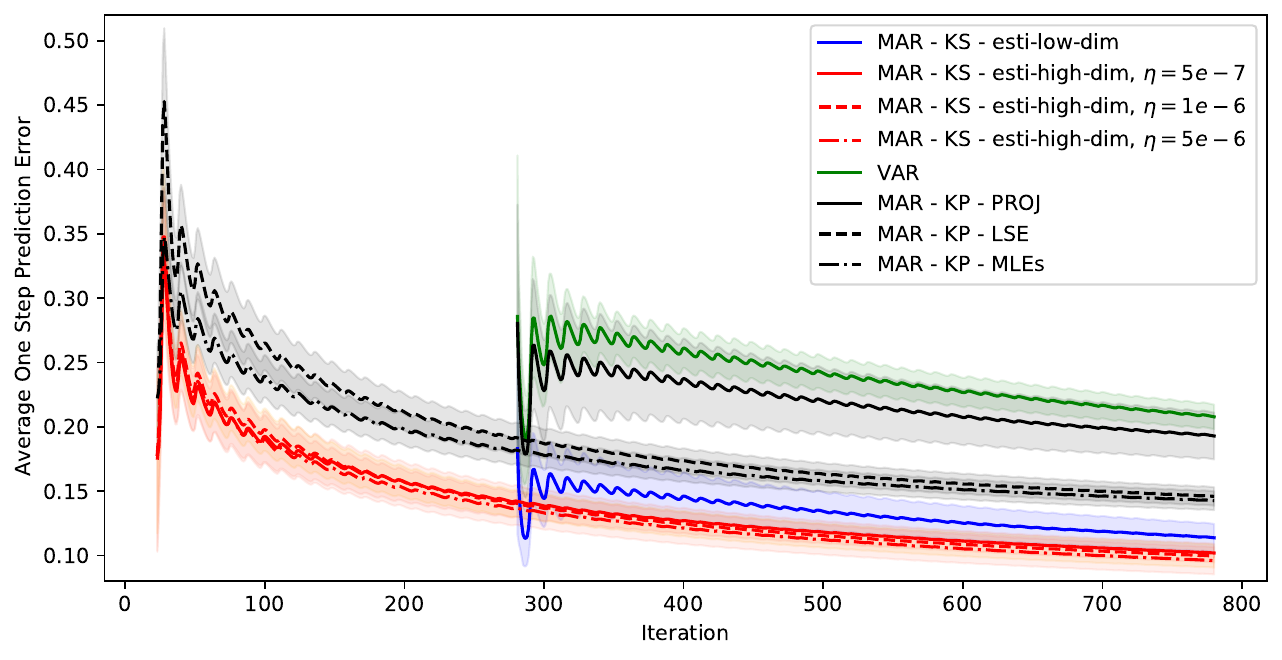}}
\caption{\textit{Average one step prediction error (true sample model is ours).} The curves are calculated with the same values as in Figure \ref{fig: Average_one_step_prediction_error_aug_comp_trueisours}, except that the calculation for our high-dimensional estimators and the LSE estimator do not use their errors available earlier than the MLEs estimator . }
\label{fig: Average_one_step_prediction_error_aug_comp_trueisours2}
\end{center}
\vskip -0.2in
\end{figure}

Now the results of the competitors {\CM are discussed}. Figures \ref{fig: rmsd_aug_comp_trueisours}, \ref{fig: Average_one_step_prediction_error_aug_comp_trueisours}, and \ref{fig: Average_one_step_prediction_error_aug_comp_trueisours2} report their estimation and prediction performance in comparison with ours. Firstly, in Figure \ref{fig: rmsd_aug_comp_trueisours}, {\CM it can be seen} that our estimators outperform the rest four estimators. Note that our low-dimensional estimator (blue) is the projected OLS estimator (green), thus the green curve represents in effect a consistent estimator. However since it is not applied techniques to reduce the dimensionality, in a fairly long period of our simulations, it can not beat the other estimators. This numerically evidences our claim in the introduction that vector models with reasonable structures in parameters are preferable than the direct use of vector models. Note that the KP structure is not the correct parameter structure in the true sample model, but it is still related to the KS structure. 

Between our methods and those of \cite{chen2021autoregressive}, since the true model is not fair to the latters, we focus on the minimal sample sizes required by different methods. Firstly, our low-dimensional estimator and the Proj estimator of \cite{chen2021autoregressive} are both projected OLS estimators. Thus they are not available before the $NF \times NF$ matrix $\widehat{\bm \Gamma}_t(0)$ is invertible. The LSE estimator of \cite{chen2021autoregressive} is defined through the following LS problem $$\argmin_{B\otimes A} \sum\limits_{\tau} \|\x_{\tau} - (B\otimes A)\x_{\tau-1}\|_{\ell_2}^2.$$ It is solved by a proposed coordinate descent algorithm, which iterates between
\begin{equation}\label{eq: coor-desc chen}
\begin{aligned}
     & B \leftarrow \left(\sum\limits_{\tau}\bm X_{\tau}^\top A \bm X_{\tau-1}\right)\left(\sum\limits_{\tau}\bm X_{\tau - 1}^\top A^\top A \bm X_{\tau-1}\right)^{-1} \\
     & A \leftarrow \left(\sum\limits_{\tau}\bm X_{\tau} B \bm X_{\tau-1}^\top\right)\left(\sum\limits_{\tau}\bm X_{\tau - 1} B^\top B \bm X_{\tau-1}^\top\right)^{-1}. 
\end{aligned}    
\end{equation}
Thus the LSE estimator is not available before the $N \times N$ matrix $\sum_{\tau}\bm X_{\tau - 1}^\top A^\top A \bm X_{\tau-1}$  and the $F \times F$ matrix $\sum_{\tau}\bm X_{\tau - 1} B^\top B \bm X_{\tau-1}^\top$ are both invertible. In Figure \ref{fig: rmsd_aug_comp_trueisours}, {\CM it can be seen} the LSE estimator starts later than our high-dimensional estimators. To avoid inversing matrices in calculation, other algorithm frameworks could replace the coordinate descend, such as projected gradient descend. Such way, a local minimiser is calculable since the beginning. However, it still can not fix the instability problem. With regard to this point, {\CM it is also found in the figure} that, the calculable LSE estimators
before $100$ samples actually have large error mean and variance. To have better performance in high dimension, extra regularization is needed. The MLEs method also estimates directly the KP components $A$ and $B$, thus similarly to the LSE method, the estimator is available in a higher dimension than the Proj one. Nevertheless, it estimates additionally the covariance of the noise, which is limited to have a KP structure as well. The authors supplemented the previous coordinate descend \eqref{eq: coor-desc chen} algorithm by extra steps, the latter requiring to inverse more matrices. {\CM It can be seen} in Figure \ref{fig: rmsd_aug_comp_trueisours} that the MLEs curves appears later than the MLEs one, but still more earlier than the low-dimensional estimators. 

We also point out the reason which results in the jumps of LSE and MLEs curves at the starting time of the low-dimensional estimators. The two estimators are calculated by algorithms which can only converge to local minimizers, thus the solutions depend on the initial values. In the paper \cite{chen2021autoregressive}, it is suggested to use the Proj estimation. However we consider higher dimension, thus before the Proj estimation is available, we use other initial values (identity matrices), the change of initialization causes the jumps.


Figure \ref{fig: Average_one_step_prediction_error_aug_comp_trueisours} reports the prediction errors of different methods. To calculate the curves for LSE and MLEs estimators,  the moving average {\CM is taken} by ignoring the unavailable estimations occurred occasionally in the past. If the estimation is not available for the whole past with respect to a given time, the error metric is given Nan for that time. {\CM It can be seen} the prediction errors differ in the same way as the estimation errors, except that around $50$ samples, the LSE and MLEs methods slightly surpass our high-dimensional method. This is because the error metric is a moving average, thus its value at a given time aggregates all the past ones. Since the high-dimensional method is the only method which can work with extremely small sample sizes, on the other hand, smaller sample size leads to worse performance in general, these inferior errors of our estimator then propagate in longer period. Therefore, we additionally plot  Figure \ref{fig: Average_one_step_prediction_error_aug_comp_trueisours2}, where  the prediction errors {\CM are re-calculated} for our high-dimensional and the LSE methods by not considering their errors which are available earlier than the MLEs method. {\CM It can be seen} that the high-dimensional method actually outperforms the LSE and MLEs for all sample sizes.

\begin{figure}[t]
\begin{center}
\centerline{\includegraphics[width=0.65\textwidth]{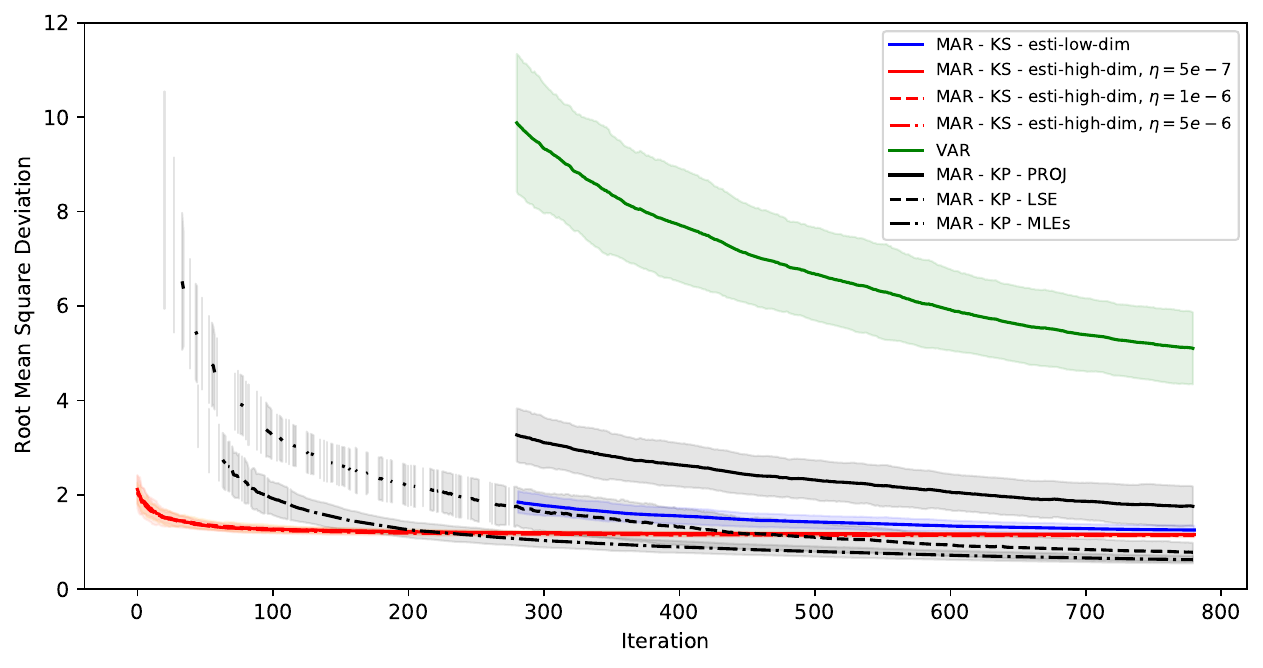}}
\caption{\textit{Root mean square deviation (true sample model is \cite{chen2021autoregressive}'s).} Other simulation settings are the same as in the caption of Figure \ref{fig: rmsd_aug_comp_trueisours}. }
\label{fig: rmsd_aug_comp_trueischens}
\end{center}
\vskip -0.2in
\end{figure}

In the previous group of simulations, both our high and low dimensional estimators outperform the competitors. The high-dimensional one is additionally more advantageous in terms of required sample size. These results are partly linked to the fact that the true sample model is ours. Now we report the results with samples generated by the MAR model of \cite{chen2021autoregressive}. Firstly, Figure \ref{fig: rmsd_aug_comp_trueischens} shows that even with the unfair sample model, our high-dimensional estimator gives still the best estimation error in high dimension, then is beaten by the MLEs estimator around $200$ samples, and by the LSE estimator around $500$ samples. As for the Proj estimator, it is not able to beat our high-dimensional estimator within the testing period, because it relies on the performance of the OLS estimator of VAR model. 

\begin{figure}[!ht]
\begin{center}
\centerline{\includegraphics[width=0.65\textwidth]{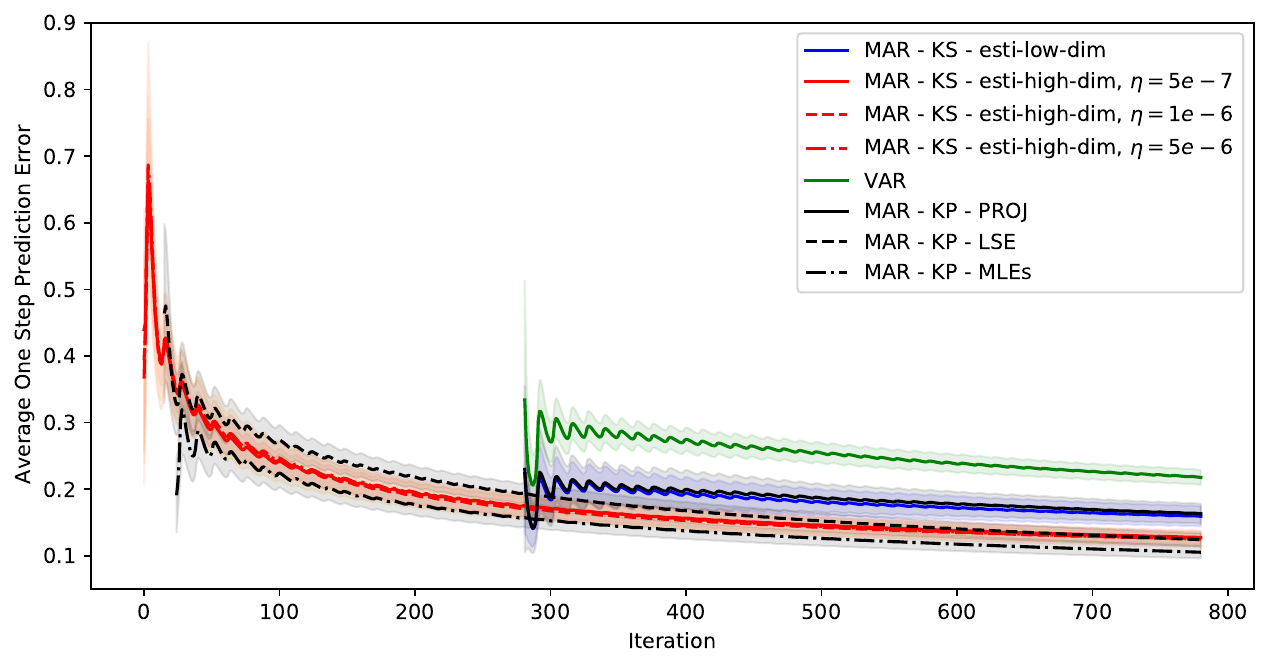}}
\caption{\textit{Average one step prediction error (true sample model is \cite{chen2021autoregressive}'s).} Other simulation settings are the same as in the caption of Figure \ref{fig: rmsd_aug_comp_trueisours}. }
\label{fig: Average_one_step_prediction_error_aug_comp_trueischens}
\end{center}
\vskip -0.2in
\end{figure}

\begin{figure}[!ht]
\begin{center}
\centerline{\includegraphics[width=0.65\textwidth]{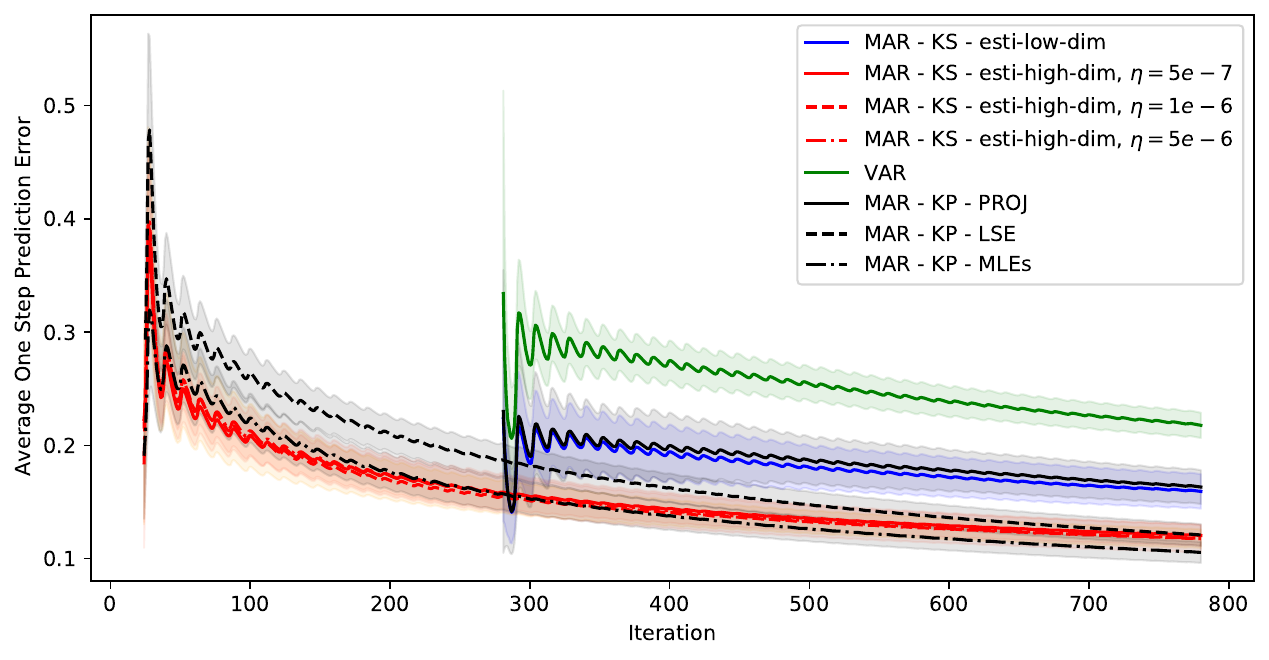}}
\caption{\textit{Average one step prediction error (true sample model is \cite{chen2021autoregressive}'s).} The curves are calculated with the same values as in Figure \ref{fig: Average_one_step_prediction_error_aug_comp_trueischens}, except that the calculation for our high-dimensional estimators and the LSE estimator do not use their errors available earlier than the MLEs estimator.}
\label{fig: Average_one_step_prediction_error_aug_comp_trueischens2}
\end{center}
\vskip -0.2in
\end{figure}

Next, Figures \ref{fig: Average_one_step_prediction_error_aug_comp_trueischens} and \ref{fig: Average_one_step_prediction_error_aug_comp_trueischens2} report the prediction performance of methods. Same conclusion can be made as the estimation performance. From these simulations, {\CM it can be seen} that the instability brought by the insufficient samples is more harmful than the model misspecification, which emphasizes the use of enough regularization in high dimension. 


\subsection{Climatology Data}\label{sec: results raw data}

In this section, a real data set {\CM is considered  that} comes from the U.S. Historical Climatology Network (USHCN) data, available at 
\url{https://www.ncdc.noaa.gov/ushcn/data-access}. The data set contains monthly averages of four climatology features, recorded at weather stations located across the United States, over years. The four features are: minimal temperature, maximal temperature, mean temperature, and precipitation. A snippet of the data set has been given in Figure \ref{fig: vector}, which illustrates these feature time series observed from a certain spatial location. A clear periodic trend can be seen from each scalar time series, with period length equal to $12$ months.
 {\CM It can also be noticed} that some observations are missing in the data set; to focus on the evaluation of learning approaches, we do not consider the stations with incomplete time series. Geographically, data {\CM are picked} only from California and Nevada for this experiment. The summary of experiment setting thus is: $N = 27, F= 4, M = 12$, total number of time points = $1523$ months (covering the years from $1894$ to $2020$). 

With this data set,  our methods {\CM are evaluated} from two aspects: as graph learning methods, we visualize the learned graphs and interpret them; as a new AR model, we compare the prediction error with the competitors. There is a third aspect of our methods which is as online algorithms. The corresponding evaluation does not depend on the nature of data. In effect, for the running time of online update, {\CM the performances are already tested and reported}  in Figure \ref{fig: run_time_aug}. On the other hand, it is self-explained that our estimators do not need to store the past samples to be updated.   

To compare the prediction errors, because the ``true'' model for real data  {\CM is not available}, the following metric formula {\CM is used} instead of the formula \eqref{eq: pred_err_aug}
\begin{equation}
\sum\limits_{\tau=1}^t\frac{\| \x_{\tau + 1} - (\widehat{\mathbf{b}}_{m(\tau+1), \tau} + \A^e_{\tau}\x_{\tau})\|_2}{t\|\x_{\tau + 1}\|_2}, \quad m(\tau+1) = (\tau + 1) \; \mbox{mod} \; M. 
\end{equation}
All methods are started the same way and same time as in the simulations. For our high-dimensional methods, we still set the initial $\lambda_0$ as $0.03$. The value does not affect the performance of high-dimensional method much, because of the adaptive tuning procedures. 



\subsubsection{Real Data Results}

\begin{figure}[!ht]
\begin{center}
\centerline{\includegraphics[width=0.65\textwidth]{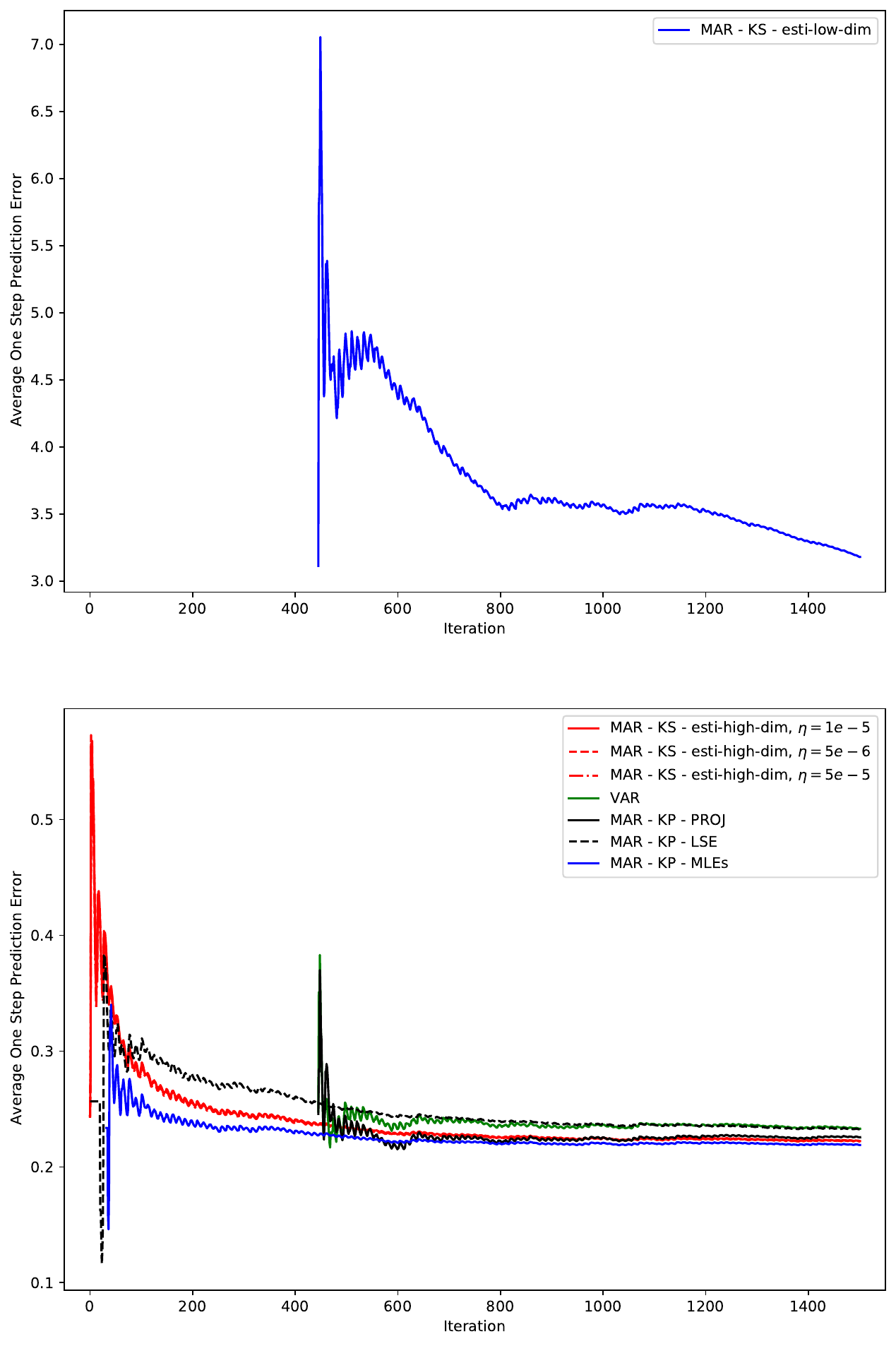}}
\caption{\textit{Average one step prediction error of raw time series.} The hyperparameter and algorithm settings are the same as in the caption of \ref{fig: rmsd_aug_comp_trueisours}.}
\label{fig: Average_one_step_prediction_error_aug_real}
\end{center}
\vskip -0.2in
\end{figure}

\begin{figure}[!ht]
\begin{center}
\centerline{\includegraphics[width=0.65\textwidth]{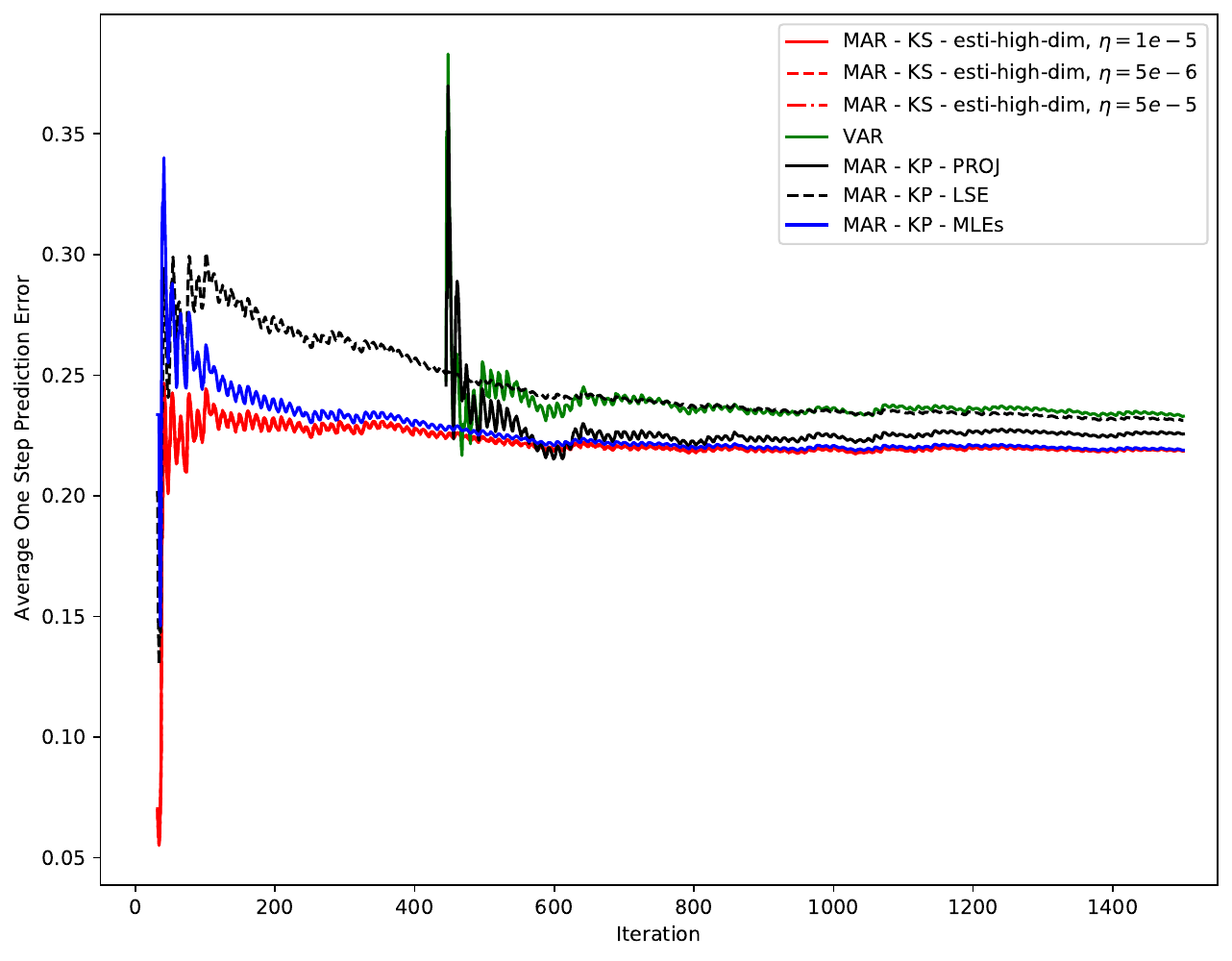}}
\caption{\textit{Average one step prediction error of raw time series.} The curves are calculated with the same values as in Figure \ref{fig: Average_one_step_prediction_error_aug_real}, except that the calculation for our high-dimensional estimators and the LSE estimator do not use their errors available earlier than the MLEs estimator.}
\label{fig: Average_one_step_prediction_error_aug_real2}
\end{center}
\vskip -0.2in
\end{figure}

\begin{figure}[!ht]
\begin{center}
\centerline{\includegraphics[width=0.65\textwidth]{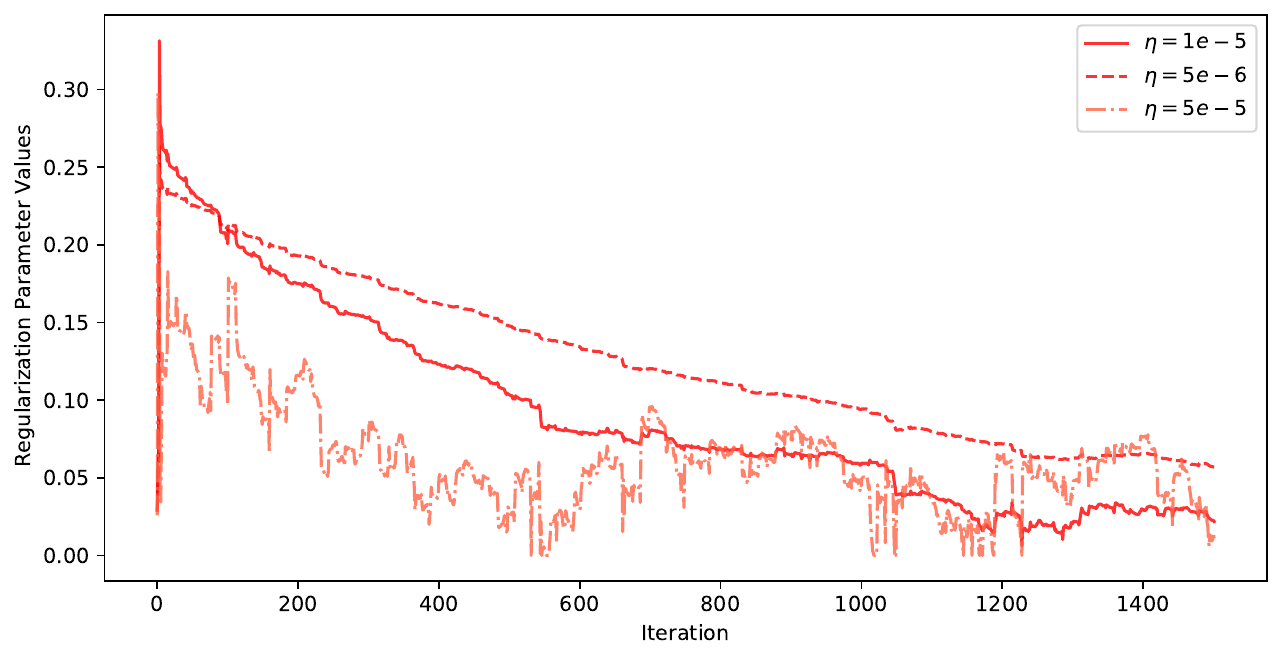}}
\caption{\textit{Regularization parameter evolution.}}
\label{fig: lambda_real_aug}
\end{center}
\vskip -0.2in
\end{figure}

The prediction errors {\CM are first reported} in Figures \ref{fig: Average_one_step_prediction_error_aug_real} and \ref{fig: Average_one_step_prediction_error_aug_real2}. To have a better visual presentation,  the curve of our low-dimensional estimator {\CM is shown} in a separate subfigure. {\CM It can be seen} that it does not perform as well as in the simulations. A possible reason can be that it fairly relies on asymptotic results, thus it is less robust to model misspecification, especially data generating models for real data are more complicated than the two MAR models, more often are implicit. By contrast, our high-dimensional estimators still outperform the rest methods for all time steps, as shown in particular by Figure \ref{fig: Average_one_step_prediction_error_aug_real2}.  Three different values of step size {\CM are tested}, they do not have significant impact on the prediction performance of the estimator as before. Comparing the counterpart results from simulations in \ref{fig: Average_one_step_prediction_error_aug_comp_trueischens2} and \ref{fig: Average_one_step_prediction_error_aug_comp_trueisours2}, we find that further the ``true" data model is away from our proposed model, less impact the step size will has on the prediction error of the estimator. The evolution of $\lambda_t$ associated to the three step sizes {\CM is also shown} in Figure \ref{fig: lambda_real_aug}. The same impact of step size on the evolution of $\lambda$ can be found as in Figure \ref{fig: lambda_aug}.

\begin{figure}[!ht]
\begin{center}
\centerline{\includegraphics[width=1.0\textwidth]{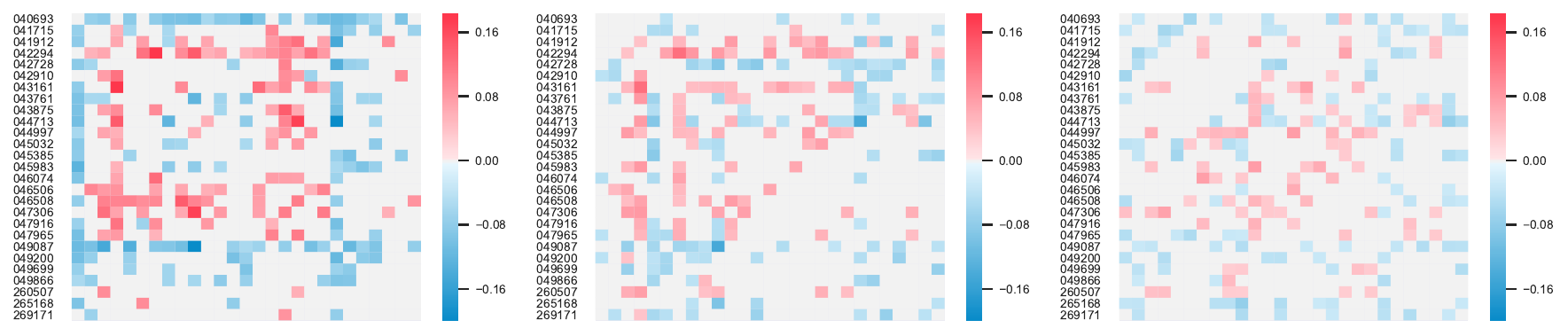}}
\caption{\textit{Updated spatial graph by the low-dimensional procedure at different times.} $t = 507$ (left), $t = 1015$ (middle), and $t = 1522$ (right). Experiment settings: $N = 27$, $F = 4$, $M = 12$, number of model parameters = $1761$, $\mbox{significance level of } \chi^2 \mbox{ test} = 0.1$, $\eta = 1e-5$, $t_{0} = 20$, $\lambda_0 = 0.03$. The row labels are the $6$-digit Cooperative Observer Identification Number of the corresponding weather stations.}
\label{fig: Est_real_Wald_aug}
\end{center}
\vskip -0.2in
\end{figure}

\begin{figure}[!ht]
\begin{center}
\centerline{\includegraphics[width=1.0\textwidth]{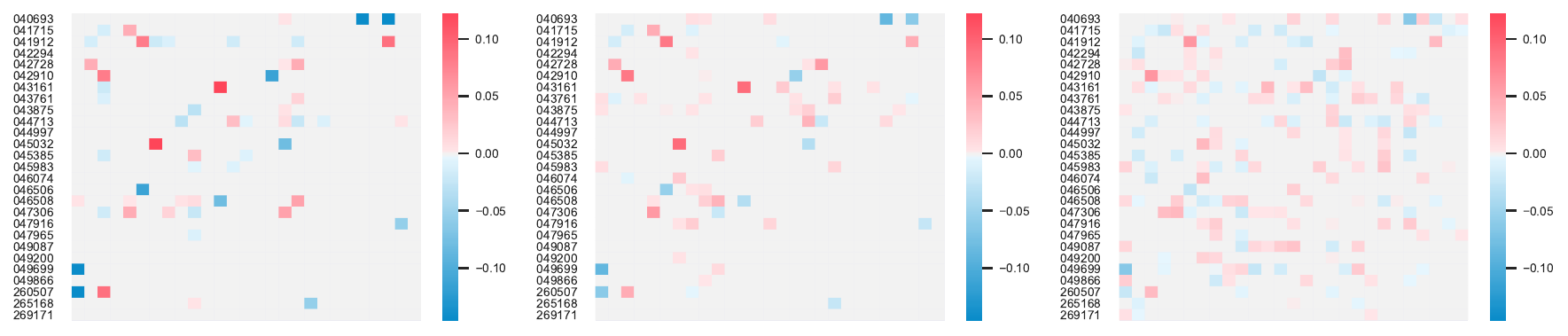}}
\caption{\textit{Updated spatial graph by the high-dimensional procedure at different times.} $t = 507$ (left), $t = 1015$ (middle), and $t = 1522$ (right). Experiment settings are the same as in Figure \ref{fig: Est_real_Wald_aug}.}
\label{fig: Est_real_aug}
\end{center}
\vskip -0.2in
\end{figure}

\begin{figure}[!ht]
\begin{center}
\centerline{\includegraphics[width=0.7\textwidth]{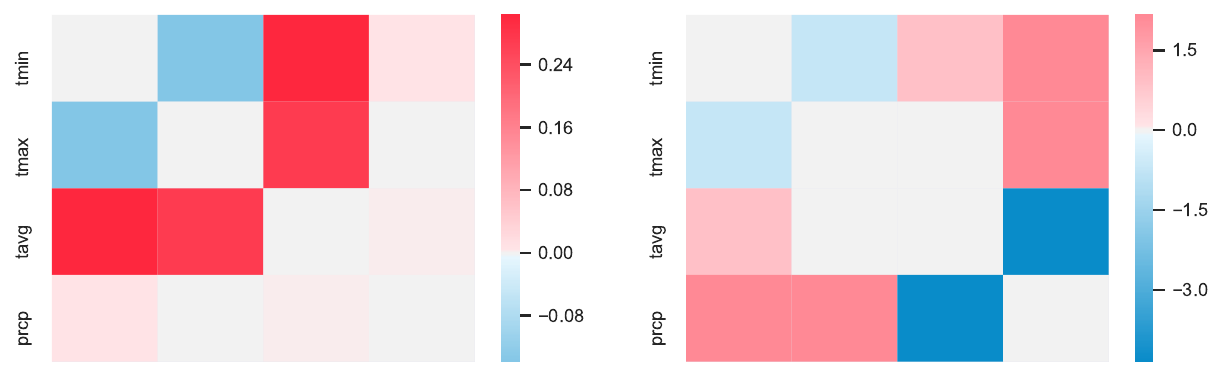}}
\caption{\textit{Updated feature graph at $t = 1522$.} the low-dimensional procedure (left), and the high-dimensional procedure (right). Experiment settings: $N = 27$, $F = 4$, $M = 12$, number of model parameters = $1761$, $\mbox{significance level of } \chi^2 \mbox{ test} = 0.1$, $\eta = 10^{-5}$, $t_{0} = 20$, $\lambda_0 = 0.03$.}
\label{fig: Est_real_AF_aug}
\end{center}
\vskip -0.2in
\end{figure}

\begin{figure}[t]
\begin{center}
\centerline{\includegraphics[width=\textwidth]{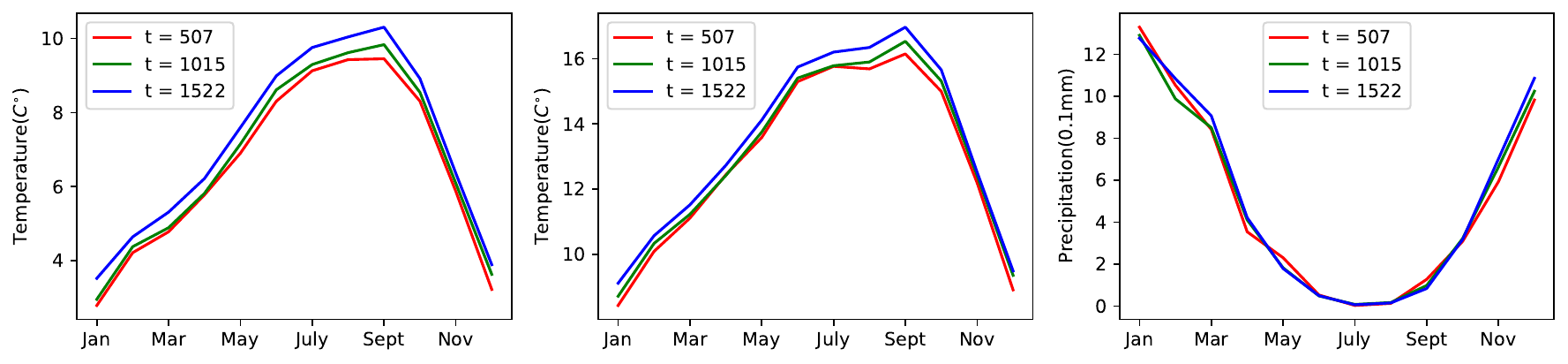}}
\caption{\textit{Estimated trends along years.} On the left, middle, right are the estimated trends at different years of Station USH00040693 for minimal temperature, average temperature, and precipitation respectively. 
Experiment settings: $N = 27$, $F = 4$, $M = 12$.}
\label{fig: trend}
\end{center}
\vskip -0.2in
\end{figure}

The inferred graphs and trends {\CM are now visualized} from our methods. We report the results corresponding to $\eta = 1e-5$ since it has converged and is less noisy. Figures \ref{fig: Est_real_Wald_aug} and \ref{fig: Est_real_aug} show the spatial graphs learned by the two proposed approaches in Section \ref{sec: online learning aug} updated at different times. {\CM It can be seen} that, for the high-dimensional method, when more observations are received, it finds that more location pairs actually have a Granger causal effect on each other. On the other hand, compared to the estimated graphs from the high-dimensional procedure, those from the low-dimensional procedure vary more along time. 
Figure \ref{fig: Est_real_AF_aug} shows the last updated feature graphs. {\CM It can be seen} that the estimated feature relationships from the two approaches coincide in tmin and tmax, tmin and tavg, tmin and prcp. However, the relationship between tavg and prcp is very weak in the high-dimensional estimation, while strong in the low-dimensional estimation. Figure \ref{fig: trend} reports the evolution of estimated trends from one representative spatial location along time, where {\CM it can be observed} the increase of temperature from the past to the present. 

\begin{figure}[t]
\begin{center}
\centerline{\includegraphics[width=0.4\textwidth]{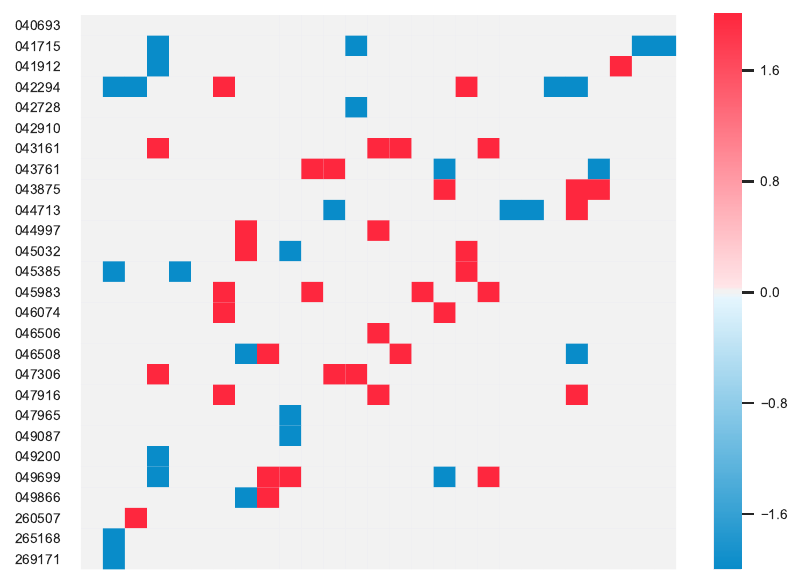}
\includegraphics[width=0.5\textwidth]{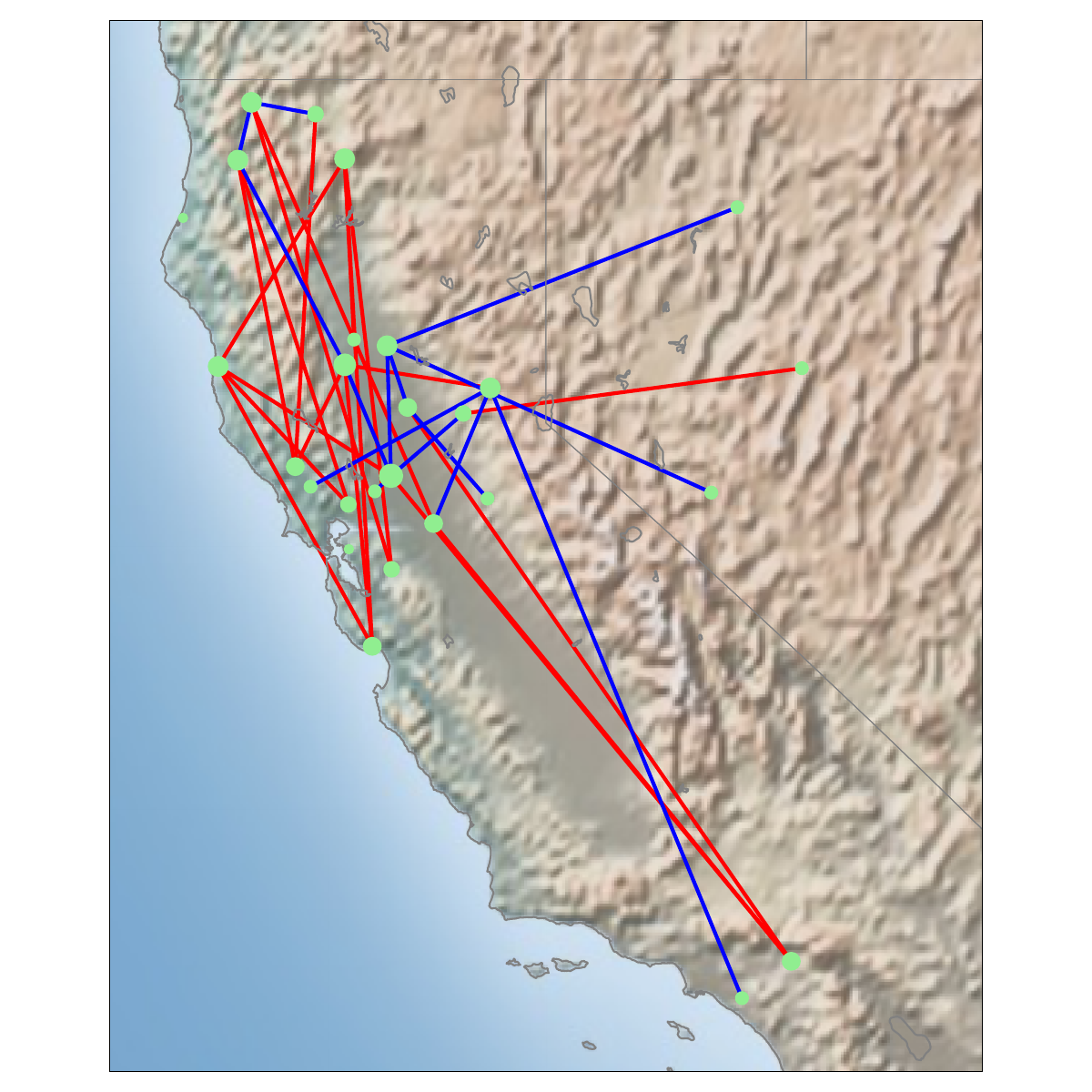}}
\caption{\textit{Overlap spatial graph.} On the left is the adjacency matrix of an unweighed undirected graph which is the overlap of the two last updated spatial graphs in Figure \ref{fig: Est_real_aug}, with the colors reporting the common edge signs. On the right is the visualization of this overlap spatial graph on the actually geographical map. The nodes with bigger sizes connect with more nodes.}
\label{fig: Est_real_overlap_aug}
\end{center}
\vskip -0.2in
\end{figure}

Lastly, in Figure \ref{fig: Est_real_overlap_aug},  the edge overlap (considering the signs of weights) of the two last updated spatial graphs {\CM is plotted}, where this spatial graph superimposed on the actual geographical graph {\CM is also visualized}. {\CM It can be seen} that the remote weather stations have less dependency with other stations, while more edges appear within the area where lots of stations are densely located together. These observations imply that the inferred graphs provide the consistent weather patterns with geographical features. 


\section{Conclusion}
In this paper, \textcolor{black}{a novel auto-regressive model for matrix-variate time series with periodic trends is proposed}. The model enables graph learning from matrix-variate time series, a data type not yet considered in the literature on graph learning. Furthermore, such graph learning is performed in an online fashion. \textcolor{black}{This represents the first aspect of contribution}. Secondly, the proposed model enriches the family of matrix auto-regressive models, \textcolor{black}{distinguished by its KS structure, in contrast to the KP structure used by other MAR models}. This distinction is justified by experiments with both synthetic and real data. Thirdly, the homotopy algorithms of classical Lasso are extended to a new Lasso-type problem. The derived algorithms present an online learning framework, where the derivations do not depend on the proposed specific structure assumed in $A$. It is important to note that the specific orthogonal basis, Equations \eqref{eq: ortho basis}, was not used in the remaining technical sections. Instead, only the symbol of the orthogonal basis, $U_k$, was manipulated, along with the symbol $K_N$, indices of the bases $U_k$ for which sparsity regularization was imposed. \textcolor{black}{As a result, the framework can be applied to the online inference of many other parametric models under structure and (partial) sparsity assumptions, as long as the structure space is a linear subspace of the parameter space.}

\textcolor{black}{It is also worth noting that} the proposed MAR model can be further extended to tensor-variate time series. The extension principle described in the introduction still applies. \textcolor{black}{The data tensor can first be vectorized}, and the classical VAR models can then be applied. Appropriate structures are then imposed onto the coefficient matrices of the VAR model. Based on VAR models, the Granger causality structure is still encoded by the sparsity structure of the coefficient matrices. For instance, in the case of VAR$(1)$ models, an appropriate structure can be a multiway KS structure \citep{wang2022kronecker, greeneWald2019tensor} $A_1 \oplus \ldots \oplus A_d$. Imposing such a structure amounts to assuming that the total causality structure factorizes into $d$ smaller causal sub-graphs. At this point, the sparsity of certain sub-graphs (modes of the tensor) can also be pursued in estimation. In the technical development, it is only necessary to assume that the first $d_0$ sub-graphs are sparse. The value of $d_0$ and which modes of the data tensor are made the first can be freely specified by different applications.

Since the multiway KS structure still leads to a linear subspace of structured parameters, \textcolor{black}{the proposed estimators and online procedures remain valid}. \textcolor{black}{It is merely necessary} to find the correct orthogonal basis $\{U_k\}$ for the new model and assign a subset of $\{U_k\}$ indexed by $K_N$ to impose sparsity. These can then be inserted into the proposed derivation to calculate the numerical results. \textcolor{black}{In this case, the fact that multiple sub-graphs are imposed with sparsity ($\ell_1$) regularization does not make a difference. This is because} $\sum_{p=1}^{d_0}\|A_p\|_{\ell_1}$ can still be expressed as $\left\|\sum_{k \in \KN} \langle U_{k},A^0\rangle  \, \frac{1}{\|U_k\|_\mathbf{F}^2}U_{k}\right\|_{\ell_1}$, where $\KN$ contains the indices of the $U_k$ corresponding to entries of $A_1 \ldots, A_{d_0}$; thus, Equation \eqref{eq: prob_A0} remains valid. \textcolor{black}{For the low-dimensional case, the Wald test is originally performed independently for each sub-graph.} This elaborated discussion also demonstrates that the proposed inference framework is fairly generic. \textcolor{black}{This completes the conclusion section.}

\section*{Acknowledgments}
This work has been partially supported by MIAI@Grenoble Alpes,
(ANR-19-P3IA-0003).
Jérémie Bigot is a member of Institut Universitaire de France (IUF), and this work has been carried out with financial support from the IUF.

\appendix

\section{Proof of Results in Section \ref{sec:app1} and the CLT for $\widehat{\A}_t$}\label{sec: proof CLT AN_hat}
\textit{Proof of Theorem \ref{thm: asym_AN}.} By Cram\'er-Wold theorem, $\sqrt{t} \, \vect(\widecheck{\A}_t - A) \xrightarrow{d} \mathcal{N}(0, \Sigma_{ols})$ is equivalent to 
\begin{equation}\label{eq: equi_CW}
\langle \Lambda, \sqrt{t} \left(\widecheck{\A}_t - A\right) \rangle \xrightarrow{d} \mathcal{N}(0, \vect(\Lambda)^\top\Sigma_{ols}\vect(\Lambda)), \quad \forall \Lambda \in \R^{NF \times NF}.
\end{equation}
On the other hand, we can express the entries of $\svec\left(\sqrt{t} \left(\widehat{\bm \AN} - \AN\right)\right)$ as a linear function of $\widecheck{\A}_t$
\begin{equation}
    \svec\left(\sqrt{t} \left(\widehat{\bm \AN} - \AN\right)\right) = \sum_{k \in \KN} \langle U_k, \sqrt{t}\left(\widecheck{\A}_t - A\right) \rangle \, \svec(E_k).
\end{equation}
Then for all $\lambda \in \R^{\frac{N(N-1)}{2}}$, we have
$$
\lambda^\top \svec\left(\sqrt{t} \left(\widehat{\bm \AN} - \AN\right)\right) = \langle \sum_{k \in \KN}  \lambda^\top\svec(E_k) U_k, \sqrt{t}\left(\widecheck{\A}_t - A\right) \rangle.
$$
Let $\Lambda$ in Equation (\ref{eq: equi_CW}) be $\sum_{k \in \KN}  \lambda^\top\svec(E_k) U_k$, then we have 
$$
\lambda^\top \svec\left(\sqrt{t} \left(\widehat{\bm \AN}_{,t} - \AN\right)\right) \rangle \xrightarrow{d} \mathcal{N}(0, \vect(\Lambda)^\top\Sigma_{ols}\vect(\Lambda)).
$$
Note that, $\vect(\Lambda)  = \sum_{k \in \KN} \lambda^\top \svec(E_k) \vect(U_k).$
Thus $\vect(\Lambda)^\top\Sigma_{ols}\vect(\Lambda) = \lambda^\top \Sigma_{\N} \lambda$.
Use Cram\'er-Wold theorem again, we can get the theorem result.

$\hfill \blacksquare$

\begin{theorem}(CLT for $\widehat{\A}_t$)
\normalfont
\begin{equation}
\sqrt{t} \, \vect(\widehat{\A}_t - \A) \xrightarrow{d} \mathcal{N}(0, \Sigma_{\G}),
\end{equation}
where $\Sigma_{\G} = \sum\limits_{k,k' \in K} \vect(U_k)^\top \Sigma_{ols} \vect(U_{k'}) \left[\vect(U_k)\vect(U_{k'})^\top\right]$.
\end{theorem}
\textit{Proof:} The proof is similar as before. Because, we can express any entries of $\widehat{\A}_{t}$ as a linear function of $\widecheck{\A}_t$:
\begin{equation}
    \widehat{\A}_{t} = \sum_{k \in K} \langle U_{k}, \widecheck{\A}_t\rangle U_{k}.
\end{equation}
Thus, for all $\Lambda' \in \R^{NF \times NF}$, we have
$$\langle \Lambda', \sqrt{t} \left(\widehat{\A}_t - A\right) \rangle = \langle \sum_{k \in K} \langle \Lambda', U_{k}\rangle U_{k}, \sqrt{t} \left(\widecheck{\A}_t - A\right) \rangle.$$
Let $\Lambda$ in Equation (\ref{eq: equi_CW}) be $\sum_{k \in K} \langle \Lambda', U_{k}\rangle U_{k}$, 
then  
$$
\langle \Lambda', \sqrt{t} \left(\widehat{\A}_t - A\right) \rangle
\xrightarrow{d} \mathcal{N}(0, \vect(\Lambda')^\top \Sigma_{\G}\vect(\Lambda')).
$$
Use Cram\'er-Wold theorem again, we can get the theorem result. The distribution in this theorem is degenerate.

$\hfill \blacksquare$

\textit{Proof of Corollary \ref{coro: Wald test}.} The proof is an adaption of \citet[Section 3.6]{lutkepohl2005new}. We first construct the following matrix:
\begin{equation}
    C = 
\begin{pmatrix}
\vdots\\
\svec(E_{h_k})^\top\\
\vdots
\end{pmatrix} \in \R^{P \times \frac{N(N-1)}{2}}.
\end{equation}
Then test $H_0$ versus $H_1$ equals to 
$$
H_0': C \svec(\AN) = 0 \; \mbox{versus} \; H_1': C \svec(\AN) \neq 0.
$$
Following CLT \ref{thm: asym_AN}, we have 
$$\sqrt{t} \, C\svec(\widehat{\bm \AN}_{,t} - \AN) \xrightarrow{d} \mathcal{N}(0, C\Sigma_{\N}C^\top).$$ Hence, when $H_0'$ holds, 
$$\sqrt{t} \, C\svec(\widehat{\bm \AN}_{,t}) \xrightarrow{d} \mathcal{N}(0, C\Sigma_{\N}C^\top).
$$ 
Then by Proposition C.2 (4) in \cite{lutkepohl2005new}, we have
$$
\sqrt{t} \, \left[C\widehat{\mb{\Sigma}}_{\N,t}C^\top\right]^{-\frac{1}{2}}C\svec(\widehat{\bm \AN}_{,t}) \xrightarrow{d} \mathcal{N}(0, I_P), 
$$
where $\widehat{\mb{\Sigma}}_{\N,t} =  \hspace{-0.2cm} \sum_{k, k' \in \KN} \hspace{-0.2cm} \vect(U_k)^\top \widehat{\mb{\Sigma}}_{ols, t} \vect(U_{k'}) \, \left(\svec(E_k)\svec(E_{k'})^\top\right)$ is the consistent estimator of $\mb{\Sigma}_{\N}$. Then by continuous mapping theorem:
$$
t \, \hat{\bm \alpha}_t^\top \left[C\widehat{\mb{\Sigma}}_{\N,t}C^\top\right]^{-1} \hat{\bm \alpha}_t \xrightarrow{d} \chi^2(P).
$$
Note that $C\svec(\widehat{\bm \AN}_{,t}) = \hat{\bm \alpha}_t$, and $(\svec(E_k))_{k \in \KN}$ are orthonormal basis in $\R^{\frac{N(N-1)}{2}}$, thus we have $
C\widehat{\mb{\Sigma}}_{\N,t}C^\top = \widehat{\mb{\Sigma}}_{W,t}$. 

$\hfill \blacksquare$

\section{Proof of Proposition \ref{prop: asymptotics new OLS}}\label{sec: proof aug asymp}

From Definition \eqref{eq: new OLS}, we have 
$$
\begin{aligned}
\spc &= \sum\limits_{m=0}^{M-1} \frac{p_{m,t}}{t}\left(\frac{\sum\limits_{\tau \in I_{m,t}}\x_{\tau-1}\x_{\tau-1}^\top}{p_{m,t}} - \underline{\x}_{m-1,t}\underline{\x}_{m-1,t}^\top\right) \\
&=  \frac{\sum\limits_{\tau = 1}^t\x_{\tau-1}\x_{\tau-1}^\top}{t} - \sum\limits_{m=0}^{M-1} \frac{p_{m,t}}{t}\left(\underline{\x}_{m-1,t}\underline{\x}_{m-1,t}^\top\right).
\end{aligned}
$$
Plug $\mb{x}_t = \mbox{b}_t^0 + \mb{x}_t^{\prime}$ in the last equation above, we can get the formula only with respect with $\mb{x}_t^{\prime}$
$$\spc = \spc^{\prime} - \sum\limits_{m=0}^{M-1} \frac{p_{m,t}}{t}\left(\underline{\x}_{m-1,t}^{\prime}[\underline{\x}_{m-1,t}^{\prime}]^\top\right),$$
where $\underline{\x}_{m-1,t}^{\prime} = \sum\limits_{\tau \in I_{m,t}} \frac{\x_{\tau-1}^{\prime}}{p_{m,t}}, \; m = 0, ..., M-1$, and $\spc^{\prime} := \frac{\sum\limits_{\tau = 1}^t\x_{\tau-1}^{\prime}[\x_{\tau-1}^{\prime}]^\top}{t}$. Note that $\underline{\x}_{-1,t}^{\prime} = \underline{\x}_{M-1,t}^{\prime}$.

Similarly, denote $\frac{\sum\limits_{\tau = 1}^t\x_{\tau}^{\prime}[\x_{\tau-1}^{\prime}]^\top}{t}$ by $\spcone^{\prime}$, we have 
$$\spcone = \spcone^{\prime} - \sum\limits_{m=0}^{M-1} \frac{p_{m,t}}{t}\left(\Bar{\x}_{m,t}^{\prime}[\underline{\x}_{m-1,t}^{\prime}]^\top\right),$$
with $\Bar{\x}_{m,t}^{\prime} = \sum\limits_{\tau \in I_{m,t}} \frac{\x_{\tau}^{\prime}}{p_{m,t}}, \; m = 0, ..., M-1$.
Since $(\x_{t}^{\prime})_t$ is the causal solution of VAR \eqref{eq: var1_vect}, we have 

\vspace{0.1in}

(a') $\spc^{\prime} \xrightarrow{p} \Gamma(0)$, $\spcone^{\prime} \xrightarrow{p} \Gamma(1)$,

\vspace{0.1in}

(b') $\spcone^{\prime}\left[\spc^{\prime}\right]^{-1}\xrightarrow{p} \Gamma(1)\left[\Gamma(0)\right]^{-1} = A$,

\vspace{0.1in}

(c') {\normalfont$\sqrt{t} \, \vect\left(\spcone^{\prime}\left[\spc^{\prime}\right]^{-1} -A\right) \xrightarrow{d} \mathcal{N}(0, \left[\Gamma(0)\right]^{-1} \otimes \Sigma$}.

\vspace{0.1in}

Thus, to reach the results in Proposition \ref{prop: asymptotics new OLS}, we need additionally the asymptotic properties of sample mean $\Bar{\x}_{m,t}^{\prime}$, which are given in Lemma 
\ref{lem: CLT sample mean}.

\begin{lemma}\label{lem: CLT sample mean}(CLT of $\Bar{\x}_{m,t}^{\prime}$)
$$\sqrt{p_{m,t}}\Bar{\x}_{m,t}^{\prime} \xrightarrow{d} \mathcal{N}(0,\Phi\Sigma_M\Phi^\top), \quad \forall m = 0, ..., M-1,$$ 
where $\Phi = \left(I_{NF} - A^M \right)^{-1}$, and $\Sigma_M = \sum_{h=0}^{M-1}A^h\mb{\Sigma}(A^h)^\top$. Therefore, $\Bar{\x}_{m,t}^{\prime} \xrightarrow{p} 0$.
\end{lemma}
\textit{Proof of Lemma \ref{lem: CLT sample mean}.}
Because of the periodicity, $(\x_{\tau}^{\prime})_{\tau \in I_{m,\infty}}$ is also a stationary process from VAR: $\widetilde{\mb{X}}_{t'} = \A^M\widetilde{\mb{X}}_{t'-1} + \Tilde{\mb{z}}_{t'}$, with $\Tilde{\mb{z}}_{t'}\sim \mathrm{IID}(0, \Sigma_M)$, for all $m = 0, ..., M-1$. Thus, apply Proposition $3.3$ in \citet{lutkepohl2005new}, we get the result. 

$\hfill \blacksquare$





\textit{Proof of Proposition \ref{prop: asymptotics new OLS}}.

(a) When $t \to \infty$,  $\spc = \spc^{\prime} - \sum\limits_{m=0}^{M-1} \frac{1}{M}\left(\Bar{\x}_{m,t}^{\prime}[\Bar{\x}_{m,t}^{\prime}]^\top\right) \xrightarrow{p} \Gamma(0) - 0 = \Gamma(0)$, and  $\spcone = \spcone^{\prime} - \sum\limits_{m=0}^{M-1} \frac{1}{M}\left(\Bar{\x}_{m,t}^{\prime}[\Bar{\x}_{m-1,t}^{\prime}]^\top\right) \xrightarrow{p} \Gamma(1)$, with $\Bar{\x}_{-1,t}^{\prime} := \Bar{\x}_{M-1,t}^{\prime}$.

(b) $\Bar{\x}_{m,t} =  \frac{\sum_{\tau \in I_{m,t}}\mbox{b}_{m}^0 + \x_{\tau}^{\prime}}{p_{m,t}} = \mbox{b}_{m}^0 + \Bar{\x}_{m,t}^{\prime} \xrightarrow{p} \mbox{b}_{m}^0, \; \forall m = 0, ..., M-1$. Since asymptotically, $\Bar{\x}_{m,t} = \underline{\x}_{m,t}$, thus both means can be used to estimate $\mbox{b}_{m}^0$. On the other hand, based on (a), using continuous mapping theorem on the matrix inverse, we have $\widecheck{\A}_t = \spcone\left[\spc\right]^{-1}\xrightarrow{p} A.$

(c) When $t \to \infty$, $\widecheck{\A}_t$ equals 
$$\left[\spcone^{\prime} - \sum\limits_{m=0}^{M-1} \frac{p_{m,t}}{t}\left(\Bar{\x}_{m,t}^{\prime}[\Bar{\x}_{m-1,t}^{\prime}]^\top\right) \right] \left[\spc^{\prime} - \sum\limits_{m=0}^{M-1} \frac{p_{m,t}}{t}\left(\Bar{\x}_{m-1,t}^{\prime}[\Bar{\x}_{m-1,t}^{\prime}]^\top\right)\right]^{-1}.$$
Use Woodbury formula on the matrix inverse, we have 
$$
\begin{aligned}
\sqrt{t}(\widecheck{\A}_t - A) &= \sqrt{t}(\spcone^{\prime}\left[\spc^{\prime}\right]^{-1} - A)  - \sqrt{t}\sum\limits_{m=0}^{M-1} \frac{p_{m,t}}{t}\left(\Bar{\x}_{m,t}^{\prime}[\Bar{\x}_{m-1,t}^{\prime}]^\top\right) \left[\spc^{\prime}\right]^{-1} \\ 
&+ \frac{\sqrt{t}}{1-g}\spcone^{\prime}\left[\spc^{\prime}\right]^{-1}\sum\limits_{m=0}^{M-1} \frac{p_{m,t}}{t}\left(\Bar{\x}_{m-1,t}^{\prime}[\Bar{\x}_{m-1,t}^{\prime}]^\top\right) \left[\spc^{\prime}\right]^{-1} \\
-\frac{\sqrt{t}}{1-g}\sum\limits_{m=0}^{M-1}& \frac{p_{m,t}}{t}\left(\Bar{\x}_{m,t}^{\prime}[\Bar{\x}_{m-1,t}^{\prime}]^\top\right) \left[\spc^{\prime}\right]^{-1}\sum\limits_{m=0}^{M-1} \frac{p_{m,t}}{t}\left(\Bar{\x}_{m-1,t}^{\prime}[\Bar{\x}_{m-1,t}^{\prime}]^\top\right)\left[\spc^{\prime}\right]^{-1}, \\
\end{aligned}
$$
$\mbox{where}, g = \mbox{tr}(\sum\limits_{m=0}^{M-1} \frac{p_{m,t}}{t}\left(\Bar{\x}_{m-1,t}^{\prime}[\Bar{\x}_{m-1,t}^{\prime}]^\top\right)\left[\spc^{\prime}\right]^{-1}).$
Based on the result of (c'), to reach the same asymptotic distribution, we only need to show that, the reminder terms, namely from the second term to the last term above, all converge to $0$ in probability.

From Slutsky's theorem and Lemma \ref{lem: CLT sample mean}, we have the asymptotic result:
$$\forall m, \frac{p_{m,t}}{\sqrt{t}}\left(\Bar{\x}_{m,t}^{\prime}[\Bar{\x}_{m-1,t}^{\prime}]^\top\right) = \frac{1}{\sqrt{M}}\left(\sqrt{p_{m,t}}\Bar{\x}_{m,t}^{\prime}\right)[\Bar{\x}_{m-1,t}^{\prime}]^\top \xrightarrow{p} 0.$$
Thus, $\sqrt{t}\sum\limits_{m=0}^{M-1} \frac{p_{m,t}}{t}\left(\Bar{\x}_{m,t}^{\prime}[\Bar{\x}_{m-1,t}^{\prime}]^\top\right) \left[\spc^{\prime}\right]^{-1} \xrightarrow{p} 0.$ 

Similarly, 
$$\sqrt{t}\sum\limits_{m=0}^{M-1} \frac{p_{m,t}}{t}\left(\Bar{\x}_{m-1,t}^{\prime}[\Bar{\x}_{m-1,t}^{\prime}]^\top\right) \left[\spc^{\prime}\right]^{-1} \xrightarrow{p} 0.$$ Since, it is obvious that $\sum\limits_{m=0}^{M-1} \frac{p_{m,t}}{t}\left(\Bar{\x}_{m-1,t}^{\prime}[\Bar{\x}_{m-1,t}^{\prime}]^\top\right) \xrightarrow{p} 0.$ Then using the properties of convergence in probability and continuous mapping theorem, it can be shown that the reminder terms all converge to $0$ in probability.

$\hfill \blacksquare$

\section{Bisection Wald Test for the Identification of Sparsity Structure of $\AN$ }\label{app: Wald}
\begin{breakablealgorithm}
   \caption{}
   \label{alg: proj_Wald}
\begin{algorithmic}
    \STATE {\bfseries Input:} $\mb{x}_{t+1}, \mb{x}_t$, $\spc$, $\spcone$, $[\spc]^{-1}$, $t$.

    \COMMENT{Update:}
    \STATE $\widehat{\mb{\Gamma}}_{t+1}(1) \gets \frac{t}{t+1}\spcone + \frac{1}{t+1}\mb{x}_{t+1}\mb{x}_{t}^\top$, $\widehat{\mb{\Gamma}}_{t+1}(0) = \frac{t}{t+1}\spc + \frac{1}{t+1}\mb{x}_{t}\mb{x}_{t}^\top$, 

    \STATE 
    $[\widehat{\mb{\Gamma}}_{t+1}(0)]^{-1} \gets \frac{t+1}{t}[\spc]^{-1} -  \frac{t+1}{t} \frac{[\spc]^{-1}\mb{x}_t\mb{x}_t^\top[\spc]^{-1}}{t + \mb{x}_t^\top[\spc]^{-1}\mb{x}_t}$, \\
    
    \STATE
    $\widehat{\mb{\Sigma}}_{t+1} \gets \widehat{\mb{\Gamma}}_{t+1}(0) - \widehat{\mb{\Gamma}}_{t+1}(1)\widehat{\mb{\Gamma}}_{t+1}(0)^{-1}\widehat{\mb{\Gamma}}_{t+1}(1)^\top$,  $\widecheck{\A}_{t+1} \gets \widehat{\mb{\Gamma}}_{t+1}(1)[\widehat{\mb{\Gamma}}_{t+1}(0)]^{-1}$.
    
   \COMMENT{Projection:}
   
   \STATE $\widehat{\A}_{t+1} = \projKG(\widecheck{\A}_{t+1})$, retrieve  $\widehat{\A_{\mathrm{D}}}_{,t+1}, \widehat{\bm \AF}_{,t+1}, \widehat{\bm \AN}_{,t+1}$ using Equation \eqref{eq: projection_KG}.
   
   \STATE Sort such that: 
   $|(\widehat{\bm \AN}_{,t+1})_{i_1, j_1}| \leq ... \leq |(\widehat{\bm \AN}_{,t+1})_{i_{|\KN|}, j_{|\KN|}}|$.
   
   \COMMENT{Bisection Wald test procedure:}
   
   \STATE Initialize $p_l = 1, \; p_r = |\KN|, \; p_m = \texttt{Floor}(\frac{p_l + p_r}{2})$.
   \\
   \STATE Construct the corresponding test statistic $\lambda_{W,t+1}$ or $\lambda_{F,t+1}$ using Equation \eqref{eq: Wald statistic}. 
   \\
   \STATE Perform tests $H(1)$ and $H(|\KN|)$ based on Corollary \ref{coro: Wald test}.
   
   \IF{$H(1), \; H(|\KN|)$ are not rejected} 
   \STATE $\widehat{\bm \AN}_{,t+1} \gets 0$, 
   \ELSE
   \IF{$H(1), \; H(|\KN|)$ are both rejected}
   \STATE No changes are made to $\widehat{\bm \AN}_{,t+1}$, 
   
   \ELSE
   
   \WHILE{$p_l + 1 < p_r$}
   
   \STATE $p_m \gets \texttt{Floor}(\frac{p_l + p_r}{2})$, perform $H(p_m)$.
   \IF{$H(p_m)$ is not rejected}
   
   \STATE $p_l \gets p_m$,
   
   \ELSE
   \STATE $p_r \gets p_m$,
   \ENDIF
   \ENDWHILE
   
   \STATE Let $(\widehat{\bm \AN}_{,t+1})_{i_1, j_1} = ... = (\widehat{\bm \AN}_{,t+1})_{i_{p_l}, j_{p_l}} = 0$.

   \ENDIF
   
   \ENDIF

   \STATE $\widehat{\A}_{t+1} \gets \widehat{\A_{\mathrm{D}}}_{,t+1} +  \widehat{\bm \AF}_{,t+1} \otimes  \widehat{\bm \AN}_{,t+1}$, $t \gets t+1$.

   \STATE {\bfseries Output:} $\widehat{\A}_{t+1}, \widehat{\mb{\Gamma}}_{t+1}(0), \widehat{\mb{\Gamma}}_{t+1}(1), \widehat{\mb{\Gamma}}_{t+1}(0)^{-1}$, $t$.
\end{algorithmic}
\end{breakablealgorithm}

Note that, since multiplication with $\vect(U_{h_k})$ amounts to extracting elements in the matrix from the corresponding locations, in practice, we take the elements directly from $\left[\spc\right]^{-1}$ and $\widehat{\mb{\Sigma}}_t$, to compose $\widehat{\mb{\Sigma}}_{W,t}$ as:
$$
\begin{aligned}
\left(\widehat{\mb{\Sigma}}_{W,t}\right)_{k, k'} &= \left(\widehat{\mb{\Sigma}}_{W,t}\right)_{k', k} \\
& = \langle\mb{\Sigma}_{ii}^{k, k'}, \mb{\Gamma}_{jj}^{k, k'}\rangle + \langle\mb{\Sigma}_{jj}^{k, k'}, \mb{\Gamma}_{ii}^{k, k'}\rangle +
\langle\mb{\Sigma}_{ij}^{k, k'}, \mb{\Gamma}_{ji}^{k, k'}\rangle +
\langle\mb{\Sigma}_{ji}^{k, k'}, \mb{\Gamma}_{ij}^{k, k'}\rangle,
\end{aligned}
$$
where $\mb{\Sigma}_{ii}^{k, k'} = \left[\widehat{\mb{\Sigma}}_t\right]_{I_k,I_{k'}}, \; \mb{\Gamma}_{jj}^{k, k'} = \left[\spc^{-1}\right]_{J_k,J_{k'}}$, $\mb{\Sigma}_{jj}^{k, k'} = \left[\widehat{\mb{\Sigma}}_t\right]_{J_k,J_{k'}}\;, \mb{\Gamma}_{ii}^{k, k'} = \left[\spc^{-1}\right]_{I_k,I_{k'}}$, $\mb{\Sigma}_{ij}^{k, k'} = \left[\widehat{\mb{\Sigma}}_t\right]_{I_k,J_{k'}}\;, \mb{\Gamma}_{ji}^{k, k'} = \left[\spc^{-1}\right]_{J_k,I_{k'}}$, and $\; \Sigma_{ji}^{k, k'} = \left[\widehat{\mb{\Sigma}}_t\right]_{J_k,I_{k'}}\;$, $\mb{\Gamma}_{ij}^{k, k'} = \left[\spc^{-1}\right]_{I_k,J_{k'}}$, with order indices $I_k := \{i_k, i_k + F, ..., i_k + (N-1)F\}$, $I_{k'} := \{i_{k'}, i_{k'} + F, ..., i_{k'} + (N-1)F\}$, $J_k := \{j_k, j_k + F, ..., j_k + (N-1)F\}$, $J_{k'} := \{j_{k'}, j_{k'} + F, ..., j_{k'} + (N-1)F\}$.

\section{Extended Algorithm \ref{alg: proj_Wald} for the Augmented Model}\label{app: Wald ext}
\begin{algorithm}[H]
   \caption{}
   \label{alg: proj_Wald aug}
\begin{algorithmic}
    \STATE {\bfseries Input:} $\mb{x}_{t+1}, \mb{x}_t$, $\spc$, $\spcone$, $[\spc]^{-1}$, $\Bar{m}$, $t$, $\{p_{m,t}\}_{m=0}^{M-1}$, $\{\underline{\x}_{m,t}\}_{m=0}^{M-1}$.

    \STATE Update $\widehat{\mb{\Gamma}}_{t+1}(0)$, \, $\widehat{\mb{\Gamma}}_{t+1}(1)$ from Equation \eqref{eq: update autocov}.

    \STATE Update:
    
    \STATE \quad $[\widehat{\mb{\Gamma}}_{t+1}(0)]^{-1} \gets \frac{t+1}{t}[\spc]^{-1} -  \frac{t+1}{t} \frac{[\spc]^{-1}(\x_t - \underline{\x}_{\Bar{m}-1,t})(\x_t - \underline{\x}_{\Bar{m}-1,t})^\top[\spc]^{-1}}{t(1+1/p_{\Bar{m},t}) + (\x_t - \underline{\x}_{\Bar{m}-1,t})^\top[\spc]^{-1}(\x_t - \underline{\x}_{\Bar{m}-1,t})}$,  
    
    \quad $\widecheck{\A}_{t+1} \gets \widehat{\mb{\Gamma}}_{t+1}(1)[\widehat{\mb{\Gamma}}_{t+1}(0)]^{-1}$,
   
   \quad $\widehat{\mb{\Sigma}}_{t+1} \gets \widehat{\mb{\Gamma}}_{t+1}(0) - \widehat{\mb{\Gamma}}_{t+1}(1)[\widehat{\mb{\Gamma}}_{t+1}(0)]^{-1}\widehat{\mb{\Gamma}}_{t+1}(1)^\top$.
    
   \STATE Step \textit{Projection} to \textit{Bisection Wald test procedure} are identical to Algorithm \ref{alg: proj_Wald}.

   \STATE Let $\widehat{\A}_{t+1} = \widehat{\A_{\mathrm{D}}}_{,t+1} +  \widehat{\bm \AF}_{,t+1} \otimes  \widehat{\bm \AN}_{,t+1}$.
   
   \STATE Update:  
   \STATE \quad$\underline{\x}_{\Bar{m}-1,t+1} \gets \frac{p_{\Bar{m},t}}{p_{\Bar{m},t} + 1}\underline{\x}_{\Bar{m}-1,t} + \frac{1}{p_{\Bar{m},t} + 1}\x_{t}$, 
   \STATE \quad  $\underline{\x}_{m,t+1} \gets \underline{\x}_{m,t}, \forall m \neq \Bar{m}-1$,
    \STATE \quad $p_{\Bar{m},t+1} \gets p_{\Bar{m},t} + 1$,
    \STATE \quad $p_{m,t+1} \gets p_{m,t}, \forall m \neq \Bar{m}$.
   
   \STATE \quad $t \gets t+1$.

   \STATE {\bfseries Output:} $\widehat{\A}_{t+1}, \widehat{\mb{\Gamma}}_{t+1}(0), \widehat{\mb{\Gamma}}_{t+1}(1), [\widehat{\mb{\Gamma}}_{t+1}(0)]^{-1}$, $t$, $\{p_{m,t}\}_{m=0}^{M-1}$, $\{\underline{\x}_{m,t}\}_{m=0}^{M-1}$.
\end{algorithmic}
\end{algorithm}

\section{Proximal Gradient Descent for the proposed Lasso}\label{app: pgd}
The implementation of proximal gradient descent for Lasso \eqref{eq: matLasso} is given as follows.
\begin{equation}\label{eq-intro: pdg}
\begin{aligned}
    \A^{k+1} &= \mbox{prox}(\A^{k} - \eta^k \nabla f(\A^{k}) ), \\
    & = \argmin_{A \in \KG} \frac{1}{2\eta^k}  \left\|A -  \left(\A^{k} - \eta^k \nabla f(\A^{k})\right)\right\|_{\ell_2}^2 + \lambda_t F \left\|\AN\right\|_{\ell_1} \\
    & = \argmin_{A \in \KG} \frac{1}{2\eta^k}  \left\|A -  \projKG\left(\A^{k} - \eta^k \nabla f(\A^{k})\right)\right\|_{\ell_2}^2 + \lambda_t F \left\|\AN\right\|_{\ell_1}\\\vspace{0.1in}
    & \iff
    \begin{cases}
    &\A^{k+1}_{\mathrm{N}} =  \argmin\limits_{\AN} \left\|\AN -  \projKN\left(\A^{k} - \eta^k \nabla f(\A^{k})\right)\right\|_{\ell_2}^2 + 2\eta^k\lambda_t \left\|\AN\right\|_{\ell_1},   \\\vspace{0.05in}
    & \A^{k+1}_{\mathrm{F}} = \projKF\left(\A^{k} - \eta^k \nabla f(\A^{k})\right), \\\vspace{0.05in}
    & \diag(\A^{k+1}) = \projKD\left(\A^{k} - \eta^k \nabla f(\A^{k})\right),
    \end{cases}
\end{aligned}
\end{equation}
where $\nabla f(\A^{k}) = \A^{k}\spc - \spcone$, we denote $\A^{k+1}(t, \lambda_t)$ by $\A^{k+1}$ to avoid the heavy notation. 

The forward step requires to calculate the gradient only in $\R^{NF \times NF}$, then the backward step amounts to a classical Lasso after projecting the gradient onto $\KG$. Thus the structure constraint and the partial sparsity do not pose additional difficulties. 

\section{Matrix representation of the optimality condition of the proposed Lasso}\label{app: op mat}

The matrix representation of the optimality condition of Lasso \eqref{eq: matLasso} is given in terms of $\A$. To present it, we introduce the projections onto sub-spaces $\mathcal{K}_{\mathrm{N}^1} := \mbox{span}\{U_k: k \in \KN^1\}$ and $\mathcal{K}_{\mathrm{N}^0} := \mbox{span}\{U_k: k \in \KN^0\}$, denoted respectively by $\mbox{Proj}_{\mathrm{N}^1}$ and $\mbox{Proj}_{\mathrm{N}^0}$. Note that Equation \eqref{eq: direct sum} in fact admits 
$$\KG = \bigoplus_{k \in K}\mbox{span}\{U_k\}.$$
Thus 
$$
\mbox{Proj}_{\mathrm{N}^1}(B) = \sum_{k\in \KN^1} \langle U_k,B\rangle \, \frac{1}{\|U_k\|_\mathbf{F}^2} U_k =  I_F \otimes \left[\sum_{k\in \KN^1}\langle U_k,B\rangle E_k\right],
$$
and 
$$
\mbox{Proj}_{\mathrm{N}^0}(B) = \sum_{k\in \KN^0} \langle U_k,B\rangle \, \frac{1}{\|U_k\|_\mathbf{F}^2} U_k =  I_F \otimes \left[\sum_{k\in \KN^0}\langle U_k,B\rangle E_k\right].
$$
Then Equations \eqref{eq: op cond DF}, \eqref{eq: op cond KN1}, and \eqref{eq: op cond KN0} are equivalent respectively to 
\begin{equation}
  \mbox{Proj}_{\mathrm{DF}}\left(\A\spc - \spcone\right) = 0,
\end{equation}
\begin{equation}
  \mbox{Proj}_{\KN^1}\left(\A\spc - \spcone\right) + \lambda I_F \otimes \left[\sum_{k \in \KN^1} \mbox{sign}\langle E_k,\bm \AN\rangle \, E_k\right] = 0,
\end{equation}
\begin{equation}
  \mbox{Proj}_{\KN^0}\left(\A\spc - \spcone\right) + \lambda I_F \otimes \left[\sum_{k \in \KN^0} \partial |\langle E_k,\bm \AN\rangle| \, E_k\right] = 0,
\end{equation}
where $\A \in \KG$, $\mbox{Proj}_{\mathrm{DF}} = \mbox{Proj}_{\mathrm{D}} +  \mbox{Proj}_{\mathrm{F}}$, and $\partial |\langle E_k,\bm \AN\rangle| \in [-1,1]$.

\section{Homotopy Algorithm for Regularization Path $\A(t, \lambda_1)$ to $\A(t, \lambda_2)$}\label{app: homo1}
\begin{breakablealgorithm}
   \caption{}
   \label{alg: HomoAlgo1}
\begin{algorithmic}
    \STATE {\bfseries Input:} $N, F$, $\Ga$, $\gaone$, $\KN^1$ (ordered list), $\wNone$, $\lambda_1$, $\lambda_2$, $\left[\Ga^1\right]^{-1}$, where $\KN^1$, $\wNone$, $\left[\Ga^1\right]^{-1}$ are associated with $\A(t, \lambda_1)$, and $\wNone = [\mb{w}]_{\KN^1}$.

    \STATE \textbf{Initialization:} $\lambda \gets \lambda_1$, $\KN^0 \gets \KN \setminus \KN^1$, $K^1 \gets \KD + \KF + \KN^1$, where $+$ is the ordered append of two lists.
    
    \COMMENT{Computing the regularization path (the steps in parentheses are the modifications for the case $\lambda_1 > \lambda_2$):}
    
    \WHILE{$\lambda < \lambda_2$ (or $\lambda > \lambda_2$)}
   
    \STATE Generate $\Ga^0$, $\gaone^1$, $\gaone^0$, $\mb{w}_1$, based on Proposition \ref{prop: opcon}.
   
    \STATE Compute $\lambda_r$ (or $\lambda_l$), based on Equations \eqref{eq: crit_point} and \eqref{eq: lambda_range}.
    
    \IF{$\lambda_r < \lambda_2$ (or $\lambda_l > \lambda_2$)} 
    
     \STATE $\lambda \gets \lambda_r$ (or $\lambda \gets \lambda_l$), 
    
     \COMMENT{Update the active set and the sign vector:} 
    
    \IF{$[\as]_{i}$ becomes zero for some $k_i \in K^1 \mbox{ and } k_i \in \KN^1$, namely, $\lambda$ comes from $\{\lambda_k^0\}_k$}
    
     \STATE $\KN^1 \gets \KN^1 \setminus \{k\}$, $K^1 \gets K^1 \setminus \{k\}$, $\KN^0 \gets \KN^0 + \{k\}$. 
     
     \STATE Remove $[\wNone]_{i-|\KD|-|\KF|}$ from $\wNone$. 
    
     \STATE Remove the $i$-th row together with the $i$-th column from $\Ga^1$, and use Sherman Morrison formula to update  $\left[\Ga^1\right]^{-1}$. 
    
    \vspace{0.02in}
    
    \ELSIF{$[\mb{w}_0]_{i}$ reaches $1$ for some $k_i \in \KN^0$, namely, $\lambda$ comes from $\{\lambda_k^+\}_k$} 
    
     \STATE $\KN^0 \gets \KN^0 \setminus \{k\}$, $\KN^1 \gets \KN^1 + \{k\}$, $K^1 \gets K^1 + \{k\}$. 
     
     \STATE Append $1$ to the end of sign vector $\wNone$.
    
     \STATE Append row $[\Ga]_{k,K^1}$, column $[\Ga]_{K^1,k}$ after the last row and last column $\Ga^1$, respectively, and use Sherman Morrison formula to update  $\left[\Ga^1\right]^{-1}$. 
    
    \ELSIF{$[\mb{w}_0]_{k}$ reaches $-1$ for some $k_i \in \KN^0$, namely, $\lambda$ comes from $\{\lambda_k^-\}_k$} 
    
     \STATE $\KN^0 \gets \KN^0 \setminus \{k\}$, $\KN^1 \gets \KN^1 + \{k\}$, $K^1 \gets K^1 + \{k\}$. 
     
     \STATE Append $-1$ to the end of sign vector $\wNone$.

     \STATE Append row $[\Ga]_{k,K^1}$, column $[\Ga]_{K^1,k}$ after the last row and last column $\Ga^1$, respectively, and use Sherman Morrison formula to update  $\left[\Ga^1\right]^{-1}$. 
    
    \ENDIF
    
    \ELSE
    \STATE $\lambda \gets \lambda_2$.
    \ENDIF
    \ENDWHILE

    \STATE Compute $\as$, using Equation \eqref{eq: closed form} and the last updated $\left[\Ga^1\right]^{-1}$, $\gaone^1$,  $\mb{w}_1$. Retrieve $\A(t, \lambda_2)$ from this $\as$.
    
   \STATE {\bfseries Output:}
   $\A(t, \lambda_2)$, $\KN^1$, $\wNone$, $\left[\Ga^1\right]^{-1}$.
\end{algorithmic}
\end{breakablealgorithm}

\section{Homotopy Algorithm for Data Path $\A(t, \frac{t+1}{t}\lambda)$ to $\A(t+1, \lambda)$}\label{app: homo2}
\begin{breakablealgorithm}
   \caption{}
   \label{alg: HomoAlgo2}
\begin{algorithmic}[1]
    \STATE {\bfseries Input:} $N, F$, $\Ga$, $\gaone$, $\KN^1$ (ordered list), $\wNone$, $\lambda$, $\left[\Ga^1\right]^{-1}$, $\x_{t+1}$, $\widetilde{\mb{X}}_{t}$, $ t$, where $\KN^1$, $\wNone$, $\left[\Ga^1\right]^{-1}$ are associated with $\A(t, \frac{t+1}{t}\lambda)$, and $\wNone = [\mb{w}]_{\KN^1}$.

    \STATE \textbf{Initialization:} $\lambda \gets \lambda_1$, $\KN^0 \gets \KN \setminus \KN^1$, $K^1 \gets \KD + \KF + \KN^1$, where $+$ is the ordered append of two lists.

   \FOR{$i = 1, ..., NF$}
   \STATE $\mu \gets 0$.
    \WHILE{$\mu < 1$}
    \STATE Generate $\Ga^0$, $\gaone^1$, $\gaone^0$, $\mb{w}_1$, based on Proposition \ref{prop: opcon}.
    \STATE Let
    \STATE $\aslow = \left[\Ga^1\right]^{-1}(\gaone^1 - (1+\frac{1}{t}) \lambda\mb{w}_1)$, \; 
    
    \STATE $e = {\bm x}_{t+1,i} - ([\widetilde{\mb{X}}_t]_{K^1,i})^\top\aslow$, \; $\mb{u} = \left[\Ga^1\right]^{-1}[\widetilde{\mb{X}}_t]_{K^1,i}$, $\alpha = ([\widetilde{\mb{X}}_t]_{K^1,i})^\top\mb{u}$,
    \STATE   $\mu^0_{k_i} = -t(\aslow)_{i} / (\alpha(\aslow)_{i} + e(\mb{u})_{i}), \, k_i \in K^1 \mbox{ such that } k_i \in \KN^1$,
    \STATE     $\mu^{\pm}_{k_i} = \frac{-t(\mbox{b}^{\pm})_{i}}{e(\Ga^0\mb{u})_{i} - e(\widetilde{\mb{X}}_t)_{k,i} + \alpha(\mbox{b}^{\pm})_{i}}, \, k_i \in \KN^0$, \;  $\mbox{b}^{\pm} = \Ga^0\aslow - \gaone^0 \pm (1+\frac{1}{t}) \lambda$,
    
    \STATE     $\mu^{\prime} = \min\big\{\min\{\mu^0_k, k \in \KN^1: \mu^0_k > \mu\}, \min\{\mu^+_k, k \in \KN^0: \mu^+_k > \mu\}, \min\{\mu^-_k, k \in \KN^0: \mu^-_k > \mu\}\big\}$.
    
    \STATE    if $\mu^{\prime} = \emptyset$, $\mu^{\prime} \gets +\infty$.

    \IF{$\mu' < 1$}
    \STATE $\mu \gets \mu'$. 
    \IF{$\mu'$ is some $\mu^0_k$}
    \STATE $\KN^1 \gets \KN^1 \setminus \{k\}$, $K^1 \gets K^1 \setminus \{k\}$, $\KN^0 \gets \KN ^0 + \{k\}$. 
    \STATE Remove $[\wNone]_{i - |\KD| - |\KF|}$ from $\wNone$.
    \STATE Remove the $i$-th row, the $i$-th column from $\Ga^1$, use Sherman Morrison formula to update $\left[\Ga^1\right]^{-1}$. 
    \ELSIF{$\mu'$ is some $\mu^+_k$ (or $\mu^-_k$)}
    \STATE $\KN^0 \gets \KN^0 \setminus \{k\}$, $\KN^1 \gets \KN^1 + \{k\}$, $K^1 \gets K^1 + \{k\}$. 
    
    \STATE Append $1$ (or $-1$) to the end of sign vector $\wNone$.
    \STATE Append row $[\Ga]_{k,K^1}$, column $[\Ga]_{K^1,k}$ after the last row and last column $\Ga^1$, respectively, and use Sherman Morrison formula to update  $\left[\Ga^1\right]^{-1}$. 
    \ENDIF
    \ELSE 
        \STATE $\mu \gets 1$.
    \ENDIF
    \ENDWHILE
    \STATE   $[\Ga^1]^{-1} \stackrel{\mbox{rank 1 update}}{\longleftarrow} [\Ga^1 + \frac{1}{t}[\widetilde{\mb{X}}_t]_{K^1,i}([\widetilde{\mb{X}}_t]_{K^1,i})^\top]^{-1}$.
    
    \STATE   $\Ga \gets \Ga + \frac{1}{t}[\widetilde{\mb{X}}_t]_{:,i}[\widetilde{\mb{X}}_t]_{:,i}^\top$. $\gaone \gets \gaone + \frac{1}{t}{\bm x}_{t+1,i}[\widetilde{\mb{X}}_t]_{:,i}$.
    
    \ENDFOR
    \STATE $\as \gets \aslow + e\mb{u}/(t + \alpha)$. Retrieve $\A(t+1, \lambda)$ based on $K^1$ and $\as$. 
    
    \STATE $[\Ga^1]^{-1} \gets \frac{t+1}{t}[\Ga^1]^{-1}$, $\Ga \gets \frac{t}{t+1}\Ga$, $\gaone \gets \frac{t}{t+1}\gaone$.
    
   \STATE {\bfseries Output:}
   $\A(t+1, \lambda)$, $\KN^1$, $\wNone$, $\left[\Ga^1\right]^{-1}$, $\Ga$, $\gaone$.
\end{algorithmic}
\end{breakablealgorithm}

\section{Online Graph and Trend Learning from Matrix-variate Time Series in High-dimensional Regime}\label{app: homo aug}
\begin{breakablealgorithm}
   \caption{}
   \label{alg: online matrix Lasso aug}
\begin{algorithmic}
   \STATE {\bfseries Input:} $\A(t, \lambda_t)$, $\Ga$, $\gaone$, $\KN^1$ (ordered list), $\wNone$, $\lambda_t$, $\left[\Ga^1\right]^{-1}$, $\x_{t+1}$, $\widetilde{\mb{X}}_{t}$, $\Bar{m}$, $t$, $M$,  $(p_{m,t})_{m=0}^{M-1}$,  $(\underline{\x}_{m,t})_{m=0}^{M-1}$, $\mb{b}_{\Bar{m},t}$, where $\KN^1$, $\wNone$, $\left[\Ga^1\right]^{-1}$ are associated with $\A(t, \lambda_t)$.
   \STATE Select $\lambda_{t+1}$ according to the end of Section \ref{sec:app2 aug}.
   \STATE Update $\A(t, \lambda_t) \rightarrow \A(t, \frac{t+1}{t}\lambda_{t+1})$ using algorithm \ref{alg: HomoAlgo1}.
   \STATE Center $\x_{t+1} \gets \x_{t+1} - \underline{\x}_{\Bar{m},t}$. Compose $\underline{\widetilde{\mb{X}}}_{\Bar{m}-1,t}$ as  $[\underline{\widetilde{\mb{X}}}_{\Bar{m}-1,t}]_{k,i} = [U_k]_{i,:}\underline{\x}_{\Bar{m}-1,t}$, and center  $\widetilde{\mb{X}}_t \gets \widetilde{\mb{X}}_t - \underline{\widetilde{\mb{X}}}_{\Bar{m}-1,t}$.
   \STATE Update $\A(t, \frac{t+1}{t}\lambda_{t+1}) \rightarrow \A(t+1, \lambda_{t+1})$ using algorithm \ref{alg: HomoAlgo2}, with modifications:
   
   \quad Line $9$ change to $\alpha = [\widetilde{\mb{X}}_t]_{K^1,i}^\top\mb{u} + p_{\Bar{m},t}$,
   
   \quad Line $29, 30$ change respectively to: 
   
   \quad \quad $[\Ga^1]^{-1} \stackrel{\mbox{rank 1 update}}{\longleftarrow} [\Ga^1 + \frac{p_{\Bar{m},t}}{t(p_{\Bar{m},t}+1)}[\widetilde{\mb{X}}_t]_{K^1,i}[\widetilde{\mb{X}}_t]_{K^1,i}^\top]^{-1}$,
    
    \quad \quad $\Ga \gets \Ga + \frac{p_{\Bar{m},t}}{t(p_{\Bar{m},t}+1)}[\widetilde{\mb{X}}_t]_{:,i}[\widetilde{\mb{X}}_t]_{:,i}^\top$,
    
    \quad \quad $\gaone \gets \gaone + \frac{p_{\Bar{m},t}}{t(p_{\Bar{m},t}+1)}{\bm x}_{t+1,i}[\widetilde{\mb{X}}_t]_{:,i}$.

    \STATE Update:
    
    \STATE \quad $\underline{\x}_{\Bar{m}-1,t+1} \gets \frac{p_{\Bar{m},t}}{p_{\Bar{m},t} + 1}\underline{\x}_{\Bar{m}-1,t} + \frac{1}{p_{\Bar{m},t} + 1}\x_{t}$,
    \STATE \quad $\underline{\x}_{m,t+1} \gets \underline{\x}_{m,t}, \forall m \neq \Bar{m}-1$,
    \STATE \quad $p_{\Bar{m},t+1} \gets p_{\Bar{m},t} + 1$, 
    \STATE \quad $p_{m,t+1} \gets p_{m,t}, \forall m \neq \Bar{m}$,
   \STATE \quad $\Bar{m}' \gets (t+2) \; \mbox{mod} \; M$,
   \STATE\quad $\mb{b}_{\Bar{m}',t+1} \gets \underline{\x}_{\Bar{m}',t+1} - \A(t+1, \lambda_{t+1})\underline{\x}_{\Bar{m},t+1}$,
   \STATE \quad $t \gets t + 1.$
   \STATE {\bfseries Output:} $\A(t+1, \lambda_{t+1})$, $\Ga$, $\gaone$, $\KN^1$, $\wNone$, $\lambda_{t+1}$, $\left[\Ga^1\right]^{-1}$, $t$, $(p_{m,t+1})_{m=0}^{M-1}$,  $(\underline{\x}_{m,t+1})_{m=0}^{M-1}$, $\mb{b}_{\Bar{m}',t+1}$.
\end{algorithmic}
\end{breakablealgorithm}

\bibliography{main}
\bibliographystyle{elsarticle-harv} 
\end{document}